\documentclass[10pt,twocolumn,letterpaper]{article}

\usepackage{iccv}
\usepackage{times}
\usepackage{epsfig}
\usepackage{graphicx}
\usepackage{amsmath}
\usepackage{amssymb}

\usepackage[ruled,vlined]{algorithm2e}
\usepackage{multirow}
\usepackage{caption}
\usepackage{subcaption}
\usepackage[pagebackref=true,breaklinks=true,letterpaper=true,colorlinks,bookmarks=false]{hyperref}
\usepackage{float}

\usepackage[breaklinks=true,bookmarks=false]{hyperref}

\iccvfinalcopy 

\def\iccvPaperID{10790} 

\ificcvfinal\fi

\begin{document}

\title{Reconstructing Small 3D Objects in front of a Textured Background
\thanks{The authors are affiliated with the Czech Institute of Informatics, Robotics and Cybernetics, Czech Technical University in Prague, CZ. This research was supported by the EU Structural and Investment Funds, Operational Programe Research, Development and Education under the project IMPACT (reg.\ no.\ CZ$.02.1.01/0.0/0.0/15\_003/0000468$), EU H2020 ARtwin No.~856994, and EU H2020 SPRING No.~871245.}}
\author{Petr Hruby\\
CIIRC, CTU in Prague\\
{\tt\small hrubype7@fel.cvut.cz}
\and
Tomas Pajdla\\
CIIRC, CTU in Prague\\
{\tt\small pajdla@cvut.cz}
}
\maketitle
\begin{abstract}
\noindent 
We present a technique for a complete 3D reconstruction of small objects moving in front of a textured background. It is a particular variation of multibody structure from motion, which specializes to two objects only. The scene is captured in several static configurations between which the relative pose of the two objects may change. We reconstruct every static configuration individually and segment the points locally by finding multiple poses of cameras that capture the scene's other configurations. Then, the local segmentation results are combined, and the reconstructions are merged into the resulting model of the scene. In experiments with real artifacts, we show that our approach has practical advantages when reconstructing 3D objects from all sides. In this setting, our method outperforms the state-of-the-art. We integrate our method into the state of the art 3D reconstruction pipeline COLMAP.
\end{abstract}

\section{Introduction}

\noindent Structure-from-Motion (SfM)~\cite{schoenberger2016sfm} is an important problem of estimating the 3D model of a scene from two-dimensional images of the scene~\cite{DBLP:journals/tog/SnavelySS06,DBLP:journals/ijcv/SnavelySS08,schoenberger2016sfm,DBLP:conf/iccvw/EnqvistKO11,DBLP:conf/iccv/MagerandB17}, image matching~\cite{DBLP:conf/nips/RoccoCATPS18,Dusmanu-ICCV-2019}, visual odometry~\cite{DBLP:conf/cvpr/NisterNB04,DBLP:journals/ral/AlismailKBL17} and visual localization~\cite{DBLP:journals/pami/SattlerLK17,DBLP:journals/pami/SvarmEKO17,DBLP:conf/cvpr/TairaOSCPSPT18}. 

\begin{figure}[t]
  \begin{center}
    \begin{tabular}{cc}
      \hspace*{-0.5cm}\begin{tabular}[b]{cc}
        \includegraphics[width=0.2\linewidth]{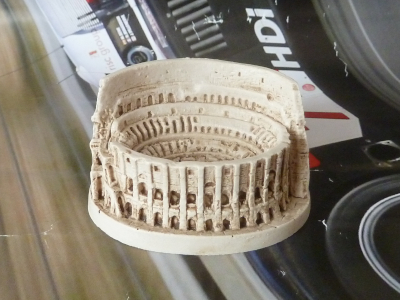}&
        \includegraphics[width=0.2\linewidth]{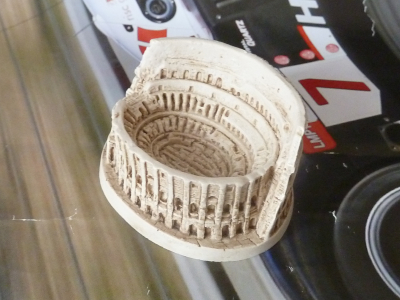}\\
        \includegraphics[width=0.2\linewidth]{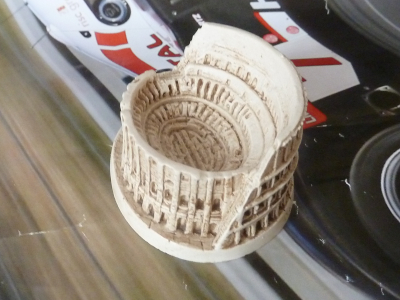}&
        \includegraphics[width=0.2\linewidth]{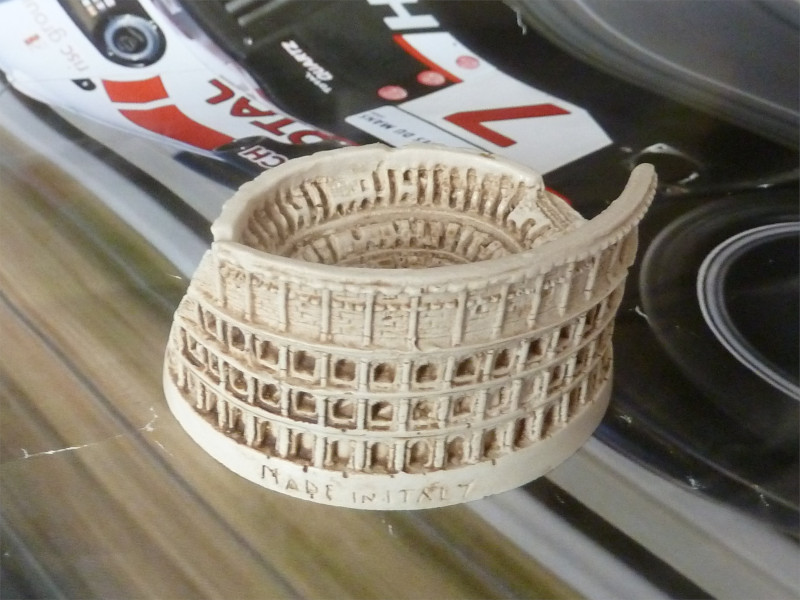}
      \end{tabular}
      & \hspace*{-0.5cm}\includegraphics[width=0.5\linewidth]{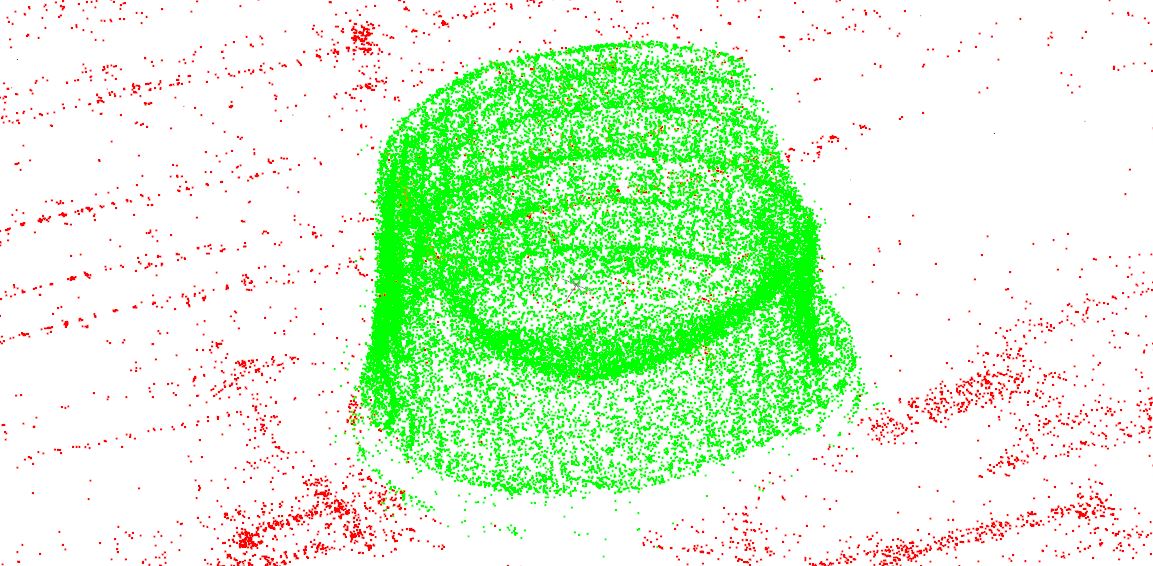}
    \end{tabular}
\end{center}
\caption{(left) Multiple images of an artefact moving in front of a textured background. (right) Reconstructed 3D model capturing the artefact from all sides (green) together with the reconstructed background (red). Our TBSfM segments artifacts from the background and delivers 3D reconstruction of both rigid bodies. Background helps to obtain more 3D points artefacts and reduces reprojection errors, see Tab.~\ref{tab:error}.}
    \label{fig:output}
\end{figure}

In the digitization of cultural heritage~\cite{8939087}, movie~\cite{DBLP:conf/grapp/BullingerBA21} and game~\cite{epic-games} industries, it is crucial to reconstruct complete models of small objects. Often, this is done by running an SfM pipeline on images of small objects presented from all sides on a featureless background~\cite{DBLP:conf/avr/PaolisLGDP20}. Small objects, however, often have repetitive structures and are close to planar in some views. Hence, they do not provide reliable camera poses, which leads to low quality 3D reconstructions. This can be remedied by capturing a structured background together with the object, Fig.~\ref{fig:output}. Then, however, segmentation is {\em needed} to reconstruct the moving object in front of the background. To this end, we develop the TBSfM method, a particular variation of two-body SfM for high-quality complete 3D reconstruction of objects moving in front of a textured background. 

SfM is well understood for static scenes with a single rigid object and has many practical applications in, e.g., cartography~\cite{cartography}, archaeology~\cite{archaeology}, and film industry~\cite{specialeffects}. In the case of dynamic scenes, the majority of works~\cite{DBLP:conf/eccv/FitzgibbonZ00,Tola05,DBLP:journals/tip/QianCZ05,DBLP:journals/ijcv/SchindlerSW08,DBLP:journals/pami/OzdenSG10,DBLP:conf/ismar/RoussosRGA12} and \cite{DBLP:conf/iccv/FayadRA11,DBLP:conf/eccv/RussellYA14,DBLP:conf/iccv/KunduKJ11,DBLP:conf/iccv/KumarDL17,DBLP:conf/cvpr/VoNS16,DBLP:conf/nips/FragkiadakiSAM14} assume video input, while the case of unordered input images has not been fully explored yet. To our best knowledge, there is no full Multi-Body SfM (MBSfM) pipeline producing satisfactory results in industrial quality. Existing partial results~\cite{DBLP:journals/tip/QianCZ05,DBLP:journals/pami/OzdenSG10,DBLP:journals/ijcv/SchindlerSW08,Srajer16} indicate that the multi-body segmentation in SfM is a difficult task. Hence, we address a specialized version of an MBSfM problem by restricting ourselves to two objects and to a semidynamic acquisition of the images of the scene. Although the setting is more restrictive when compared to the general MBSfM, the problem is still challenging and has practical use. 


We consider unordered and unstructured input images, which is a typical setup for SfM 3D reconstruction. Therefore, the works which assume video input, for example~\cite{DBLP:conf/eccv/FitzgibbonZ00,Tola05,DBLP:journals/tip/QianCZ05,DBLP:journals/ijcv/SchindlerSW08,DBLP:journals/pami/OzdenSG10,DBLP:conf/ismar/RoussosRGA12,DBLP:conf/iccv/FayadRA11,DBLP:conf/eccv/RussellYA14,DBLP:conf/iccv/KunduKJ11,DBLP:conf/iccv/KumarDL17,DBLP:conf/cvpr/VoNS16,DBLP:conf/nips/FragkiadakiSAM14}, cannot solve our problem. We also assume that the images may capture the scene from different viewpoints and most of the points are observed only by a small portion of cameras. The MBSfM works based on factorization, e.g.,~\cite{DBLP:conf/cvpr/VidalMS03,LSA,DBLP:journals/ijcv/VidalTH08,DBLP:conf/cvpr/RaoTVM08,DBLP:journals/pami/LiuLYSYM13,DBLP:journals/pami/ElhamifarV13,DBLP:conf/cvpr/LiV15,DBLP:conf/iccv/JiSL15,DBLP:journals/ijcv/CosteiraK98,DBLP:conf/cvpr/LiKSV07,DBLP:conf/icra/SabzevariS14,DBLP:conf/3dim/KumarDL16}, use a track completion, as the factorization itself requires complete tracks. However, in our setting, the track completion is an underconstrained problem, as~\cite{DBLP:journals/tit/CandesT10} proves the existence of a lower bound on the number of samples present in a matrix, below which no algorithm can guarantee a successful matrix completion. (Sec.~\ref{sec:exp1}) Other MBSfM works, e.g.~\cite{DBLP:conf/iccv/LiGCZ13,DBLP:journals/tits/LaiWYCZ17,Xu18}, evaluate the distances between each pair of points and therefore are not suitable for large scenes with more than tens of thousands of points because the distances are unable to fit into computer memory. We assume arbitrarily large scenes, therefore, these methods are not suitable for our setting too.

\subsection{Contribution}
We present Two-Body Structure from Motion for Semidynamic Scenes, a generalization of SfM to two independently moving objects in the scene, assuming that the scene is captured in several static configurations between which the relative pose of the two objects may change. Our TBSfM  reconstructs every static configuration individually and segments the points locally by finding multiple poses of cameras which capture the other configurations of the scene. We demonstrate that TBSfM has a practical use in the reconstruction of small artifacts and that it outperforms the state-of-the-art. It is able to improve the quality of the reconstruction of small and repetitive objects when placed on a textured background. Most of the existing MBSfM pipelines reconstruct each object individually, and therefore, they are not able to resolve the problem of scale consistency. To overcome this problem, we reconstruct the foreground object together with the background and segment points afterwards, which is possible thanks to the semidynamic setting. Then, the scale of the foreground model is consistent with the scale of the background model.

Our method has two significant advantages when compared to the general approach to MBSfM, which is to first segment tracks and then reconstruct each object individually. First, we evade the segmentation of the 2D tracks, which is a difficult task which might be degenerated \cite{Hartley:2003:MVG:861369}. Secondly, the experiments show that the presence of the background in the reconstruction step may significantly improve the result. (See Section \ref{sec:exp2})

\subsection{Related work}\label{sec:previous-work} 
\noindent General MBSfM is closely related to Motion segmentation (MS), which clusters points into different motions. Two view MS can be achieved with multiple model fitting. MS methods use consensus analysis~\cite{DBLP:journals/prl/XuOK90,Vincent01,Zuliani05,DBLP:conf/cvpr/MagriF16}, preference analysis~\cite{DBLP:conf/eccv/ZhangK06,DBLP:conf/eccv/ToldoF08,DBLP:conf/cvpr/ChinSW10,DBLP:conf/cvpr/MagriF14,DBLP:conf/cvpr/PhamCYS12,DBLP:conf/bmvc/MagriF15}, energy minimization~\cite{PEARL,DBLP:conf/eccv/BarathM18}, branch and bound~\cite{DBLP:journals/tkde/ThakoorG11}, subspace segmentation \cite{DBLP:journals/ijcv/VidalMSS06}, or information theory \cite{Torr98}. Three-view MS is described in~\cite{DBLP:journals/ijcv/VidalTH08}. Some multiview MS methods use subspace segmentation~\cite{DBLP:conf/cvpr/VidalMS03,LSA,DBLP:journals/ijcv/VidalTH08,DBLP:conf/cvpr/RaoTVM08,DBLP:journals/pami/LiuLYSYM13,DBLP:journals/pami/ElhamifarV13,DBLP:conf/cvpr/LiV15,DBLP:conf/iccv/JiSL15}. The subspace segmentation requires complete tracks. Therefore, these methods demand a completion of the tracks to pass from incomplete tracks to the complete ones. The completion of tracks works only when a small number of observations are missing. In our case, the points are observed by a few cameras and, therefore, the track completion is an underconstrained problem. Other methods combine results from two view methods~\cite{DBLP:conf/iccv/LiGCZ13,DBLP:journals/tits/LaiWYCZ17,Xu18,DBLP:journals/corr/abs-1905-09043}. Except for \cite{DBLP:journals/corr/abs-1905-09043}, these methods evaluate the distances between each pair of points and, therefore, are not suitable for large scenes due to their spatial complexity. Work~\cite{DBLP:conf/iros/KunduKS09} assumes that the motion is known a priory and detects the moving objects as points inconsistent with the motion. The goal of MBSfM is to segment moving objects in the scene and to provide their 3D reconstruction. Works~\cite{DBLP:conf/eccv/FitzgibbonZ00,Tola05,DBLP:journals/tip/QianCZ05,DBLP:journals/ijcv/SchindlerSW08,DBLP:journals/pami/OzdenSG10,DBLP:conf/ismar/RoussosRGA12,DBLP:conf/iccv/FayadRA11} and \cite{DBLP:conf/eccv/RussellYA14,DBLP:conf/iccv/KunduKJ11,DBLP:conf/iccv/KumarDL17,DBLP:conf/cvpr/VoNS16,DBLP:conf/nips/FragkiadakiSAM14} require video sequences as the input. Work \cite{DBLP:conf/eccv/FitzgibbonZ00} introduces a constraint on internal parameters and segments points together with building the tracks. In~\cite{Tola05}, the initial segmentation is found and propagated throughout the video sequence. In~\cite{DBLP:journals/tip/QianCZ05} a matrix indicating the similarity between two points is found and decomposed with SVD. \cite{DBLP:journals/ijcv/SchindlerSW08} builds tentative motions and selects the most relevant ones with an information criterion. In~\cite{DBLP:journals/pami/OzdenSG10}, 3D reconstruction and motion segmentation are performed simultaneously. The work handles incomplete tracks and changing number of motions. In~\cite{DBLP:conf/ismar/RoussosRGA12}, the structure, motion and segmentation are found simultaneously with energy minimization and the output is a dense reconstruction. Work~\cite{DBLP:conf/iccv/KunduKJ11} performs SfM with additional feedback from motion segmentation and from a particle filter. Work~\cite{DBLP:conf/iccv/FayadRA11} handles articulated motions. Non-rigid motions are addressed in~\cite{DBLP:conf/eccv/RussellYA14,DBLP:conf/iccv/KumarDL17,DBLP:conf/cvpr/VoNS16,DBLP:conf/nips/FragkiadakiSAM14}. Works~\cite{DBLP:journals/ijcv/CosteiraK98,DBLP:conf/cvpr/LiKSV07,DBLP:conf/icra/SabzevariS14,DBLP:conf/3dim/KumarDL16} use subspace clustering. All the above methods require complete tracks. While~\cite{DBLP:journals/ijcv/CosteiraK98} assumes affine projection, \cite{DBLP:conf/cvpr/LiKSV07,DBLP:conf/icra/SabzevariS14} handle perspective projection. \cite{DBLP:conf/3dim/KumarDL16} handles multiple nonrigid motions. Work~\cite{DBLP:journals/cviu/ZappellaBLS13} requires the motion segmentation as its input, it outputs a corrected segmentation together with the 3D model.

{\bf The most relevant previous work} is YASFM \cite{Srajer16}. Unlike the majority of the methods, \cite{Srajer16} handles unordered input images and does not require complete tracks. In \cite{Srajer16}, the planes are segmented with growing homographies and grouped together if they belong to the same motion according to a planar homology score. Unfortunately, due to the greedy nature of the grouping algorithm, \cite{Srajer16} typically fails if there is no motion between some images. Our work is an extension of COLMAP SfM pipeline~\cite{schoenberger2016sfm}, which already contains some elements used in MBSfM. For instance, it can verify matches from different objects. However, \cite{schoenberger2016sfm} does not deal with dynamic scenes with multiple objects.
\section{Problem formulation}
\noindent We will now formulate TBSfM problem. We use the nomenclature from~\cite{Hartley:2003:MVG:861369}, $\pi$ denotes the calibrated perspective projection, i.e., $\pi([x,y,z]) = [x/z, y/z]$. $[1,n]$ is an abbreviation of $1,...,n$.

The scene we wish to reconstruct consists of two objects; the background and the foreground. The scene contains $n$ points and is captured in $k$ different static configurations by image sequences called {\em takes}. The background remains static, while the foreground moves between the takes. 

Let $i \in [1,n]$ denote a point id. We introduce point-labeling function $o(i) \in \{ B, F, U \}$, which assigns points to the foreground (F), background (B), and marks as unknown (U) points that were not assigned to F or B, e.g.\ due to mismatches. Then, $X_{i, \delta}^{o(i)} \in \mathbb{R}^3$ denotes the $i^{th}$ point in the initial configuration and in the world coordinate system $\delta$.

\begin{figure}[t]
\begin{center}
\begin{tabular}{c}
    \includegraphics[width=0.85\linewidth]{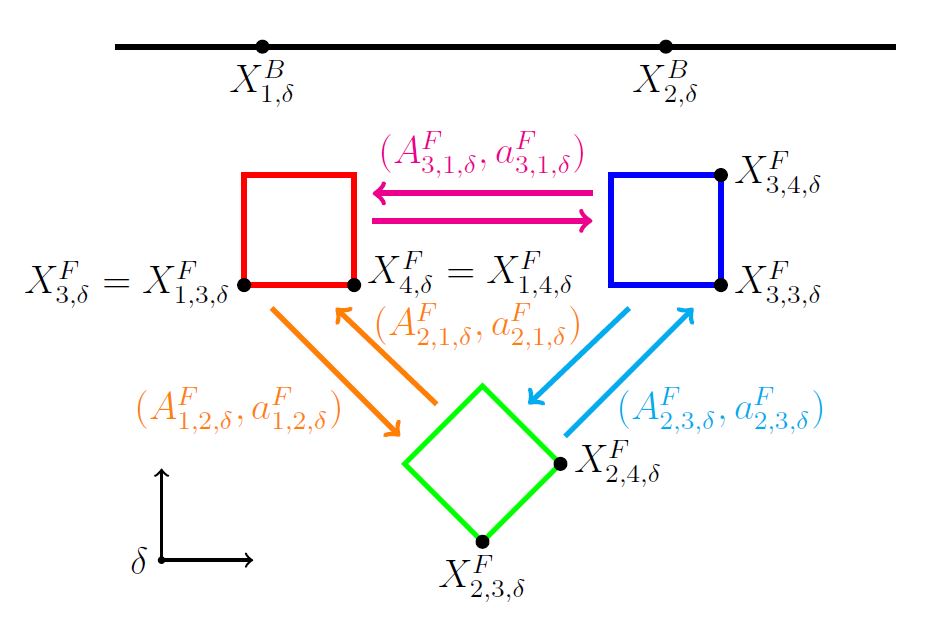} \\
    (a)\\
    \includegraphics[width=0.85\linewidth]{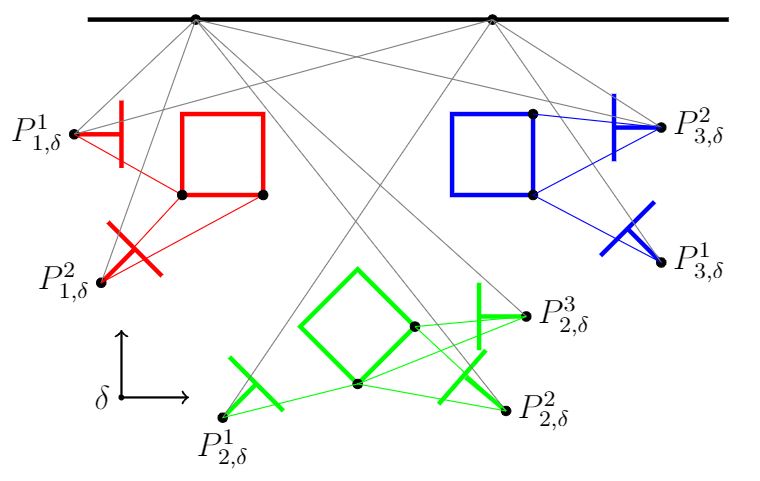} \\
    (b)
\end{tabular}
\end{center}
   \caption{Example of a scene with $k=3$ takes and $m=4$ points. The red, green and blue boxes represent three different positions of the object. The black line represents the background. (a) The arrows represent the motions between those. (b) The scene is observed with 7 cameras. Every take is observed by the cameras of the corresponding color. }
   \label{fig:scene1}
\end{figure}

\subsection{Motion between the takes}
\noindent We will now describe how the points move between different configurations.

The points appear in different positions according to take $t \in [1,k]$, in which they are captured. Let $X_{t, i, \delta}^{o(i)} \in \mathbb{R}^3$ be the position of point $i$ in take $t$ and in the world coordinate system $\delta$. Let $(A_{t,\delta}^{o}, a_{t,\delta}^{o})$, with rotation $A_{t,\delta}^{o}$ and translation  $a_{t,\delta}^{o}$, denote the motion of object $o \in \{ B, F \}$ from the initial configuration to take $t$. Then we have
\begin{equation}
    X_{t, i, \delta}^{o(i)} = A_{t,\delta}^{o(i)} X_{i, \delta}^{o(i)} + a_{t,\delta}^{o(i)}
\end{equation}
The background remains static and the initial position is equal to take $1$, therefore if $o=B$ or $t=1$, we can write
\begin{equation}
    A_{t,\delta}^o = I; a_{t,\delta}^o = 0
\end{equation}
Let $(A_{s,t,\delta}^{o}, a_{s,t,\delta}^{o})$ denote the motion of object $o$ from take $s$ to take $t$. This motion is composed of the motions between the takes and the initial position as
\begin{equation}
    A_{s,t,\delta}^{o} = A_{t, \delta}^o (A_{s, \delta}^o)^{-1}; \
    a_{s,t,\delta}^{o} = a_{t,\delta}^{o(i)} - A_{t,\delta}^{o(i)} (A_{s,\delta}^{o(i)})^{-1} a_{s,\delta}^{o(i)}
\end{equation}
\noindent A scene with the motions is illustrated in Fig.~\ref{fig:scene1}(a).

\subsection{Cameras observing the configurations}
\noindent Next, we explain how the configurations of the scene are observed by cameras. 

Every take $t \in [1,k]$ is captured by $m_t \geq 2$ cameras. We assume the perspective camera model~\cite{Hartley:2003:MVG:861369}. Let $m=\sum_{t=1}^k~m_t$ be the total number of cameras. Let~$j~\in~[1,m]$ be the id of a camera. Let $t(j)$ be the take observed by the camera $j$. Let $P_{\delta}^{j} = K^{j} \begin{bmatrix} R_{\delta}^{j} \ | -R_{\delta}^{j} c_{\delta}^{j} \end{bmatrix}$ denote the camera matrix and $x_{t(j), i}^{j}$ denote the projection of point $X_{t(j), i, \delta}^{o(i)}$ into camera $P_{\delta}^{j}$. The point is projected as
\begin{eqnarray}
    \alpha x_{t(j), i}^{j} &=& K^{j} R_{\delta}^{j} (X_{t(j), i, \delta}^{o(i)} - c_{\delta}^{j}); \ \alpha \in \mathbb{R} \\
    \alpha x_{t(j), i}^{j} &=& K^{j} R_{\delta}^{j} (A_{t(j), \delta}^{o(i)} X_{i, \delta}^{o(i)} + a_{t(j), \delta}^{o(i)} - c_{\delta}^{j})
\end{eqnarray}
The cameras observing the scene are illustrated in Fig.~\ref{fig:scene1}(b).
\subsection{The two-body SfM task}
\noindent We will now formulate our main task. 

First, we define the reprojection error of the observation $x_{t(j),i}^{j}$ as follows. For point labels $o(i) \in \{B, F\}$, we set
\begin{equation*}
e_{t(j),i}^{j} = \lVert \pi(x_{t(j),i}^{j}) - \pi(R_{\delta}^{j} ( A_{t(j),\delta}^{o(i)} X_{t(j), i, \delta}^{o(i)} + a_{t(j),\delta}^{o(i)} - c_{\delta}^{j}) \lVert 
\end{equation*}
For $o(i) \in \{U\}$, we set it to a constant value $\epsilon$, i.e., \ $e_{t,i}^{j_t} = \epsilon$.

Then, introducing points $\theta_1 = \{X_{i,\delta}^{o(i)}; i = [1,n]\}$, foreground motions $\theta_2 = \{(A_{t, \delta}^F, a_{t, \delta}^F); t = [2, k] \}$, and camera projections  $\theta_3 = \{ P_{\delta}^{j}; , j = [1,m] \} $, the scene can be described with parameters $\theta = \{ o(i), \theta_1, \theta_2, \theta_3 \}$.

\vspace*{1ex}
\noindent {\bf The TBSfM task is to}

\noindent{\em  1) Segment the scene into two objects, F and B, and points with label U by finding the  optimal labeling
\begin{equation}
    o(i)^* = \underset{\theta}{\operatorname{argmax}} \#(e_{t(j), i}^{j} < \epsilon) \label{eq:task1}
\end{equation}
with $o(i)^* = U$ when $e_{t(j),i}^{j} \geq \epsilon$.

\noindent 2) Reconstruct the two objects in the initial position from input observations $x_{t(j), i}^{j}$} by finding optimal parameters $\phi = \{ \theta_1, \theta_2, \theta_3 \}$
\begin{equation}
    \phi^* = \underset{\phi}{\operatorname{argmin}} \sum_{i, j, t(j)} (e_{t(j), i}^{j})^2 \label{eq:task2}
\end{equation}
%
The value of the function \eqref{eq:task2} does not depend on the points labeled by U because their reprojection error is set to the constant $\epsilon$.

\section{Proposed TBSfM method}
\noindent TBSfM task is known to be hard and its exact solution is not feasible~\cite{Nister-ICCV-2007}. We thus propose a greedy approximate solution and show that it achieves good results in practice.

Our method, which is concisely presented in Alg.~\ref{alg:TBFfM}, starts with reconstructing static configurations with a state-of-the-art pipeline~\cite{schoenberger2016sfm}. Then, it finds the poses of the cameras from other takes towards the background and towards the foreground. Inliers of these poses give {\em local segmentations}. Then, the poses are grouped to obtain the global segmentation to merge partial reconstructions of the static configurations into the resulting model.
\begin{algorithm}[b]
\SetAlgoLined
\SetKwInOut{Input}{input}\SetKwInOut{Output}{output}
\Input{Images of the scene in different takes}
\Output{Model of the scene with point labelling $o(i)$}
\For{take $t \in [1,k]$}
{
    (Sec.~\ref{sec:recostruct-takes}) Reconstruct model of take $t$\;
    (Sec.~\ref{sec:cam-registration}) Sequentially register cameras from other takes $s, s \neq t$ towards $t$\;
    (Sec.~\ref{sec:loc-group}) Merge the observations of the sequentially registered cameras\;
}
(Sec.~\ref{sec:segmentation}) Merge the groups of points from different takes\;
(Sec.~\ref{sec:rec-merge}) Merge the models according to \cite{Arun87}\;
Perform Bundle Adjustment \cite{Triggs00} where function \eqref{eq:task2} is minimized\;
\caption{Two Body SfM.}\label{alg:TBFfM}
\end{algorithm}
\noindent We will next describe the individual elements of  the TBSfM algorithm.
\subsection{Reconstruction of the takes}\label{sec:recostruct-takes}
\noindent The input of the method is a set of images grouped into takes. The first step is to reconstruct every take separately by COLMAP SfM pipeline~\cite{schoenberger2016sfm}. The output is a set of $k$ models with matches between the corresponding 3D points.

The models reconstructed separately from the takes are in their own coordinate systems $\delta_t$, $t \in [1, k]$. Let $X_{t, i, \delta_t}^{o(i)}$ be the position of point $X_{t, i, \delta}^{o(i)}$ in coordinate system $\delta_t$. Let $(B_{t}, b_{t}, \beta_t)$, $B_t \in SO(3), b_t \in \mathbb{R}^3, \beta_t \in \mathbb{R}^+$ denote the change of coordinates between systems $\delta$ and $\delta_t$. Then,
\begin{equation}
    X_{t, i, \delta_t}^{o(i)} = \beta_t B_t X_{t, i, \delta}^{o(i)} + b_t
\end{equation}
The change of coordinates $(B_{s,t}, b_{s,t}, \beta_{s,t})$ between systems $\delta_s$, $\delta_t$ is composed from the changes of coordinates between the systems and the world coordinate system $\delta$ as
\begin{equation}
    B_{s,t} = B_t B_s^{-1};\ b_{s,t} = b_t - \frac{\beta_t}{\beta_s} B_t B_s^{-1} b_s;\ \beta_{s,t} = \frac{\beta_t}{\beta_s}
\end{equation}
Let $P_{t, \delta_t}^{j_t}$ denote camera $P_{t, \delta}^{j_t}$ in system $\delta_t$, and $(A_{s, t, \delta_t}^o, a_{s, t, \delta_t}^o)$ denote the motion $(A_{s, t, \delta}^o, a_{s, t, \delta}^o)$ of object $o$ in coordinate system $\delta_t$.

\begin{figure}
\begin{center}
  \begin{subfigure}[b]{0.9\linewidth}
    \includegraphics[width=\linewidth]{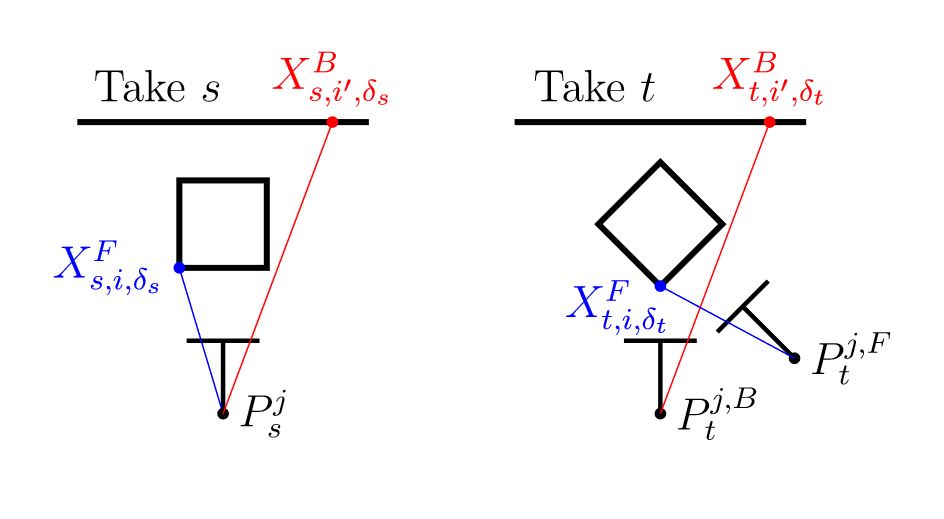}
    \caption{}
    \label{fig:twoposes}
  \end{subfigure}
  \begin{subfigure}[b]{\linewidth}
    \begin{tabular}{c c c}
    \includegraphics[width=0.3\linewidth]{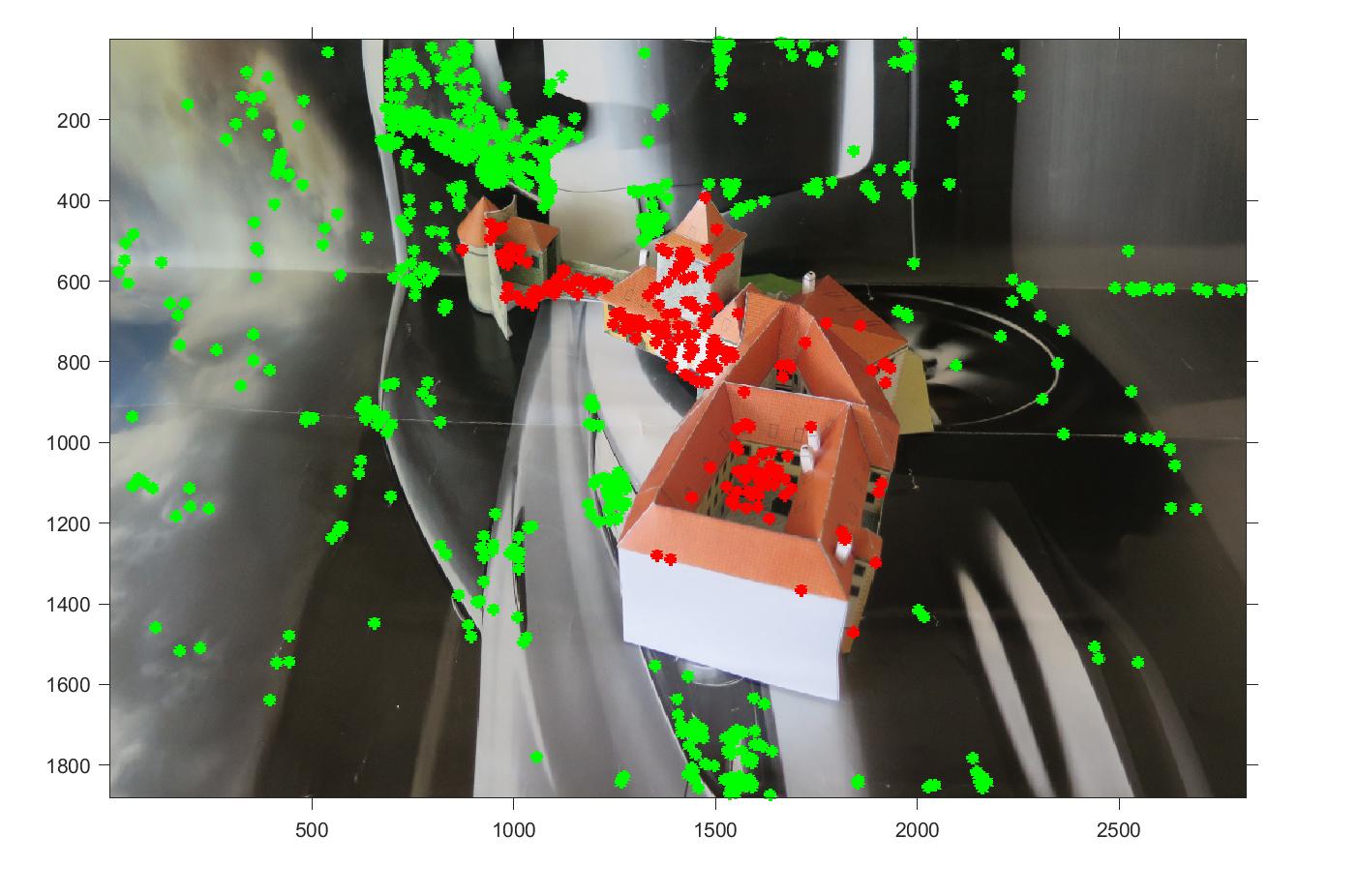}
    \includegraphics[width=0.3\linewidth]{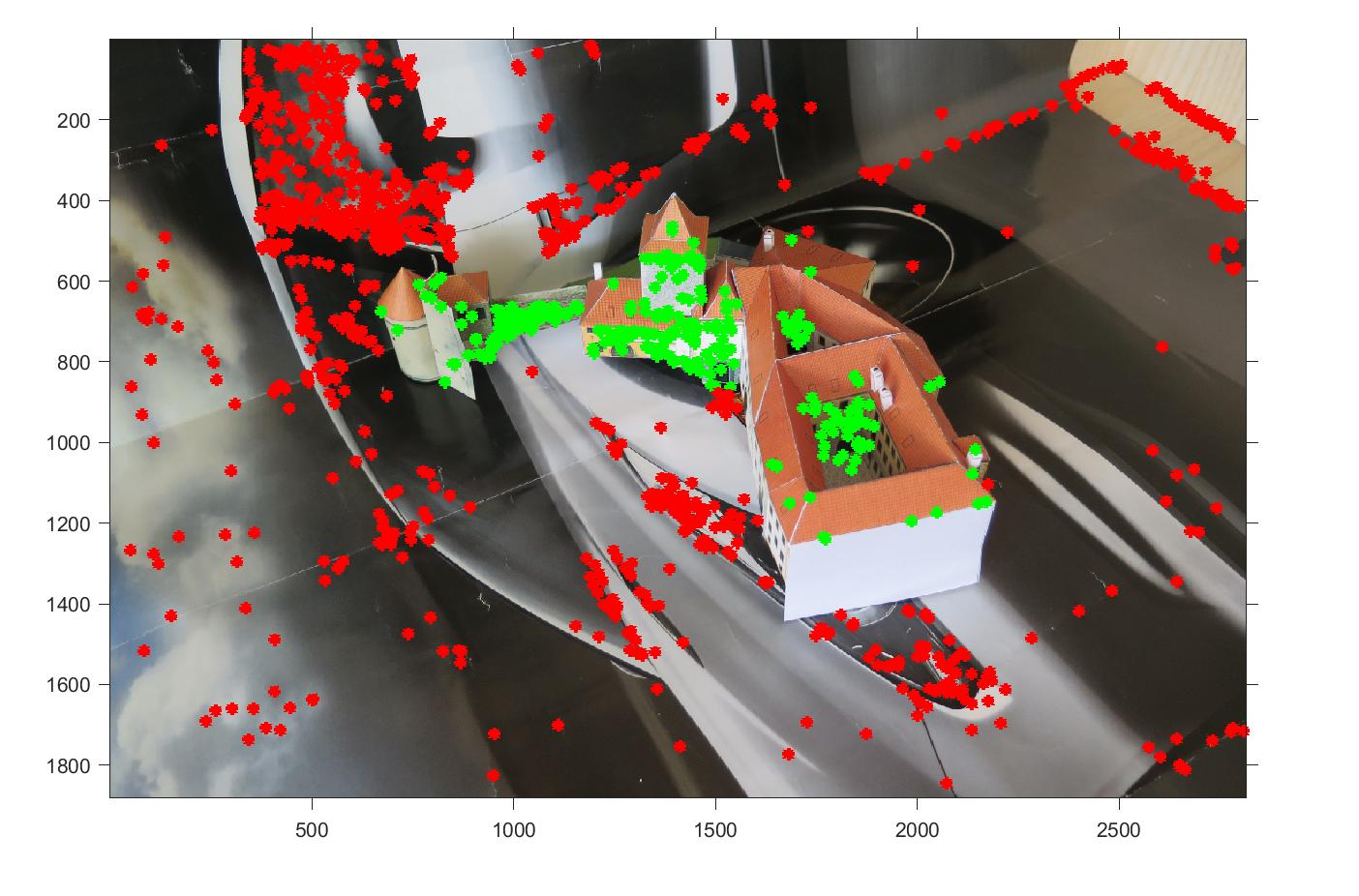}
    \includegraphics[width=0.3\linewidth]{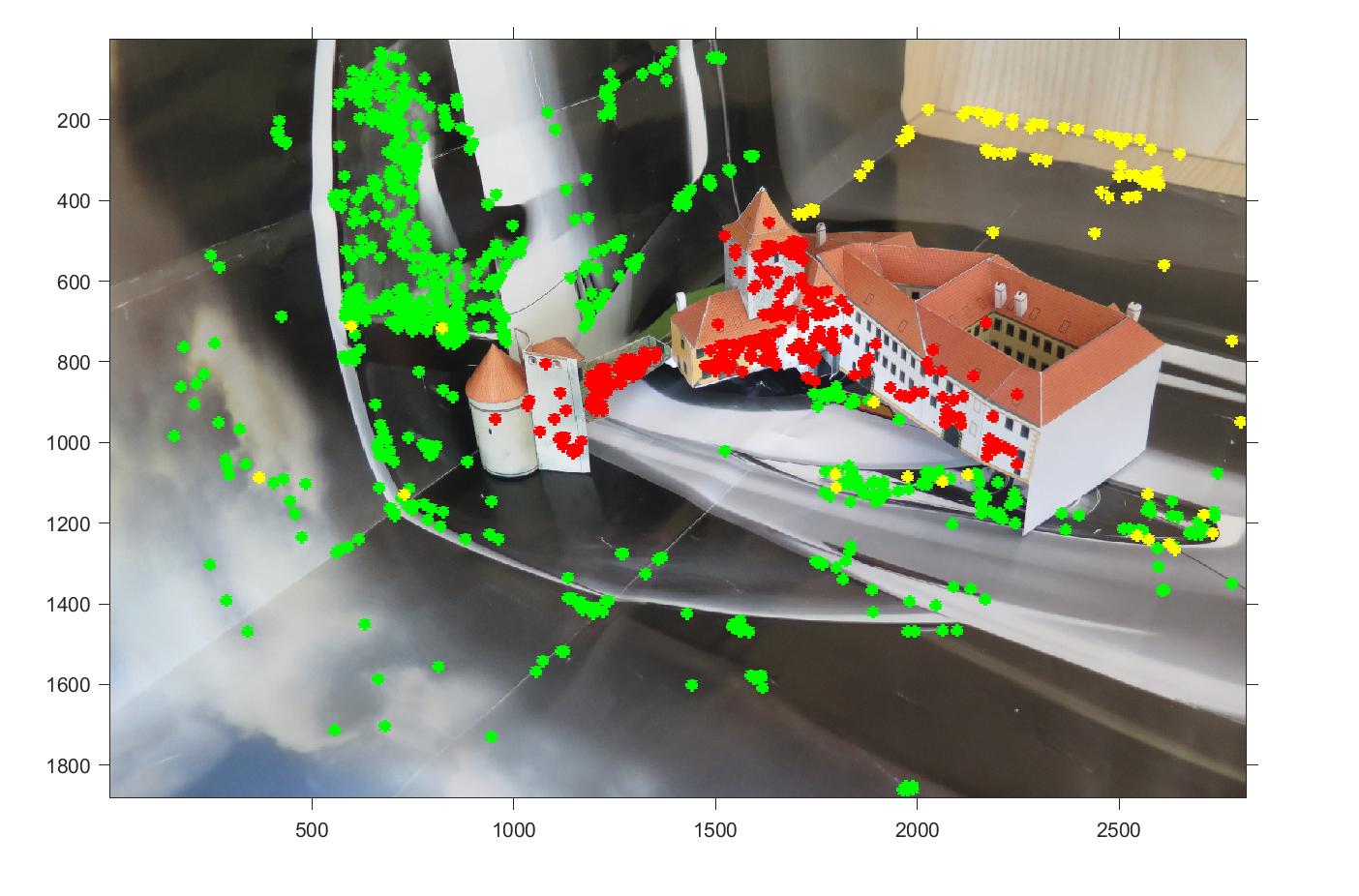}\\
    \end{tabular}
    \caption{}
    \label{fig:seqran}
  \end{subfigure}
\caption{(a) Illustration of two poses of a camera from take $s$ towards take $t$. The figure shows that a camera has two different poses towards a different take; one towards the background and one towards the foreground. (b) Different results of the sequential PnP. Three cameras from take $s$ which were registered towards take $t$. Inliers to the first pose are green, inliers to the second pose are red, inliers to the third pose are yellow.}
\end{center}
\end{figure}

\subsection{Registration of cameras from other takes}\label{sec:cam-registration}
\noindent The next task is to find the segmentation of the points and to combine the $k$ models into one. We will first show how the poses of the cameras from other takes can be used to label the points.

The key observation is that camera $P_{s}^{j}$ from take $s$ has two different poses towards take $t \neq s$; one towards the background and one towards the foreground, as shown in Fig.~\ref{fig:twoposes}. Let us denote by $P_{t}^{j, B}$ the pose towards the background and by $P_{t}^{j, F}$ the pose towards the foreground.

Then, the projection of point $X_{t, i, \delta_t}^F$ from the foreground onto camera $P_{t}^{j, F}$ is the same as the projection of point $X_{s, i, \delta_s}^F$ onto camera $P_{s}^{j}$. So, the reprojection error of points from foreground on camera $P_{t}^{j, F}$ is small. This holds analogously for points from background and camera $P_{t}^{j, B}$. Therefore, if the poses are known, the inliers to pose $P_{t}^{j, B}$ can be assigned to the background, while the inliers to pose $P_{t}^{j_s, F}$ can be assigned to the foreground.

The poses of camera $P_{s}^{j}$ towards take $t$ are obtained with Sequential RANSAC~\cite{Vincent01}. It extracts minimal random samples out of the correspondences and computes poses with PnP~\cite{DBLP:journals/jmiv/WuH06}. The pose maximizing the number of inliers is accepted and the subsequent model fitting is performed on the remaining data. The output is a set of poses $\mathcal{P}_{t}^{j}$. If the registration is not successful, the set is empty.

We use Sequential RANSAC to register poses of every camera towards every take. Fig.~\ref{fig:seqran} shows that Sequential RANSAC does not guarantee the order of the objects. In addition, a camera may be registered multiple times towards the same object. Thus, we aim at grouping the observations of the found poses such that one group will contain the observations of the background and the other group will contain the observations of the foreground.
\subsection{Local observation grouping}\label{sec:loc-group}
\noindent We proceed from the sets $\mathcal{P}_{t}^{j}$ of camera poses found in the sequential registration of the cameras. The goal is to find two groups of the points: one from the background and the other one from the foreground. We use a two step-scheme to group the camera poses:

\noindent{\em  1) For every take $t$ cluster the observations of the poses of cameras registered towards take $t$.}

\noindent{\em 2) Group the clusters found in \textit{1)} together.}

Now, we will show how the observations of the cameras sequentially registered towards the same take $t$ can be clustered. For given take $t$, we select all sets of poses $\mathcal{P}_{t}^{j}, s \neq t$ obtained with the sequential registration of camera $P_{t}^{j}$ observing take $t$. From each of these sets, we extract camera pairs $(P_{t,1}^{j}, P_{t,2}^{j})$ and we cluster these pairs using a linkage procedure similar to \cite{Magri14, Toldo08}:

Let us have two pairs of cameras, $(P_{t,1}^{i}, P_{t,2}^{i})$ and $(P_{t,1}^{j}, P_{t,2}^{j})$. Let $\mathcal{O}_{t,\gamma}^{i}$ denote a set of points, which are observed by a camera $P_{t,\gamma}^{i}$, where $\gamma \in \{1, 2\}$. If the first camera $P_{t,1}^{i}$ in the first pair observes the same object as the first camera $P_{t,1}^{j}$ in the second pair and, at the same time, the second camera $P_{t,2}^{i}$ in the first pair observes the same object as the second camera $P_{t,2}^{j}$ in the second pair, we can expect the intersections $\mathcal{O}_{t,1}^{i} \cap \mathcal{O}_{t,1}^{j}$, $\mathcal{O}_{t,2}^{i} \cap \mathcal{O}_{t,2}^{j}$ to be large and the intersections $\mathcal{O}_{t,1}^{i} \cap \mathcal{O}_{t,2}^{j}$, $\mathcal{O}_{t,2}^{i} \cap \mathcal{O}_{t,1}^{j}$ to be small or empty. If, on the other hand, the first camera in the first pair observes the same object as the second camera in the second pair and, at the same time, the second camera in the first pair observes the same object as the first camera in the second pair, we can expect the intersections $\mathcal{O}_{t,1}^{i} \cap \mathcal{O}_{t,2}^{j}$, $\mathcal{O}_{t,2}^{i} \cap \mathcal{O}_{t,1}^{j}$ to be large and the intersections $\mathcal{O}_{t,1}^{i} \cap \mathcal{O}_{t,1}^{j}$, $\mathcal{O}_{t,2}^{i} \cap \mathcal{O}_{t,2}^{j}$ to be empty. Therefore, we propose the following linkage scheme\footnote{An additional optional criterion based on the motion of the object, which is suitable for takes containing many objects, is described in the Supplementary material. }:
\begin{multline}
    (i^*,j^*) = argmax_{i,j} (\lvert \mathcal{O}_{t,1}^i \cap \mathcal{O}_{t,1}^j \rvert + \lvert \mathcal{O}_{t,2}^i \cap \mathcal{O}_{t,2}^j \rvert) \\
    s.t. \lvert \mathcal{O}_{t,1}^i \cap \mathcal{O}_{t,2}^j \rvert < \theta_{1,2}, \lvert \mathcal{O}_{t,2}^i \cap \mathcal{O}_{t,1}^j\rvert < \theta_{2,1} \label{eq:linkage1}
\end{multline}
\begin{multline}
    (i^{**},j^{**}) = argmax_{i,j} (\lvert \mathcal{O}_{t,1}^i \cap \mathcal{O}_{t,2}^j \rvert + \lvert \mathcal{O}_{t,2}^i \cap \mathcal{O}_{t,1}^j \rvert) \\
    s.t. \lvert \mathcal{O}_{t,1}^i \cap \mathcal{O}_{t,1}^j \rvert < \theta_{1,1}, \lvert \mathcal{O}_{t,2}^i \cap \mathcal{O}_{t,2}^j\rvert < \theta_{2,2} \label{eq:linkage2}
\end{multline}
In every step, we select the pair out of $(i^*,j^*)$, $(i^{**},j^{**})$, which produces the larger union and we merge the corresponding observations.
Namely, if $\lvert \mathcal{O}_{t,1}^{i^*} \cap \mathcal{O}_{t,1}^{j^*} \rvert + \lvert \mathcal{O}_{t,2}^{i^*} \cap \mathcal{O}_{t,2}^{j^*} \rvert >= \lvert \mathcal{O}_{t,1}^{i^{**}} \cap \mathcal{O}_{t,2}^{j^{**}} \rvert + \lvert \mathcal{O}_{t,2}^{i^{**}} \cap \mathcal{O}_{t,1}^{j^{**}} \rvert$, then ${O}_{t,1}^{i^*} := {O}_{t,1}^{i^*} \cup {O}_{t,1}^{j^*}$; ${O}_{t,2}^{i^*} := {O}_{t,2}^{i^*} \cup {O}_{t,2}^{j^*}$. Otherwise $\mathcal{O}_{t,1}^{i^{**}} := \mathcal{O}_{t,1}^{i^{**}} \cup \mathcal{O}_{t,2}^{j^{**}}$; $\mathcal{O}_{t,2}^{i^{**}} := \mathcal{O}_{t,2}^{i^{**}} \cap \mathcal{O}_{t,1}^{j^{**}}$. If there is no pair, that can be merged, we terminate. Threshold $\theta_{\rho \sigma}$ is selected empirically as $2\%$ of $\lvert \mathcal{O}_{t,\rho}^i \cap \mathcal{O}_{t,\sigma}^j \rvert$, where $\rho, \sigma \in \{ 1, 2\}$. 

During the linkage, we remember from which cameras the resulting sets of observed points arise. In the subsequent steps of the algorithm, we use the pair of sets of points, which has been merged from the highest number of cameras. The rationale behind this is that, although some of the cameras might have been registered wrongly, we might expect large consensus on the correct sets, while the individual incorrect results differ from each other. We denote the selected pair of sets of points $(\mathcal{G}_t^1, \mathcal{G}_t^2)$ We may expect that one set from this pair contains points from the background, while the other set contains points from the foreground.

\subsection{Track building}\label{sec:track-build}

To group the clusters, we use the fact, that if two cameras observe the same points, they observe the same object. However, after reconstructing models, we do not know which points from different models correspond to the same real points. Therefore, we need to find the corresponding points before we group the clusters.

We build a graph, where the points are the vertices. We connect two points $X_{t, i, \delta_t}^{o(i)}$, $X_{s, i, \delta_s}^{o(i)}$ from different models $s, t$ with an edge, if a camera $P_{\delta}^{j}$ with poses $P_{s}^{j}$, $P_{t}^{j}$ exists, such that $P_{s}^{j}$ projects $X_{t, i, \delta_t}^{o(i)}$ onto the same observation onto which $P_{t}^{j}$ projects $X_{s, i, \delta_s}^{o(i)}$.

We call a set of all reconstructed 3D points from different models, which correspond to the same real point, a track. Every track is built in such way that it contains the points that belong to the same connected component in the graph.

\subsection{Point segmentation}\label{sec:segmentation}
Now, we will describe how the points are labeled.

For every take $t$, we have found a pair of sets of points $(\mathcal{G}_{t}^1, \mathcal{G}_{t}^2)$, such that the first set contains the points from the background and the second set contains the points from the foreground, or vice versa. Our task is to merge the points from reconstructions of different takes in such a way, that the first group contains the points from the background and the second one contains the points from the foreground, or vice versa.

First, we replace the points with the tracks, which have been found in Section \ref{sec:track-build}. Then, we use a greedy scheme in order to merge the track sets arising from different takes.

First, we select the reference take $r$ as the take whose pair $(\mathcal{G}_{r}^1, \mathcal{G}_{r}^2)$ which has been merged from the highest number of cameras. In every step, we select one pair $(\mathcal{G}_{t}^1, \mathcal{G}_{t}^2)$ which maximizes the linkage criterion \eqref{eq:linkage1} or \eqref{eq:linkage2} and merge it with $(\mathcal{G}_r^1, \mathcal{G}_r^2)$ in the appropriate order. We stop if all pairs are merged or if no remaining pair can be merged.

We have found a pair of sets of points $(\mathcal{G}_r^1, \mathcal{G}_r^2)$, where one set contains points from the background and the other set contains the points from the foreground. From now on we assume that $\mathcal{G}_r^1$ contains the points from the background.

We define the labelling $o(i) \in [1, n] \rightarrow \{B, F, U\}$ as
follows
\begin{itemize}
    \item{$X_{i, \delta}^{o(i)} \in \mathcal{G}_r^1$} Point $X_{i, \delta}^{o(i)}$ belongs to the background, so $o(i) = B$.
   \item{$X_{i, \delta}^{o(i)} \in \mathcal{G}_r^2$} Point $X_{i, \delta}^{o(i)}$ belongs to the foreground, so $o(i) = F$.
    \item{$X_{i, \delta}^{o(i)} \notin \mathcal{G}_r^1$, $X_{i, \delta}^{o(i)} \notin \mathcal{G}_r^2$} We do not know to which object the point belongs; thus we have $o(i) = U$.
\end{itemize}
Typically, the majority of the points end up being unlabeled after the first step. We, therefore proceed to the second step.

For every unlabeled point $X_{i, \delta}^{o(i)}$ we find $k$ nearest labeled points among all models. If all the nearest points are assigned to the background, we assign the point to the background $o(i) = B$. If all the nearest points are assigned to the foreground, we assign the point to the foreground $o(i) = F$. Otherwise, we leave the point unlabeled.

\subsection{Reconstruction merging}\label{sec:rec-merge}
\noindent We have found an approximate solution to task \eqref{eq:task1}, therefore, we can proceed to the solution to task \eqref{eq:task2}. We aim at merging the models of the takes $[1,k]$ into a single model.


We use the reference take $r$ from Section \ref{sec:segmentation}. The order $t_1, ..., t_k$ in which the models are transformed is the same as the order, in which the pairs were merged in Section \ref{sec:segmentation}.

For every take $t \in [1,k] \setminus \{ r \}$ in the given order, the transformation $(B_{t, r}^o, b_{t, r}^o, \beta_{t, r}^o)$ is found using the least squares according to \cite{Arun87}. All points in model of take $t$ are transformed with the obtained transformation and added to the model of take $r$. 

Next, we transform every camera\footnote{See the description of the camera transformation in the Supplementary Material.} $P_{s}^{j, o}, t(j) = s$ to its pose $P_{r}^{j, B}$ in the reference take towards the background. The formula for transformation of the camera depends on take $s$ from which the camera origins, take $t$ towards which it was registered and object $o$ which it observes.

After all the models have been merged, we perform the bundle adjustment~\cite{Triggs00}, where the function \eqref{eq:task2} is minimized using the gradient descent. To prevent the model from overfitting, we use constraints which require that the motion between the two poses of the same camera is the same for every camera from the same take.
\begin{figure}[t]
\begin{center}
\begin{tabular}{c c c c}
\includegraphics[width=0.20\linewidth]{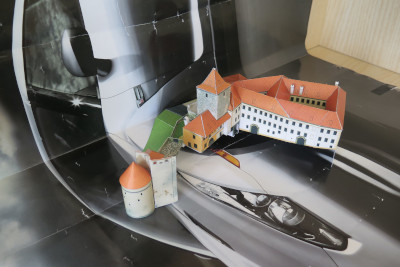}&
\includegraphics[width=0.20\linewidth]{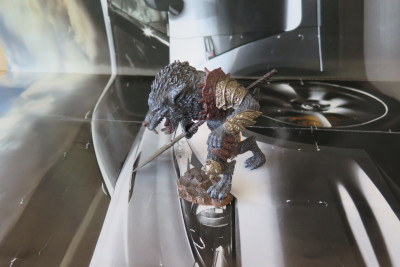}&
\includegraphics[width=0.20\linewidth]{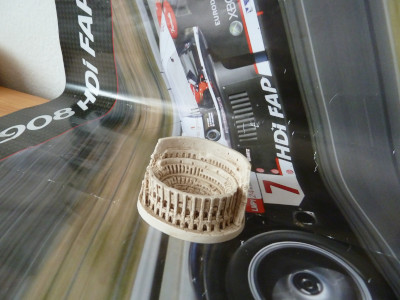}&
\includegraphics[width=0.20\linewidth]{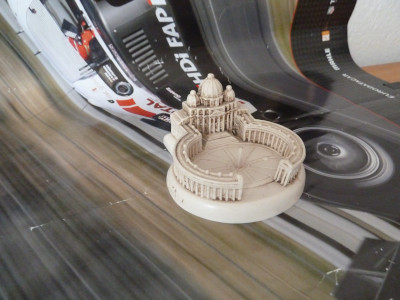}\\
(a) Dal. & (b) Lyc. & (c) Col. & (d) Vat.\\
\includegraphics[width=0.20\linewidth]{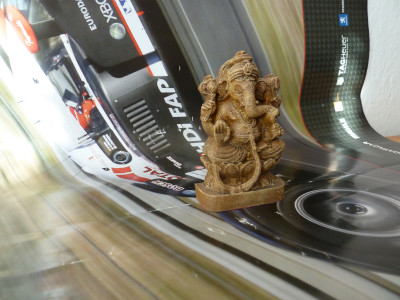}&
\includegraphics[width=0.20\linewidth]{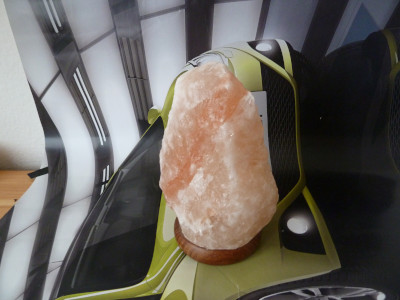}&
\includegraphics[width=0.20\linewidth]{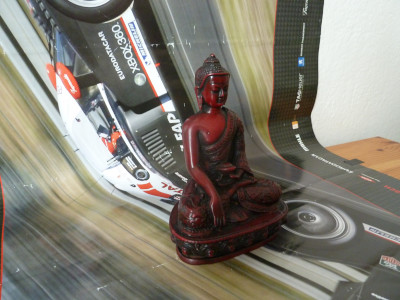}&
\includegraphics[width=0.20\linewidth]{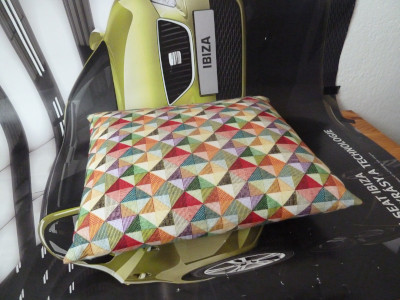}\\
(e) Gan. & (f) Lamp & (g) Bud. & (h) Pil. \\
\includegraphics[width=0.20\linewidth]{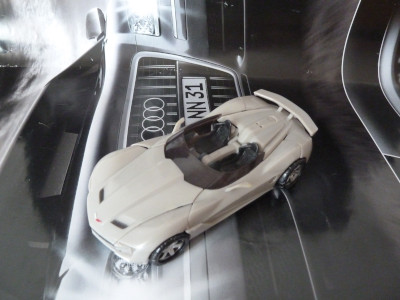}&
\includegraphics[width=0.20\linewidth]{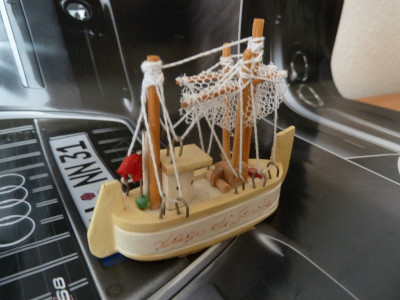}&
\includegraphics[width=0.20\linewidth]{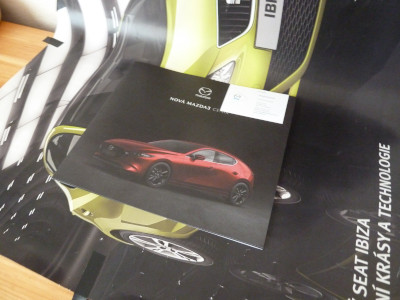}&
\includegraphics[width=0.20\linewidth]{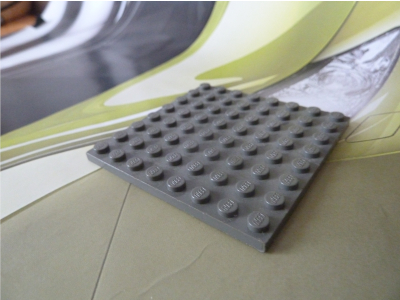}\\
(i) Car & (j) Ship & (k) Cat. & (l) Lego\\
\end{tabular}
\begin{tabular}{|l|c|c|c|c|c|c|c|c|c|c|c|c|}
\hline
Dataset & (a) & (b) & (c) & (d) & (e) & (f) \\
\hline\hline
\#Takes & 8 & 8 & 8 & 8 & 4 & 8 \\
\#Imgs w. bg. & 338 & 140 & 294 & 270 & 160 & 330 \\ 
\#Imgs w/o bg. & - & - & 326 & 304 & 187 & 302 \\
\hline\hline
Dataset & (g) & (h) & (i) & (j) & (k) & (l) \\
\hline\hline
\#Takes & 8 & 8 & 8 & 8 & 5 & 4 \\
\#Imgs w. bg. & 300 & 278 & 329 & 277 & 150 & 264\\
\#Imgs w/o bg. & - & 275 & 290 & 294 & 151 & 307\\

\hline
\end{tabular}
\end{center}
\caption{Sample images from our dataset. For every object, the table shows numbers of takes and images  as well as the number of images without background, which are used for the comparison with COLMAP.}
\label{fig:dataset}
\end{figure}
\section{Experiments}
\noindent The problem described in this paper has not exactly been considered before, and no datasets existed just for it. Thus first, to demonstrate that our approach delivers good and interesting results, we have created a new dataset, and we evaluate the performance of our method. Our motivation in this experiment is to reconstruct small 3D artifacts and the dataset corresponds to this setting. We considered 12 different scenes, each of which consists of a foreground object and a textured background. We acquired $150$ to $330$ images of every scene; each one was depicted in $4$~-~$8$ different configurations (takes), Fig.~\ref{fig:dataset}.

\subsection{Qualitative results}
\begin{figure}[t]
\begin{center}
\begin{tabular}{c c c c c c}
\includegraphics[width=0.20\linewidth]{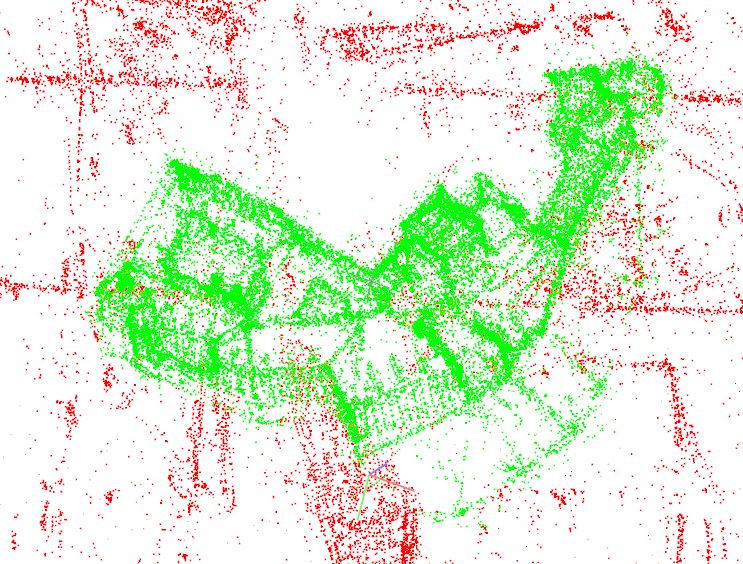}&
\includegraphics[width=0.20\linewidth]{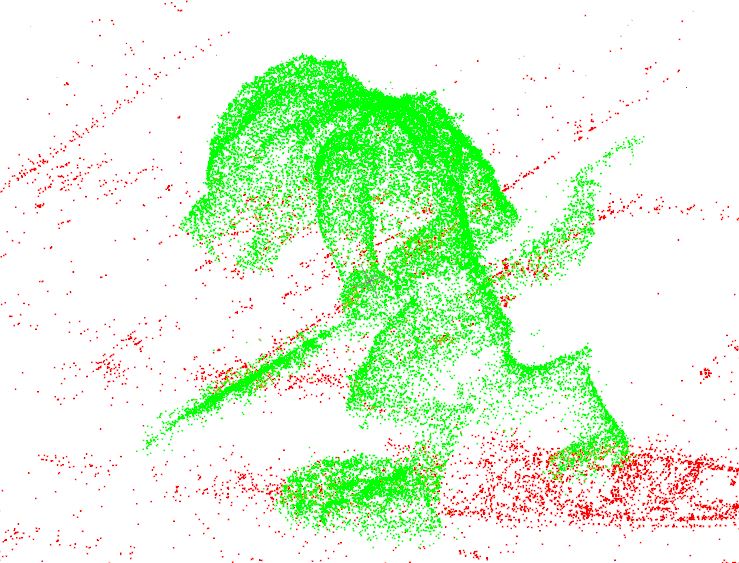}&
\includegraphics[width=0.20\linewidth]{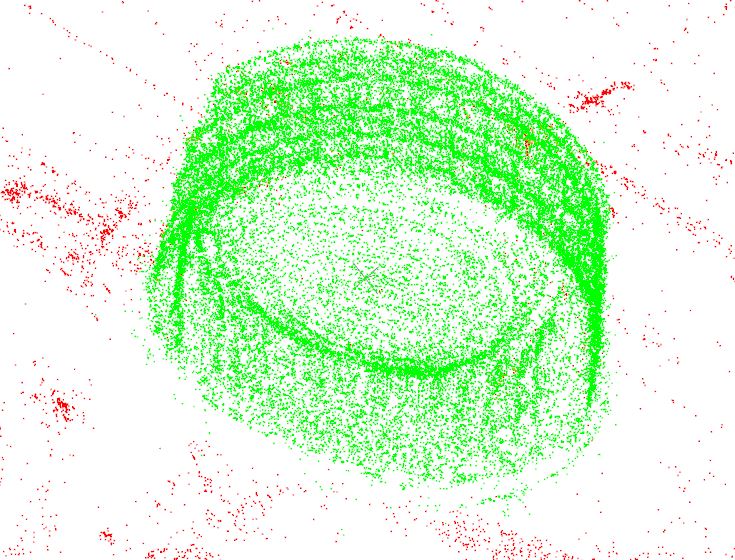}&
\includegraphics[width=0.20\linewidth]{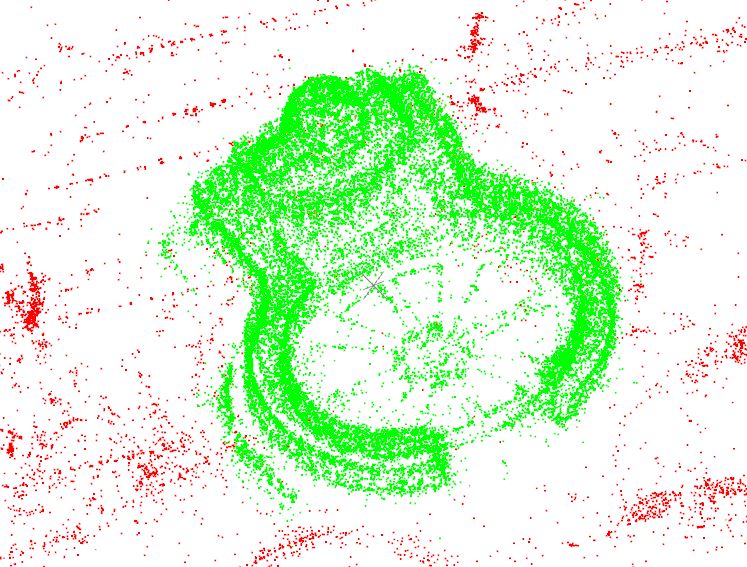}\\
(a) Dal. & (b) Lyc. & (c) Col. & (d) Vat.\\
\includegraphics[width=0.20\linewidth]{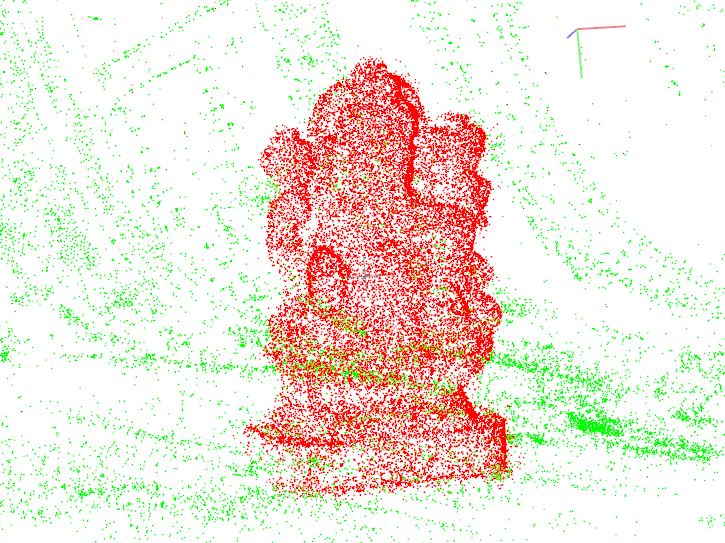}&
\includegraphics[width=0.20\linewidth]{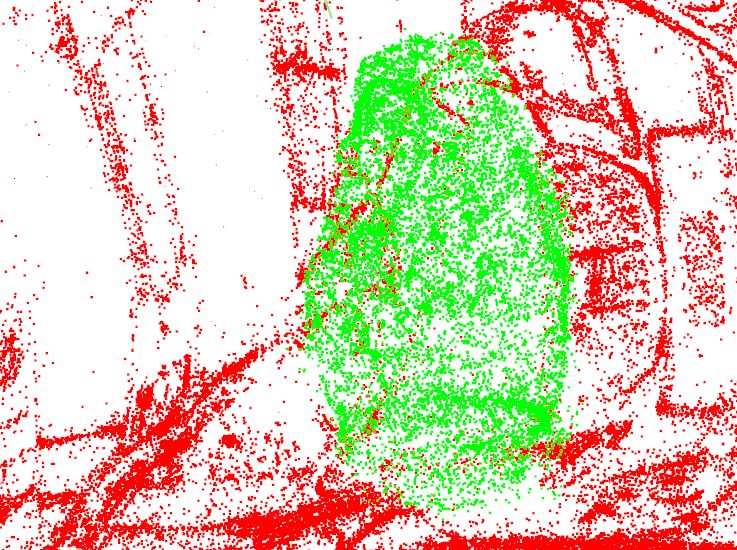}&
&
\includegraphics[width=0.20\linewidth]{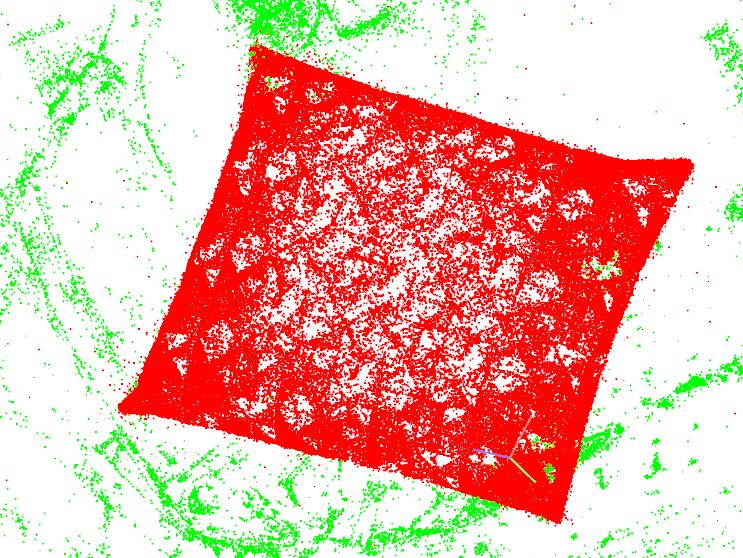}\\
(e) Gan. & (f) Lamp & (g) Bud. & (h) Pil. \\
\\
\includegraphics[width=0.20\linewidth]{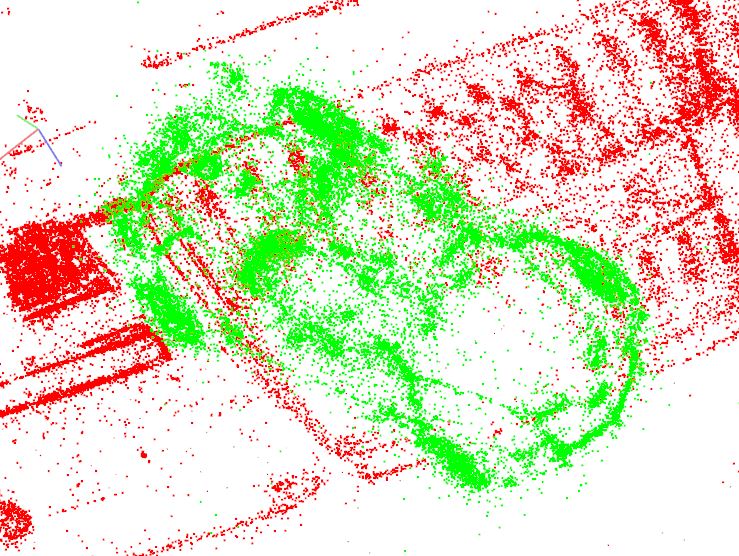}&
\includegraphics[width=0.20\linewidth]{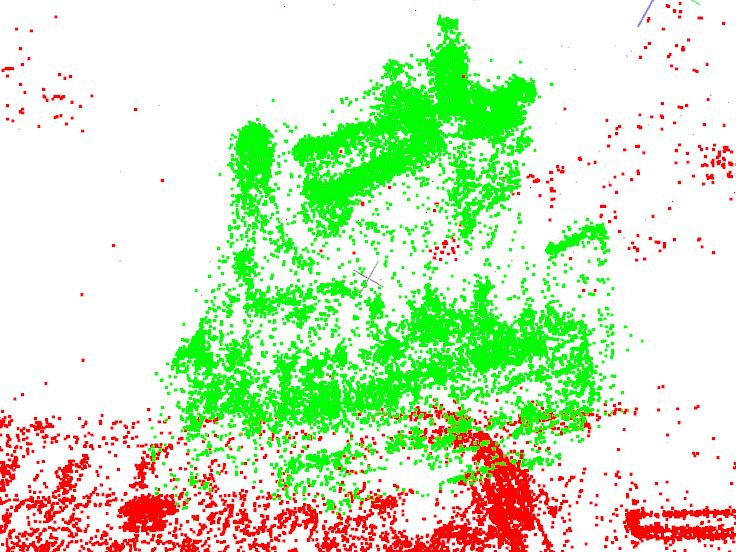}&
\includegraphics[width=0.20\linewidth]{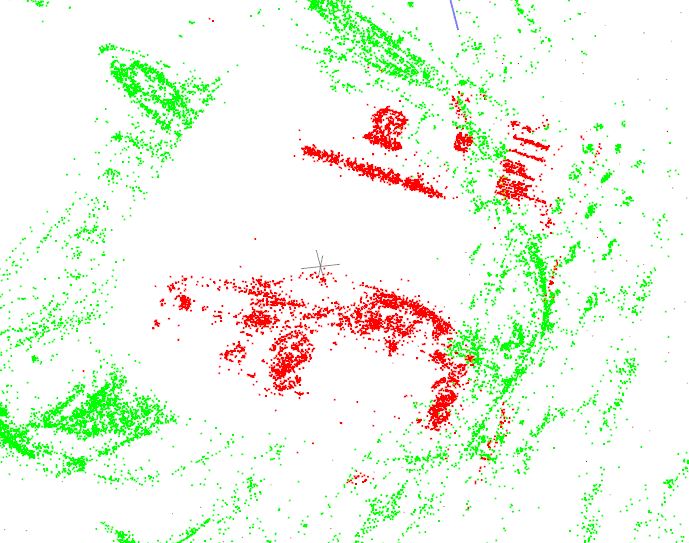}&
\includegraphics[width=0.20\linewidth]{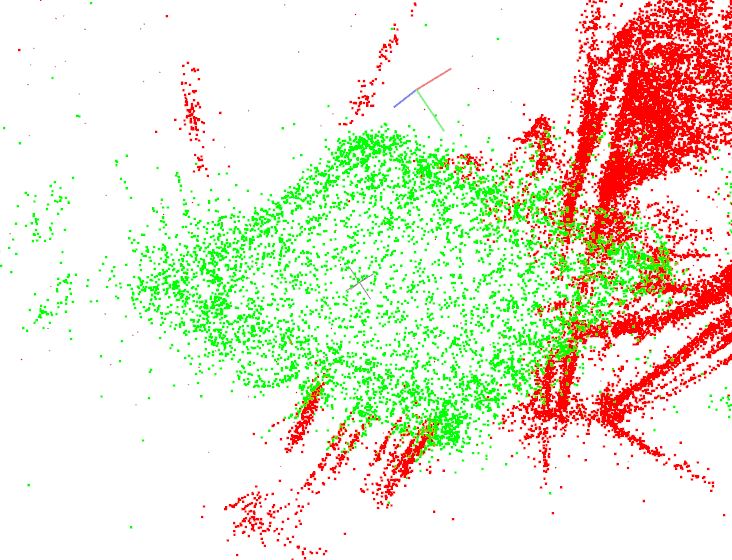}\\
(i) Car & (j) Ship & (k) Cat. & (l) Lego\\
\end{tabular}
\end{center}
\caption{Results of our metx    hod. Foreground is green, background is red.}
\label{fig:results}
\end{figure}

\noindent  Fig.~\ref{fig:results} shows the results of our method on our dataset. The hyper-parameters were tuned mostly using datasets Daliborka, Lycan, Salt Lamp and Colosseum. The other models were reconstructed with the same settings as the previously mentioned ones. The method did not manage to reconstruct sequence "Buddha". We assume, Sec.~\ref{sec:segmentation},  that the first cameras observe the background and the second cameras observe the foreground. This assumption was not met in datasets (e), (h), (k). Hence, the background and the foreground are swapped.

\subsection{Comparison with State-of-the-Art}\label{sec:exp1}
\noindent A direct competitors to our method is hard to do, as this problem had not been really addressed before, but we compare our approach to the most relevant previous approaches. 

It makes sense to compare our method to a general approach to motion segmentation and 3D reconstruction. However, as we use unordered input images, we omit the comparison with the methods assuming video input. Also, in our setup, the points are observed by a small portion of cameras, thus the tracks are very sparse, and the track completion is an under-constrained task. The methods based on factorization, therefore, fail, as shows \cite{DBLP:journals/tit/CandesT10}, as well as the results of ALC \cite{DBLP:conf/cvpr/RaoTVM08}, SSC \cite{DBLP:journals/pami/ElhamifarV13} and RSIM \cite{DBLP:conf/iccv/JiSL15}.

For the motion segmentation, we used six state-of-the-art algorithms: SSC \cite{DBLP:journals/pami/ElhamifarV13}, RSIM \cite{DBLP:conf/iccv/JiSL15}, ALC \cite{DBLP:conf/cvpr/RaoTVM08}, Subset \cite{Xu18}, MODE \cite{DBLP:journals/corr/abs-1905-09043} and YASFM \cite{Srajer16}. However, none of the considered methods has produced a satisfactory motion segmentation of sequences from our dataset. SSC outputs wrong results, where almost all points are assigned to one motion. RSIM reports badly conditioned matrices, ALC reports unbounded system during track completion. Subset and MODE run out of memory. If we make the input data shorter, up to 20 images, both Subset and MODE give an incorrect segmentation. YASFM splits the points into more than two models, most of which contain points from both the foreground and the background.\footnote{The results of Subset, MODE and YASFM are shown in the Supplementary Material.}
\subsection{Comparison with Single body SfM}\label{sec:exp2}
\noindent The motivation for our paper was to find out whether the presence of a background can improve the quality of a model of a given object. To answer this question, we reconstructed the same objects in a classical SfM pipeline COLMAP\cite{schoenberger2016sfm}\footnote{The models reconstructed with TBSfM and with COLMAP are shown in the Supplementary Material.}, which is considered to be the state-of-the-art method. COLMAP cannot work with multiple objects, therefore, we run COLMAP on a different dataset\footnote{Example of images from such dataset is in Supplementary Material.} with a similar number of images, where the object is depicted without background. The number of the images for the datasets with and without background is in table in Fig.~\ref{fig:dataset}.

We compared two metrics: (1) the number of reconstructed points and (2) the median of the reprojection error, Tab.~\ref{tab:error}. In the case of our method, we count only the points on the foreground.

In the most cases, our method reconstructed models with more points than the standard method. In addition to that, our method achieved lower median reprojection errors for every object. Note, that our method reconstructed the "Lego" model, which COLMAP could not reconstruct. We can see that the camera poses can be better estimated from a large object with many features when compared to a small object with few features. Thus, the presence of the background in the scene may improve the reconstruction.

\subsubsection{Computation time}
To compare computation time of TBSfM with the standard SfM, we evaluate the computation times for dataset (e). We get: 2.3 mins feature extraction, 1065 mins feature matching, 14.3 mins 3D reconstruction. Original COLMAP for the same dataset takes 2.3 mins extraction, 449 mins matching, 5.9 mins 3D reconstruction. Our method takes 2.37 times more time in feature matching and 2.4 times more in 3D reconstruction. We solve for two objects, instead of one. It roughly doubles the computation times. Further speedups would be possible if the pipeline was not build on COLMAP but directly designed for multiple objects.

\begin{table}[t]
\begin{center}
\begin{tabular}{|c|c|c||c|c|}
\hline
Dataset & \multicolumn{2}{c}{Points} \vline & \multicolumn{2}{c}{Error} \vline\\
\hline
& TBSfM & COLMAP & TBSfM & COLMAP \\
\hline
(c) & \textbf{52397} & 49418 & \textbf{1.206} & 1.483\\
\hline
(d) & \textbf{39024} & 38833 & \textbf{1.418} & 1.792\\
\hline
(e) & \textbf{35179} & 34539 & \textbf{1.142} & 1.852\\
\hline
(f) & \textbf{15305} & 10049 & \textbf{1.298} & 1.889\\
\hline
(h) & 134086 & \textbf{198728} & \textbf{1.147} & 2.342\\
\hline
(i) & \textbf{25229} & 14372 & \textbf{1.463} & 2.644\\
\hline
(j) & 18244 & \textbf{20571} & \textbf{1.326} & 1.739\\
\hline
(k) & 8004 & \textbf{10143} & \textbf{2.350} & 2.832\\
\hline
(l) & \textbf{7692} & 0 & \textbf{1.818} & -\\
\hline
\end{tabular}
\end{center}
\caption{Comparison of the number of points and the median reprojection error between our method and COLMAP.}
\label{tab:error}
\end{table}

\subsection{Experiments with ETH 3D dataset}

To further evaluate our method, we have chosen sequences motion\_1, motion\_2, motion\_3 and motion\_4 from the SLAM benchmark of the ETH 3D dataset\footnote{https://www.eth3d.net/slam\_datasets}. According to the website of the dataset, sequences motion\_3 and motion\_4 remain unsolved. These four sequences depict a dynamic scene captured by a rig of two synchronized cameras. Although the objects in the scene perform an independent motion, the objects do not move between any two images taken at the same time by the two cameras of the rig. Therefore, each pair of images taken at the same time builds one take. If we limit ourselves only to such parts of sequence where one object is moving, we may be able to reconstruct the parts of sequences using our method.

We have extracted parts of sequences motion\_1, motion\_2, motion\_3 and motion\_4 where there is always one moving object. When compared to our dataset, the images from the ETH 3D dataset have less textured background, worse quality of images, more takes, each of which consists only of two images. In addition to that, there is a person moving the objects, that may confuse the segmentation.

\begin{figure}[t]
\begin{center}
\begin{tabular}{c c c c}
\includegraphics[width=0.2\linewidth]{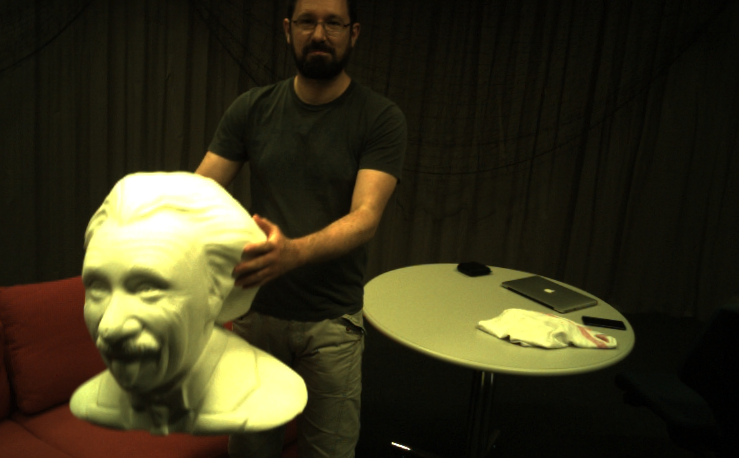}&
\includegraphics[width=0.23\linewidth]{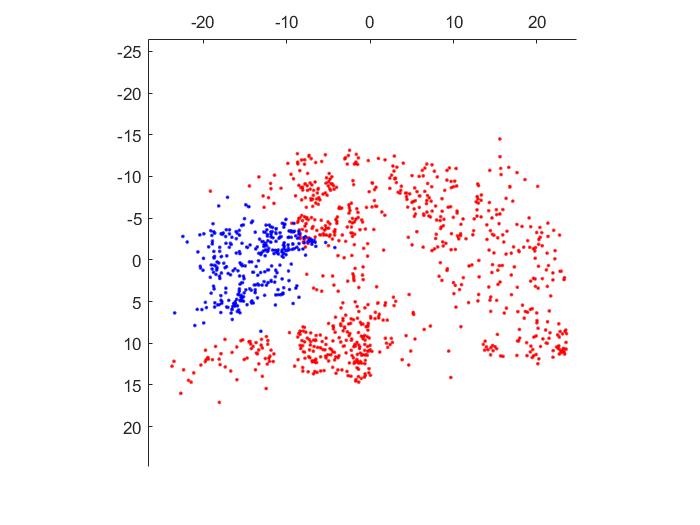}&
\includegraphics[width=0.2\linewidth]{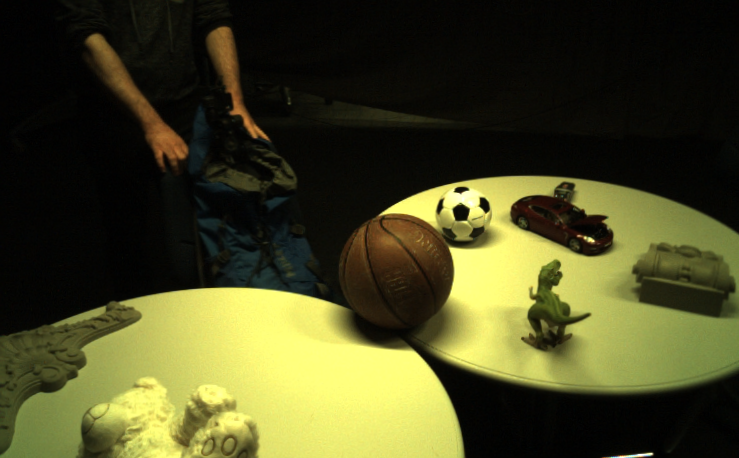}&
\includegraphics[width=0.23\linewidth]{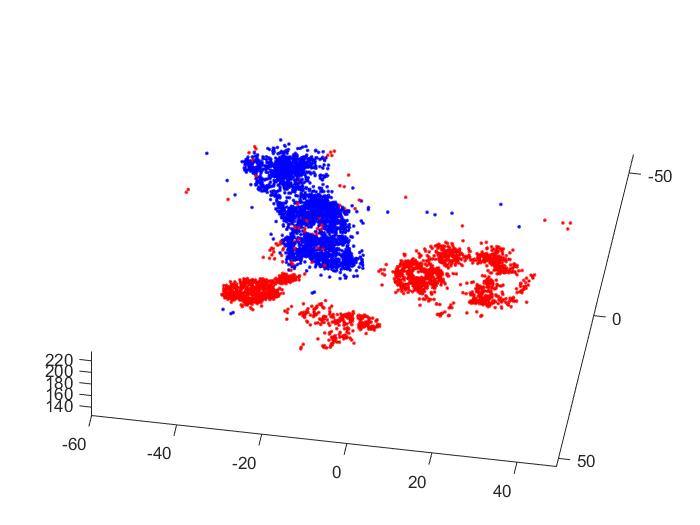}\\
\multicolumn{2}{c}{(a)} & \multicolumn{2}{c}{(b)} \\

\includegraphics[width=0.2\linewidth]{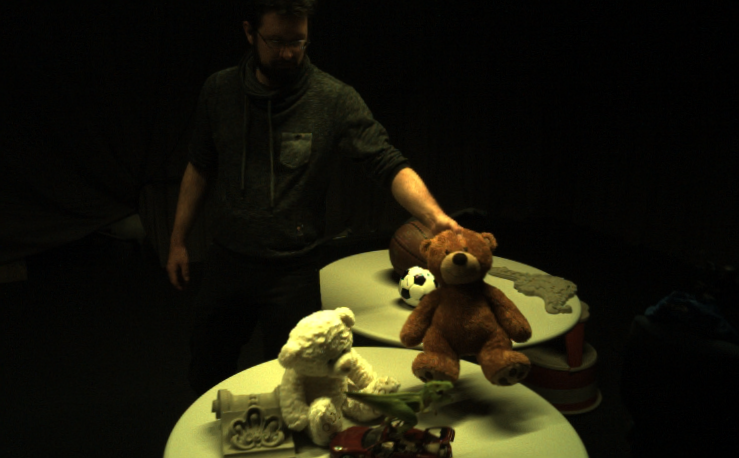}&
\includegraphics[width=0.23\linewidth]{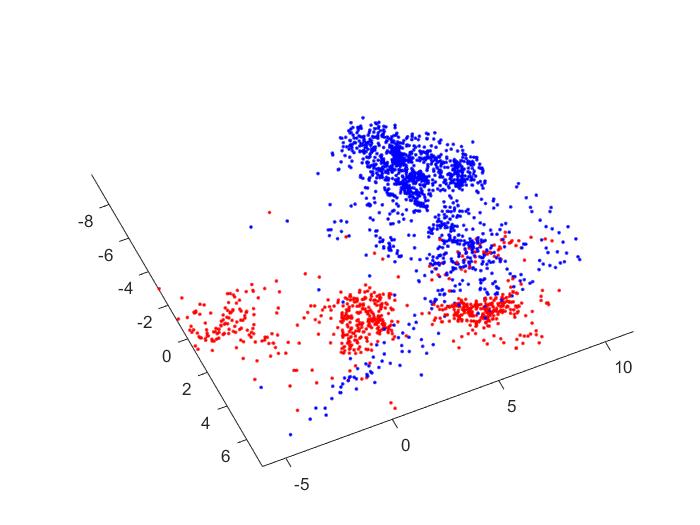}&
\includegraphics[width=0.2\linewidth]{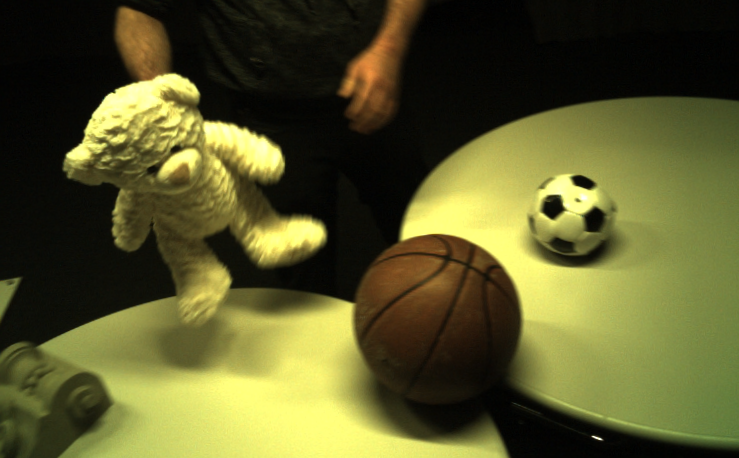}&
\includegraphics[width=0.23\linewidth]{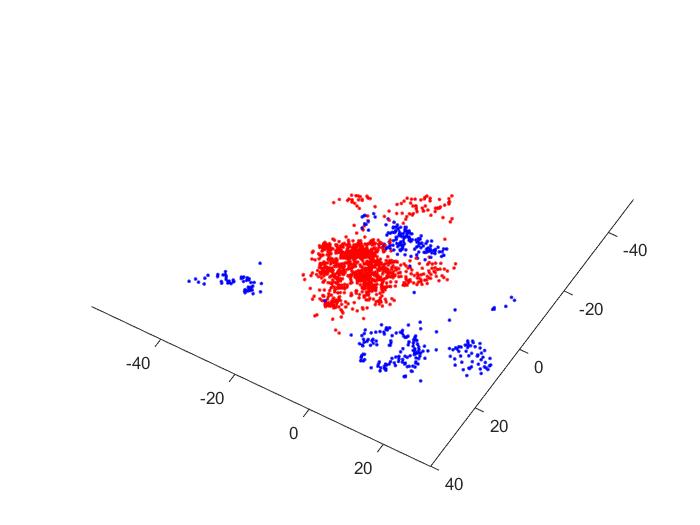}\\
\multicolumn{2}{c}{(c)} & \multicolumn{2}{c}{(d)} \\

\includegraphics[width=0.2\linewidth]{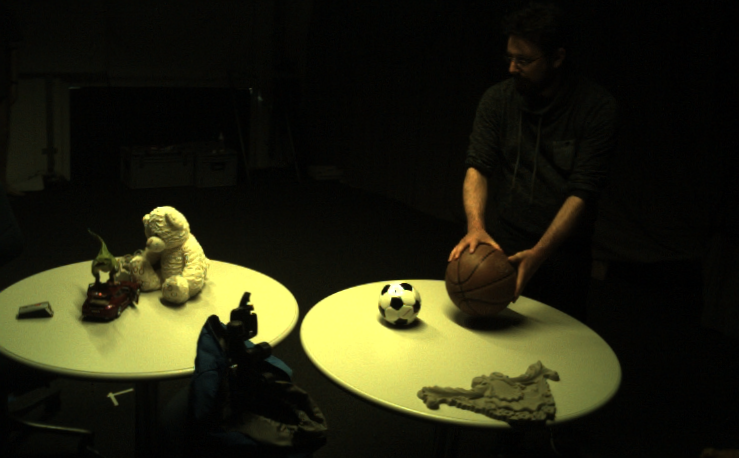}&
\includegraphics[width=0.23\linewidth]{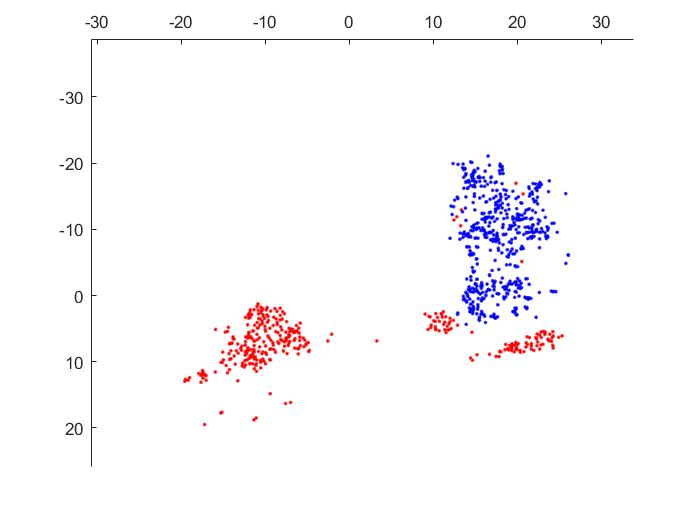}&
\includegraphics[width=0.2\linewidth]{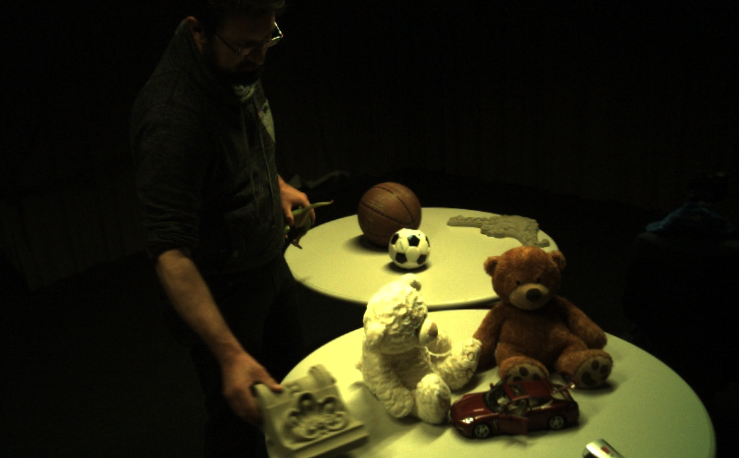}&
\includegraphics[width=0.23\linewidth]{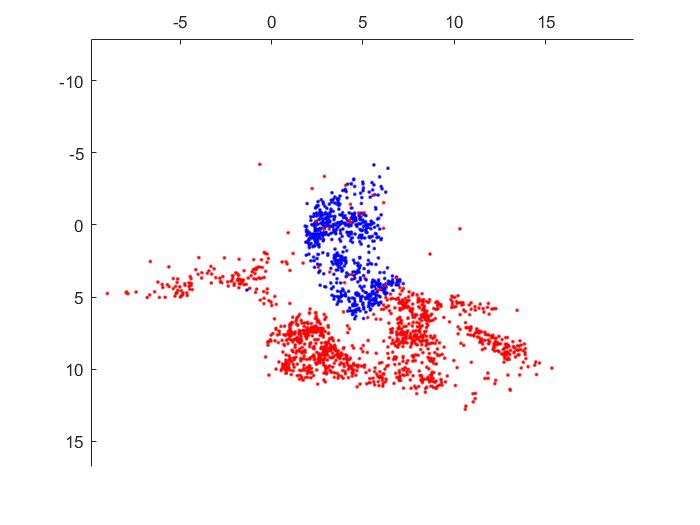}\\
\multicolumn{2}{c}{(e)} & \multicolumn{2}{c}{(f)} \\

\includegraphics[width=0.2\linewidth]{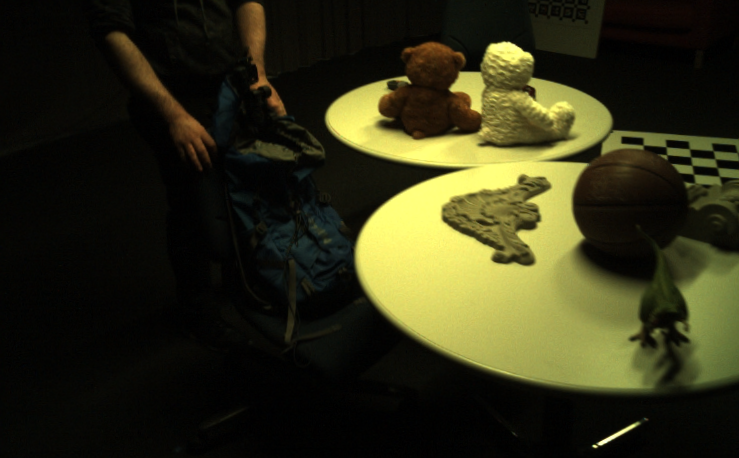}&
\includegraphics[width=0.23\linewidth]{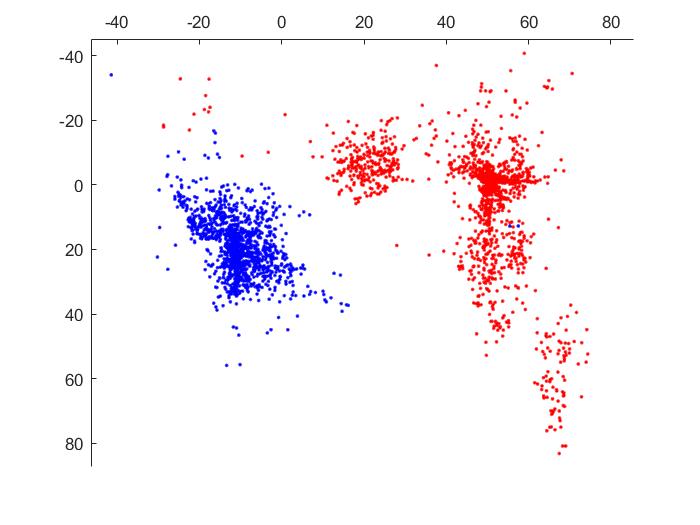}&
\includegraphics[width=0.2\linewidth]{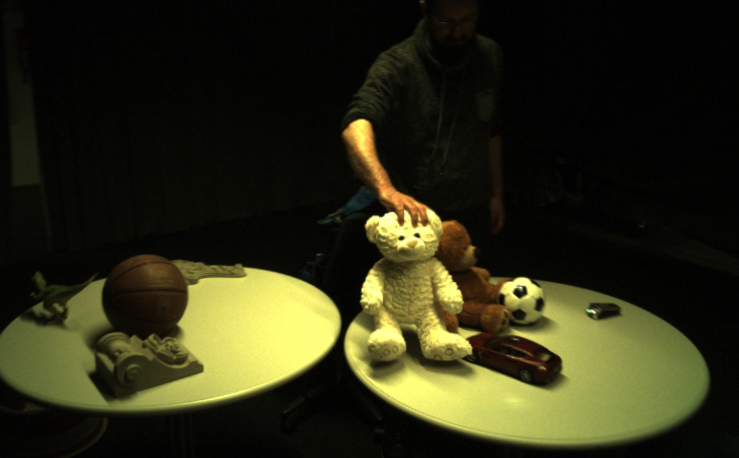}&
\includegraphics[width=0.23\linewidth]{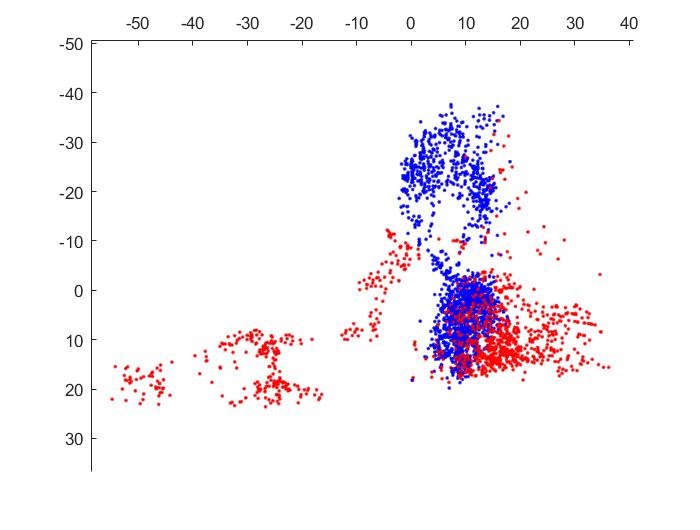}\\
\multicolumn{2}{c}{(g)} & \multicolumn{2}{c}{(h)} \\

\includegraphics[width=0.2\linewidth]{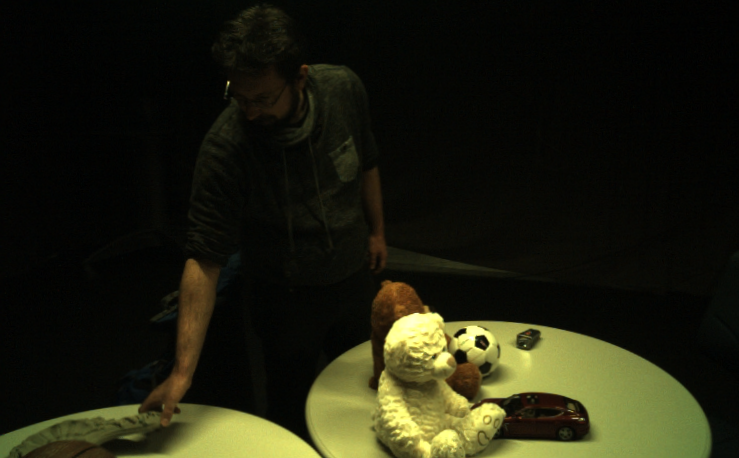}&
\includegraphics[width=0.23\linewidth]{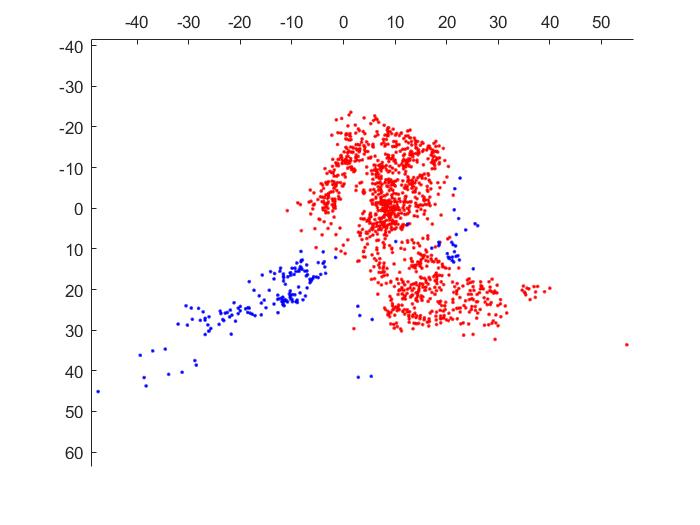}&
\includegraphics[width=0.2\linewidth]{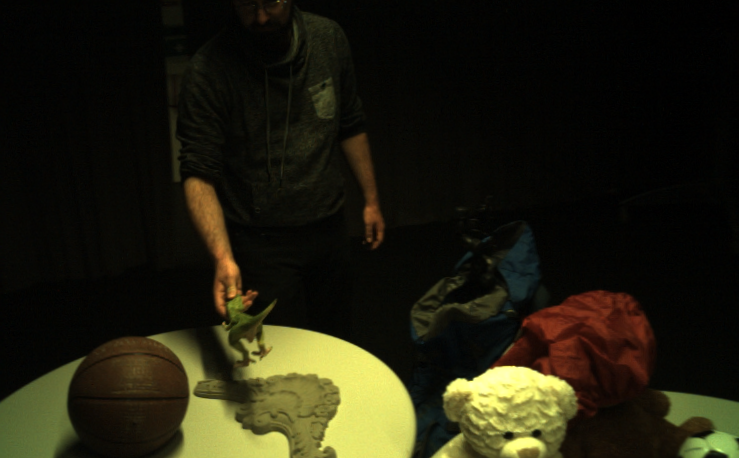}&
\includegraphics[width=0.23\linewidth]{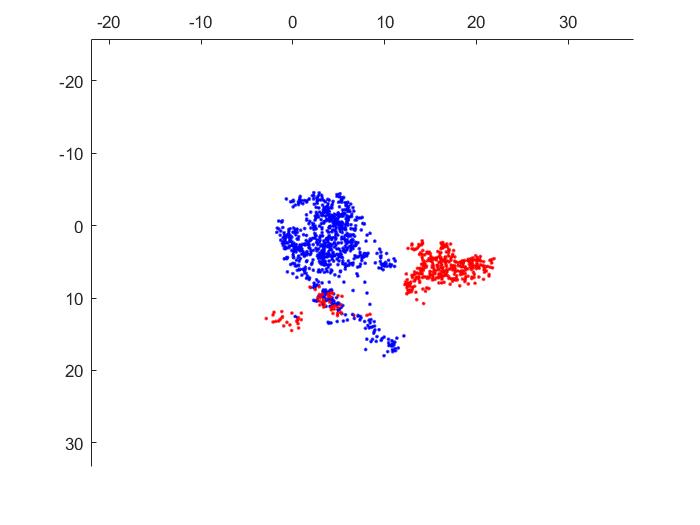}\\
\multicolumn{2}{c}{(i)} & \multicolumn{2}{c}{(j)} \\

\includegraphics[width=0.2\linewidth]{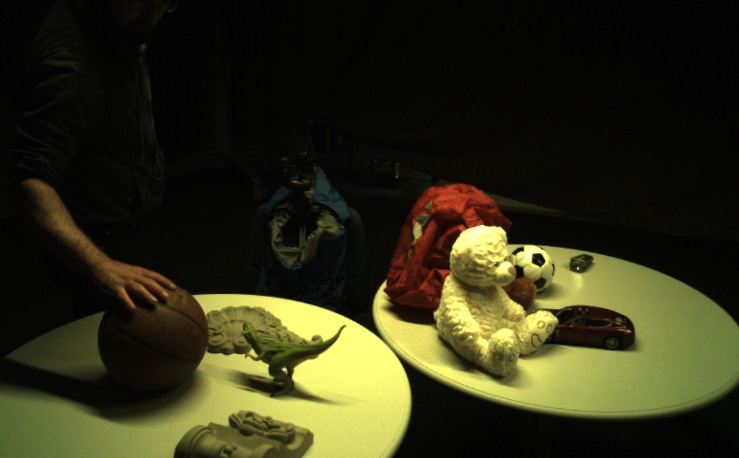}&
\includegraphics[width=0.23\linewidth]{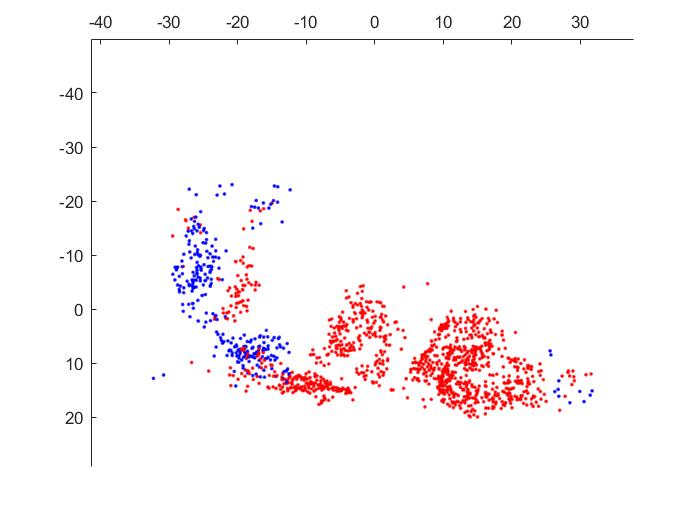}&
\includegraphics[width=0.2\linewidth]{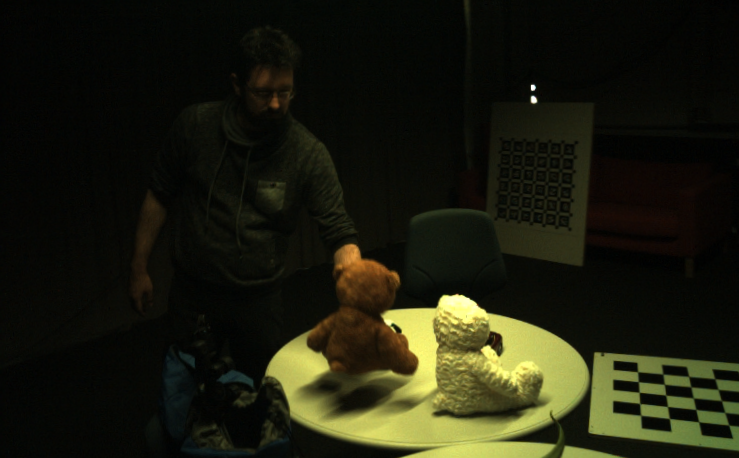}&
\includegraphics[width=0.23\linewidth]{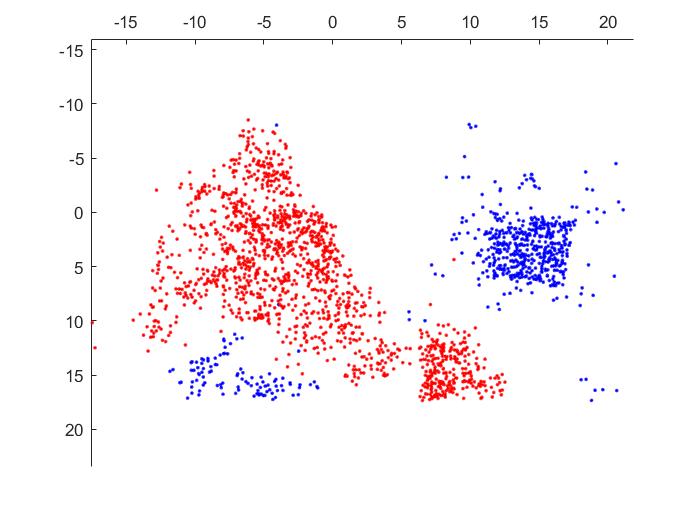}\\
\multicolumn{2}{c}{(k)} & \multicolumn{2}{c}{(l)} \\
\end{tabular}
\end{center}
\caption{Results of our method on the sequences extracted from the ETH 3D dataset. Foreground is blue, background is red.}
\label{fig:results_eth}
\end{figure}

The results of our method on the sequences from the ETH Dataset are in Figure \ref{fig:results_eth}. Our method reconstructs and segments all sequences successfully. In most of the results, a person is assigned to the foreground. In sequences (b), (d), (g) the person is not reconstructed. In sequences (a), (i) a person is assigned to the background. In sequence (k) the hand of the person is assigned to the foreground, while the rest of the body is assigned to the background. In sequence (l), the background and the foreground are swapped.

\section{Conclusion}
\noindent Our TBSfM method for complete 3D reconstruction of objects moving in front of a textured background outperforms the state of the art, Table~\ref{tab:error}. Real experiments on real data show that it reconstructs objects moving in front of a textured background,  Fig.~\ref{fig:results}, with more points and lower reprojection errors on the objects. All code and data will be made open source.

{\small
\bibliographystyle{ieee_fullname}
\bibliography{arxiv_paper}

\begin{thebibliography}{10}\itemsep=-1pt

\bibitem{DBLP:journals/ral/AlismailKBL17}
Hatem Alismail, Michael Kaess, Brett Browning, and Simon Lucey.
\newblock Direct visual odometry in low light using binary descriptors.
\newblock {\em {IEEE} Robotics and Automation Letters}, 2(2):444--451, 2017.

\bibitem{DBLP:journals/corr/abs-1905-09043}
Federica Arrigoni and Tom{\'{a}}s Pajdla.
\newblock Robust motion segmentation from pairwise matches.
\newblock {\em CoRR}, abs/1905.09043, 2019.

\bibitem{DBLP:conf/eccv/BarathM18}
Daniel Barath and Jiri Matas.
\newblock Multi-class model fitting by energy minimization and mode-seeking.
\newblock In {\em Computer Vision - {ECCV} 2018 - 15th European Conference,
  Munich, Germany, September 8-14, 2018, Proceedings, Part {XVI}}, pages
  229--245, 2018.

\bibitem{specialeffects}
Alastair Barber, Darren Cosker, Oliver James, Ted Waine, and Radhika Patel.
\newblock Camera tracking in visual effects an industry perspective of
  structure from motion.
\newblock pages 45--54, 07 2016.

\bibitem{DBLP:conf/grapp/BullingerBA21}
Sebastian Bullinger, Christoph Bodensteiner, and Michael Arens.
\newblock A photogrammetry-based framework to facilitate image-based modeling
  and automatic camera tracking.
\newblock In A.~Augusto de Sousa, Vlastimil Havran, Jos{\'{e}} Braz, and Kadi
  Bouatouch, editors, {\em Proceedings of the 16th International Joint
  Conference on Computer Vision, Imaging and Computer Graphics Theory and
  Applications, {VISIGRAPP} 2021, Volume 1: GRAPP, Online Streaming, February
  8-10, 2021}, pages 106--112. {SCITEPRESS}, 2021.

\bibitem{DBLP:journals/tit/CandesT10}
Emmanuel~J. Cand{\`{e}}s and Terence Tao.
\newblock The power of convex relaxation: near-optimal matrix completion.
\newblock {\em {IEEE} Trans. Inf. Theory}, 56(5):2053--2080, 2010.

\bibitem{DBLP:conf/cvpr/ChinSW10}
Tat{-}Jun Chin, David Suter, and Hanzi Wang.
\newblock Multi-structure model selection via kernel optimisation.
\newblock In {\em The Twenty-Third {IEEE} Conference on Computer Vision and
  Pattern Recognition, {CVPR} 2010, San Francisco, CA, USA, 13-18 June 2010},
  pages 3586--3593, 2010.

\bibitem{DBLP:journals/ijcv/CosteiraK98}
Jo{\~{a}}o~Paulo Costeira and Takeo Kanade.
\newblock A multibody factorization method for independently moving objects.
\newblock {\em International Journal of Computer Vision}, 29(3):159--179, 1998.

\bibitem{8939087}
H. {Desu}, R. {Gothandaraman}, and S. {Muthuswamy}.
\newblock 3d digital reconstruction of heritage artifacts: Parametric
  evaluation.
\newblock In {\em 2018 Tenth International Conference on Advanced Computing
  (ICoAC)}, pages 333--338, 2018.

\bibitem{cartography}
Jean Doumit.
\newblock Structure from motion technology for macro scale objects cartography.
\newblock 01 2016.

\bibitem{Dusmanu-ICCV-2019}
Mihai Dusmanu, Ignacio Rocco, Tom{\'{a}}s Pajdla, Marc Pollefeys, Josef Sivic,
  Akihiko Torii, and Torsten Sattler.
\newblock D2-net: {A} trainable {CNN} for joint description and detection of
  local features.
\newblock In {\em {IEEE} Conference on Computer Vision and Pattern Recognition,
  {CVPR} 2019, Long Beach, CA, USA, June 16-20, 2019}, pages 8092--8101, 2019.

\bibitem{DBLP:journals/pami/ElhamifarV13}
Ehsan Elhamifar and Ren{\'{e}} Vidal.
\newblock Sparse subspace clustering: Algorithm, theory, and applications.
\newblock {\em {IEEE} Trans. Pattern Anal. Mach. Intell.}, 35(11):2765--2781,
  2013.

\bibitem{DBLP:conf/iccvw/EnqvistKO11}
Olof Enqvist, Fredrik Kahl, and Carl Olsson.
\newblock Non-sequential structure from motion.
\newblock In {\em {IEEE} International Conference on Computer Vision Workshops,
  {ICCV} 2011 Workshops, Barcelona, Spain, November 6-13, 2011}, pages
  264--271. {IEEE} Computer Society, 2011.

\bibitem{DBLP:conf/iccv/FayadRA11}
Jo{\~{a}}o Fayad, Chris Russell, and Lourdes Agapito.
\newblock Automated articulated structure and 3d shape recovery from point
  correspondences.
\newblock In {\em {IEEE} International Conference on Computer Vision, {ICCV}
  2011, Barcelona, Spain, November 6-13, 2011}, pages 431--438, 2011.

\bibitem{DBLP:conf/eccv/FitzgibbonZ00}
Andrew~W. Fitzgibbon and Andrew Zisserman.
\newblock Multibody structure and motion: 3-d reconstruction of independently
  moving objects.
\newblock In {\em Computer Vision - {ECCV} 2000, 6th European Conference on
  Computer Vision, Dublin, Ireland, June 26 - July 1, 2000, Proceedings, Part
  {I}}, pages 891--906, 2000.

\bibitem{DBLP:conf/nips/FragkiadakiSAM14}
Katerina Fragkiadaki, Marta Salas, Pablo~Andr{\'{e}}s Arbel{\'{a}}ez, and
  Jitendra Malik.
\newblock Grouping-based low-rank trajectory completion and 3d reconstruction.
\newblock In Zoubin Ghahramani, Max Welling, Corinna Cortes, Neil~D. Lawrence,
  and Kilian~Q. Weinberger, editors, {\em Advances in Neural Information
  Processing Systems 27: Annual Conference on Neural Information Processing
  Systems 2014, December 8-13 2014, Montreal, Quebec, Canada}, pages 55--63,
  2014.

\bibitem{epic-games}
Epic {G}ames.
\newblock Unreal {E}ngine.
\newblock \url{www.unrealengine.com}.

\bibitem{Hartley:2003:MVG:861369}
Richard Hartley and Andrew Zisserman.
\newblock {\em Multiple View Geometry in Computer Vision}.
\newblock Cambridge University Press, New York, NY, USA, 2 edition, 2003.

\bibitem{PEARL}
Hossam Isack and Yuri Boykov.
\newblock Energy-based geometric multi-model fitting.
\newblock {\em International Journal of Computer Vision}, 97:123--147, 04 2012.

\bibitem{DBLP:conf/iccv/JiSL15}
Pan Ji, Mathieu Salzmann, and Hongdong Li.
\newblock Shape interaction matrix revisited and robustified: Efficient
  subspace clustering with corrupted and incomplete data.
\newblock In {\em 2015 {IEEE} International Conference on Computer Vision,
  {ICCV} 2015, Santiago, Chile, December 7-13, 2015}, pages 4687--4695, 2015.

\bibitem{Arun87}
D.~S.~Blostein K.~S.~Arun, T. S.~Huang.
\newblock Least-squares fitting of two 3-d point sets.
\newblock {\em IEEE transactions on pattern analysis and machine intelligence,
  vol. PAMI-9, No. 5. September 1987}, 1987.

\bibitem{DBLP:conf/3dim/KumarDL16}
Suryansh Kumar, Yuchao Dai, and Hongdong Li.
\newblock Multi-body non-rigid structure-from-motion.
\newblock In {\em Fourth International Conference on 3D Vision, 3DV 2016,
  Stanford, CA, USA, October 25-28, 2016}, pages 148--156. {IEEE} Computer
  Society, 2016.

\bibitem{DBLP:conf/iccv/KumarDL17}
Suryansh Kumar, Yuchao Dai, and Hongdong Li.
\newblock Monocular dense 3d reconstruction of a complex dynamic scene from two
  perspective frames.
\newblock In {\em {IEEE} International Conference on Computer Vision, {ICCV}
  2017, Venice, Italy, October 22-29, 2017}, pages 4659--4667. {IEEE} Computer
  Society, 2017.

\bibitem{DBLP:conf/iccv/KunduKJ11}
Abhijit Kundu, K.~Madhava Krishna, and C.~V. Jawahar.
\newblock Realtime multibody visual {SLAM} with a smoothly moving monocular
  camera.
\newblock In Dimitris~N. Metaxas, Long Quan, Alberto Sanfeliu, and Luc~Van
  Gool, editors, {\em {IEEE} International Conference on Computer Vision,
  {ICCV} 2011, Barcelona, Spain, November 6-13, 2011}, pages 2080--2087. {IEEE}
  Computer Society, 2011.

\bibitem{DBLP:conf/iros/KunduKS09}
Abhijit Kundu, K.~Madhava Krishna, and Jayanthi Sivaswamy.
\newblock Moving object detection by multi-view geometric techniques from a
  single camera mounted robot.
\newblock In {\em 2009 {IEEE/RSJ} International Conference on Intelligent
  Robots and Systems, October 11-15, 2009, St. Louis, MO, {USA}}, pages
  4306--4312. {IEEE}, 2009.

\bibitem{DBLP:journals/tits/LaiWYCZ17}
Taotao Lai, Hanzi Wang, Yan Yan, Tat{-}Jun Chin, and Wanlei Zhao.
\newblock Motion segmentation via a sparsity constraint.
\newblock {\em {IEEE} Trans. Intelligent Transportation Systems},
  18(4):973--983, 2017.

\bibitem{DBLP:conf/cvpr/LiV15}
Chun{-}Guang Li and Ren{\'{e}} Vidal.
\newblock Structured sparse subspace clustering: {A} unified optimization
  framework.
\newblock In {\em {IEEE} Conference on Computer Vision and Pattern Recognition,
  {CVPR} 2015, Boston, MA, USA, June 7-12, 2015}, pages 277--286, 2015.

\bibitem{DBLP:conf/cvpr/LiKSV07}
Ting Li, Vinutha Kallem, Dheeraj Singaraju, and Ren{\'{e}} Vidal.
\newblock Projective factorization of multiple rigid-body motions.
\newblock In {\em 2007 {IEEE} Computer Society Conference on Computer Vision
  and Pattern Recognition {(CVPR} 2007), 18-23 June 2007, Minneapolis,
  Minnesota, {USA}}, 2007.

\bibitem{DBLP:conf/iccv/LiGCZ13}
Zhuwen Li, Jiaming Guo, Loong{-}Fah Cheong, and Steven~Zhiying Zhou.
\newblock Perspective motion segmentation via collaborative clustering.
\newblock In {\em {IEEE} International Conference on Computer Vision, {ICCV}
  2013, Sydney, Australia, December 1-8, 2013}, pages 1369--1376, 2013.

\bibitem{DBLP:journals/pami/LiuLYSYM13}
Guangcan Liu, Zhouchen Lin, Shuicheng Yan, Ju Sun, Yong Yu, and Yi Ma.
\newblock Robust recovery of subspace structures by low-rank representation.
\newblock {\em {IEEE} Trans. Pattern Anal. Mach. Intell.}, 35(1):171--184,
  2013.

\bibitem{DBLP:conf/iccv/MagerandB17}
Ludovic Magerand and Alessio {Del Bue}.
\newblock Practical projective structure from motion (p2sfm).
\newblock In {\em {IEEE} International Conference on Computer Vision, {ICCV}
  2017, Venice, Italy, October 22-29, 2017}, pages 39--47. {IEEE} Computer
  Society, 2017.

\bibitem{DBLP:conf/cvpr/MagriF14}
Luca Magri and Andrea Fusiello.
\newblock T-linkage: {A} continuous relaxation of j-linkage for multi-model
  fitting.
\newblock In {\em 2014 {IEEE} Conference on Computer Vision and Pattern
  Recognition, {CVPR} 2014, Columbus, OH, USA, June 23-28, 2014}, pages
  3954--3961, 2014.

\bibitem{Magri14}
L. Magri and A. Fusiello.
\newblock T-linkage: a continuous relaxation of j-linkage for multi-model
  fitting.
\newblock {\em Computer Vision and Pattern Recognition, pages 3954–3961},
  2014.

\bibitem{DBLP:conf/bmvc/MagriF15}
Luca Magri and Andrea Fusiello.
\newblock Robust multiple model fitting with preference analysis and low-rank
  approximation.
\newblock In {\em Proceedings of the British Machine Vision Conference 2015,
  {BMVC} 2015, Swansea, UK, September 7-10, 2015}, pages 20.1--20.12, 2015.

\bibitem{DBLP:conf/cvpr/MagriF16}
Luca Magri and Andrea Fusiello.
\newblock Multiple models fitting as a set coverage problem.
\newblock In {\em 2016 {IEEE} Conference on Computer Vision and Pattern
  Recognition, {CVPR} 2016, Las Vegas, NV, USA, June 27-30, 2016}, pages
  3318--3326, 2016.

\bibitem{Nister-ICCV-2007}
David Nist{\'{e}}r, Fredrik Kahl, and Henrik Stew{\'{e}}nius.
\newblock Structure from motion with missing data is np-hard.
\newblock In {\em {IEEE} 11th International Conference on Computer Vision,
  {ICCV} 2007, Rio de Janeiro, Brazil, October 14-20, 2007}, pages 1--7, 2007.

\bibitem{DBLP:conf/cvpr/NisterNB04}
David Nist{\'{e}}r, Oleg Naroditsky, and James~R. Bergen.
\newblock Visual odometry.
\newblock In {\em 2004 {IEEE} Computer Society Conference on Computer Vision
  and Pattern Recognition {(CVPR} 2004), with CD-ROM, 27 June - 2 July 2004,
  Washington, DC, {USA}}, pages 652--659, 2004.

\bibitem{DBLP:journals/pami/OzdenSG10}
Kemal~Egemen Ozden, Konrad Schindler, and Luc~Van Gool.
\newblock Multibody structure-from-motion in practice.
\newblock {\em {IEEE} Trans. Pattern Anal. Mach. Intell.}, 32(6):1134--1141,
  2010.

\bibitem{DBLP:conf/avr/PaolisLGDP20}
Lucio Tommaso~De Paolis, Valerio~De Luca, Carola Gatto, Giovanni D'Errico, and
  Giovanna~Ilenia Paladini.
\newblock Photogrammetric {3D} reconstruction of small objects for a real-time
  fruition.
\newblock In Lucio Tommaso~De Paolis and Patrick Bourdot, editors, {\em
  Augmented Reality, Virtual Reality, and Computer Graphics - 7th International
  Conference, {AVR} 2020, Lecce, Italy, September 7-10, 2020, Proceedings, Part
  {I}}, volume 12242 of {\em Lecture Notes in Computer Science}, pages
  375--394. Springer, 2020.

\bibitem{DBLP:conf/cvpr/PhamCYS12}
Trung{-}Thanh Pham, Tat{-}Jun Chin, Jin Yu, and David Suter.
\newblock The random cluster model for robust geometric fitting.
\newblock In {\em 2012 {IEEE} Conference on Computer Vision and Pattern
  Recognition, Providence, RI, USA, June 16-21, 2012}, pages 710--717, 2012.

\bibitem{DBLP:journals/tip/QianCZ05}
Gang Qian, Rama Chellappa, and Qinfen Zheng.
\newblock Bayesian algorithms for simultaneous structure from motion estimation
  of multiple independently moving objects.
\newblock {\em {IEEE} Trans. Image Processing}, 14(1):94--109, 2005.

\bibitem{DBLP:conf/cvpr/RaoTVM08}
Shankar~R. Rao, Roberto Tron, Ren{\'{e}} Vidal, and Yi Ma.
\newblock Motion segmentation via robust subspace separation in the presence of
  outlying, incomplete, or corrupted trajectories.
\newblock In {\em 2008 {IEEE} Computer Society Conference on Computer Vision
  and Pattern Recognition {(CVPR} 2008), 24-26 June 2008, Anchorage, Alaska,
  {USA}}, 2008.

\bibitem{DBLP:conf/nips/RoccoCATPS18}
Ignacio Rocco, Mircea Cimpoi, Relja Arandjelovic, Akihiko Torii, Tom{\'{a}}s
  Pajdla, and Josef Sivic.
\newblock Neighbourhood consensus networks.
\newblock In {\em Advances in Neural Information Processing Systems 31: Annual
  Conference on Neural Information Processing Systems 2018, NeurIPS 2018, 3-8
  December 2018, Montr{\'{e}}al, Canada.}, pages 1658--1669, 2018.

\bibitem{DBLP:conf/ismar/RoussosRGA12}
Anastasios Roussos, Chris Russell, Ravi Garg, and Lourdes Agapito.
\newblock Dense multibody motion estimation and reconstruction from a handheld
  camera.
\newblock In {\em 11th {IEEE} International Symposium on Mixed and Augmented
  Reality, {ISMAR} 2012, Atlanta, GA, USA, November 5-8, 2012}, pages 31--40,
  2012.

\bibitem{DBLP:conf/eccv/RussellYA14}
Chris Russell, Rui Yu, and Lourdes Agapito.
\newblock Video pop-up: Monocular 3d reconstruction of dynamic scenes.
\newblock In {\em Computer Vision - {ECCV} 2014 - 13th European Conference,
  Zurich, Switzerland, September 6-12, 2014, Proceedings, Part {VII}}, pages
  583--598, 2014.

\bibitem{DBLP:conf/icra/SabzevariS14}
Reza Sabzevari and Davide Scaramuzza.
\newblock Monocular simultaneous multi-body motion segmentation and
  reconstruction from perspective views.
\newblock In {\em 2014 {IEEE} International Conference on Robotics and
  Automation, {ICRA} 2014, Hong Kong, China, May 31 - June 7, 2014}, pages
  23--30, 2014.

\bibitem{DBLP:journals/pami/SattlerLK17}
Torsten Sattler, Bastian Leibe, and Leif Kobbelt.
\newblock Efficient {\&} effective prioritized matching for large-scale
  image-based localization.
\newblock {\em {IEEE} Trans. Pattern Anal. Mach. Intell.}, 39(9):1744--1756,
  2017.

\bibitem{DBLP:journals/ijcv/SchindlerSW08}
Konrad Schindler, David Suter, and Hanzi Wang.
\newblock A model-selection framework for multibody structure-and-motion of
  image sequences.
\newblock {\em International Journal of Computer Vision}, 79(2):159--177, 2008.

\bibitem{schoenberger2016sfm}
Johannes~Lutz Sch\"{o}nberger and Jan-Michael Frahm.
\newblock Structure-from-motion revisited.
\newblock In {\em IEEE Conference on Computer Vision and Pattern Recognition
  (CVPR)}, 2016.

\bibitem{DBLP:conf/cvpr/SchopsSP19}
Thomas Sch{\"{o}}ps, Torsten Sattler, and Marc Pollefeys.
\newblock {BAD} {SLAM:} bundle adjusted direct {RGB-D} {SLAM}.
\newblock In {\em {IEEE} Conference on Computer Vision and Pattern Recognition,
  {CVPR} 2019, Long Beach, CA, USA, June 16-20, 2019}, pages 134--144. Computer
  Vision Foundation / {IEEE}, 2019.

\bibitem{DBLP:journals/tog/SnavelySS06}
Noah Snavely, Steven~M. Seitz, and Richard Szeliski.
\newblock Photo tourism: exploring photo collections in 3d.
\newblock {\em {ACM} Trans. Graph.}, 25(3):835--846, 2006.

\bibitem{DBLP:journals/ijcv/SnavelySS08}
Noah Snavely, Steven~M. Seitz, and Richard Szeliski.
\newblock Modeling the world from internet photo collections.
\newblock {\em International Journal of Computer Vision}, 80(2):189--210, 2008.

\bibitem{Srajer16}
Filip Srajer.
\newblock Image matching for dynamic scenes.
\newblock Master's thesis, Czech Technical University in Prague, 2016.

\bibitem{DBLP:journals/pami/SvarmEKO17}
Linus Svarm, Olof Enqvist, Fredrik Kahl, and Magnus Oskarsson.
\newblock City-scale localization for cameras with known vertical direction.
\newblock {\em {IEEE} Trans. Pattern Anal. Mach. Intell.}, 39(7):1455--1461,
  2017.

\bibitem{DBLP:conf/cvpr/TairaOSCPSPT18}
Hajime Taira, Masatoshi Okutomi, Torsten Sattler, Mircea Cimpoi, Marc
  Pollefeys, Josef Sivic, Tom{\'{a}}s Pajdla, and Akihiko Torii.
\newblock Inloc: Indoor visual localization with dense matching and view
  synthesis.
\newblock In {\em 2018 {IEEE} Conference on Computer Vision and Pattern
  Recognition, {CVPR} 2018, Salt Lake City, UT, USA, June 18-22, 2018}, pages
  7199--7209, 2018.

\bibitem{DBLP:journals/tkde/ThakoorG11}
Ninad Thakoor and Jean~X. Gao.
\newblock Branch-and-bound for model selection and its computational
  complexity.
\newblock {\em {IEEE} Trans. Knowl. Data Eng.}, 23(5):655--668, 2011.

\bibitem{Tola05}
Engin Tola, Sebastian Knorr, Evren Imre, A.~Aydın Alatan, and Thomas Sikora.
\newblock Structure from motion in dynamic scenes with multiple motions.
\newblock {\em 2nd Workshop on Immersive Communication and Broadcast Systems},
  2005.

\bibitem{DBLP:conf/eccv/ToldoF08}
Roberto Toldo and Andrea Fusiello.
\newblock Robust multiple structures estimation with j-linkage.
\newblock In {\em Computer Vision - {ECCV} 2008, 10th European Conference on
  Computer Vision, Marseille, France, October 12-18, 2008, Proceedings, Part
  {I}}, pages 537--547, 2008.

\bibitem{Toldo08}
R. Toldo and A. Fusiello.
\newblock Robust multiple structures estimation with j-linkage.
\newblock {\em European Conference on Computer Vision, volume 5302, pages
  537–547}, 2008.

\bibitem{Torr98}
Philip Torr.
\newblock Geometric motion segmentation and model selection.
\newblock {\em Philos. Trans. Roy. Soc. A}, 356, 10 1997.

\bibitem{Triggs00}
B. Triggs, P.~F. McLauchlan, R.~I. Hartley, and A.~W. Fitzgibbon.
\newblock Bundle adjustment - a modern synthesis.
\newblock {\em Proceedings of the International Workshop on Vision Algorithms:
  Theory and Practice}, 2000.

\bibitem{DBLP:conf/cvpr/VidalMS03}
Ren{\'{e}} Vidal, Yi Ma, and Shankar Sastry.
\newblock Generalized principal component analysis {(GPCA)}.
\newblock In {\em 2003 {IEEE} Computer Society Conference on Computer Vision
  and Pattern Recognition {(CVPR} 2003), 16-22 June 2003, Madison, WI, {USA}},
  pages 621--628, 2003.

\bibitem{DBLP:journals/ijcv/VidalMSS06}
Ren{\'{e}} Vidal, Yi Ma, Stefano Soatto, and Shankar Sastry.
\newblock Two-view multibody structure from motion.
\newblock {\em International Journal of Computer Vision}, 68(1):7--25, 2006.

\bibitem{DBLP:journals/ijcv/VidalTH08}
Ren{\'{e}} Vidal, Roberto Tron, and Richard~I. Hartley.
\newblock Multiframe motion segmentation with missing data using
  powerfactorization and {GPCA}.
\newblock {\em International Journal of Computer Vision}, 79(1):85--105, 2008.

\bibitem{Vincent01}
Etienne Vincent and Robert Laganiere.
\newblock Detecting planar homographies in an image pair.
\newblock pages 182 -- 187, 02 2001.

\bibitem{DBLP:conf/cvpr/VoNS16}
Minh Vo, Srinivasa~G. Narasimhan, and Yaser Sheikh.
\newblock Spatiotemporal bundle adjustment for dynamic 3d reconstruction.
\newblock In {\em 2016 {IEEE} Conference on Computer Vision and Pattern
  Recognition, {CVPR} 2016, Las Vegas, NV, USA, June 27-30, 2016}, pages
  1710--1718. {IEEE} Computer Society, 2016.

\bibitem{Luxburg07}
Ulrike von Luxburg.
\newblock A tutorial on spectral clustering.
\newblock {\em Statistics and Computing, 17 (4)}, 2007.

\bibitem{archaeology}
Mark Willis, Charles Koenig, Stephen Black, and Amanda Castaneda.
\newblock Archeological 3d mapping: the structure from motion revolution.
\newblock {\em Journal of Texas Archeology and History}, 3:1--36, 06 2016.

\bibitem{DBLP:journals/jmiv/WuH06}
Yihong Wu and Zhanyi Hu.
\newblock Pnp problem revisited.
\newblock {\em Journal of Mathematical Imaging and Vision}, 24(1):131--141,
  2006.

\bibitem{DBLP:journals/prl/XuOK90}
Lei Xu, Erkki Oja, and Pekka Kultanen.
\newblock A new curve detection method: Randomized hough transform {(RHT)}.
\newblock {\em Pattern Recognition Letters}, 11(5):331--338, 1990.

\bibitem{Xu18}
Xun Xu, Loong~Fah Cheong, and Zhuwen Li.
\newblock Motion segmentation by exploiting complementary geometric models.
\newblock In {\em 2018 {IEEE} Conference on Computer Vision and Pattern
  Recognition, {CVPR} 2018, Salt Lake City, UT, USA, June 18-22, 2018}, pages
  2859--2867, 2018.

\bibitem{LSA}
Jingyu Yan and Marc Pollefeys.
\newblock A general framework for motion segmentation: Independent,
  articulated, rigid, non-rigid, degenerate and non-degenerate.
\newblock In Ale{\v{s}} Leonardis, Horst Bischof, and Axel Pinz, editors, {\em
  Computer Vision -- ECCV 2006}, pages 94--106, Berlin, Heidelberg, 2006.
  Springer Berlin Heidelberg.

\bibitem{DBLP:journals/cviu/ZappellaBLS13}
Luca Zappella, Alessio {Del Bue}, Xavier Llad{\'{o}}, and Joaquim Salvi.
\newblock Joint estimation of segmentation and structure from motion.
\newblock {\em Computer Vision and Image Understanding}, 117(2):113--129, 2013.

\bibitem{DBLP:conf/eccv/ZhangK06}
Wei Zhang and Jana Koseck{\'{a}}.
\newblock Nonparametric estimation of multiple structures with outliers.
\newblock In {\em Dynamical Vision, {ICCV} 2005 and {ECCV} 2006 Workshops,
  {WDV} 2005 and {WDV} 2006, Beijing, China, October 21, 2005, Graz, Austria,
  May 13, 2006. Revised Papers}, pages 60--74, 2006.

\bibitem{Zuliani05}
Marco Zuliani, C.S. Kenney, and B. Manjunath.
\newblock The multiransac algorithm and its application to detect planar
  homographies.
\newblock pages III -- 153, 10 2005.

\end{thebibliography}
}

\newpage
\onecolumn
\begin{center}
    {\Large\bf Reconstructing Small 3D Objects in front of a Textured Background} \\
\end{center}


Here we present additional details, such as experiments and detailed derivations. Sec.~\ref{sec:exp} contains the experiments with TBSfM. Sec.~\ref{sec:datasets} provides a more detailed description of the datasets introduced in Sec.~4 of the main paper, as well as the resulting models reconstructed with TBSfM algorithm. Sec.~\ref{sec:comparison} compares the models reconstructed with TBSfM and the models reconstructed with a single-body SfM pipeline \cite{schoenberger2016sfm} from images without background. Sec.~\ref{sec:yasfm} shows the result of multi-body SfM pipeline YASFM \cite{Srajer16} on images from our dataset. Sec.~\ref{sec:mode} shows the result of the segmentation by MODE \cite{DBLP:journals/corr/abs-1905-09043} and Sec.~\ref{sec:subset} shows the result of the segmentation by Subset \cite{Xu18}. Sec.~\ref{sec:colmap} shows the model which outputs a single-body SfM pipeline COLMAP \cite{schoenberger2016sfm}, if it is given images from different takes. Sec~\ref{sec:eth} contains the experiments conducted with ETH 3D dataset \cite{DBLP:conf/cvpr/SchopsSP19}.

Sec.~\ref{sec:motion1} contains the derivation of the formula for foreground motion, which is used in Sec.~\ref{sec:loc-group-s}. Sec.~\ref{sec:loc-group-s} describes an additional criterion for local observation grouping, which is based on the foreground motion. Sec.~\ref{sec:trans} contains the derivation of the formula for transformation of points between reconstructed models, which is used in Sec.~\ref{sec:cameras}. Sec.~\ref{sec:cameras} describes transformation of the cameras, which is used in Sec.~3.6 of the main paper.

Code in C++ is available on Github. Please, follow the readme file to compile and run a representative experiment.

\vspace{-0.2cm}

\section{Experiments}\label{sec:exp}

\vspace{-0.2cm}
\subsection{Datasets and results of TBSfM}\label{sec:datasets}
In this section we give a detailed description of the datasets introduced in Sec.~4 of the main paper. We provide a further view on the models reconstructed by TBSfM from our datasets, which are described in Sec.~4.1 in the main paper.

Table \ref{tab:datasets1} describes the properties of the datasets, such as the number of takes, the number of images and special characteristics of the objects (planar, repetitive, ...) or the datasets (translational motion only). Properties of the reconstructed models are shown in Table \ref{tab:results}.

\begin{figure}[H]
\begin{center}
\begin{tabular}{c|c}
\includegraphics[width=0.3\linewidth]{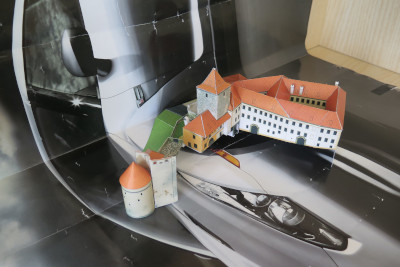}&  
\includegraphics[width=0.25\linewidth]{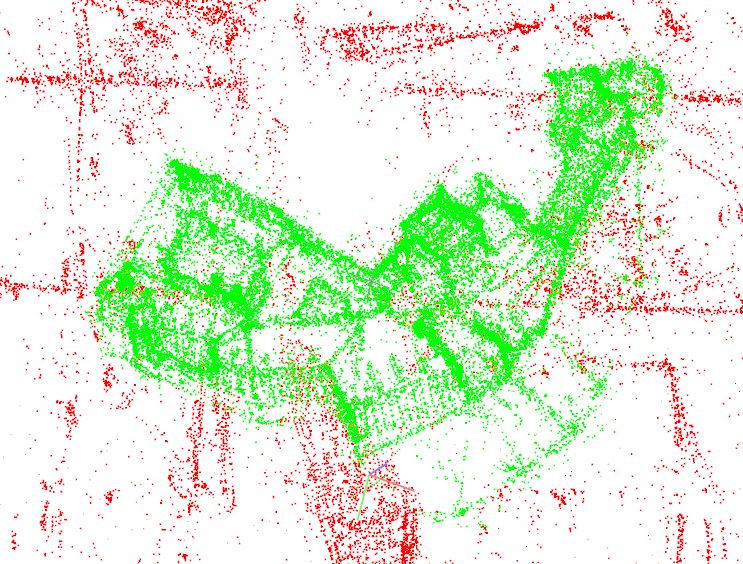}\\
(a) & (b)
\end{tabular}
\end{center}
\caption{Reconstructed datasets, Part 1.}
\label{fig:recon1}
\end{figure}

\begin{figure}[H]
\begin{center}
\begin{tabular}{c|c}
\includegraphics[width=0.3\linewidth]{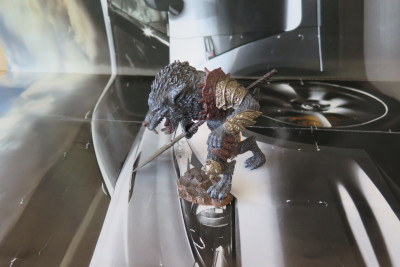}&  
\includegraphics[width=0.25\linewidth]{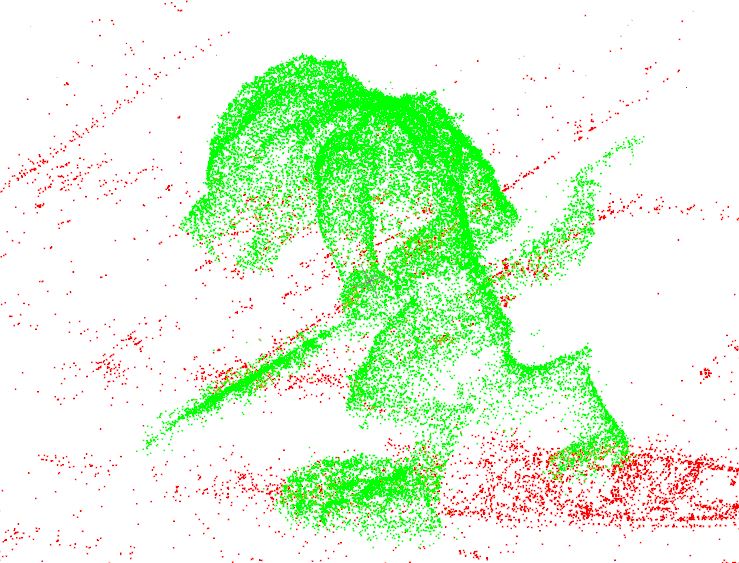}\\
(a) & (b)\\
\includegraphics[width=0.3\linewidth]{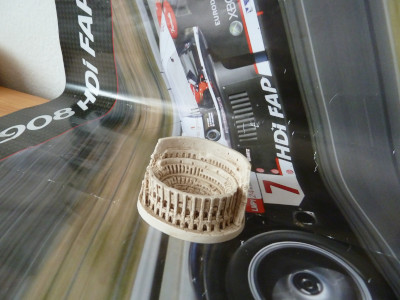}&  
\includegraphics[width=0.25\linewidth]{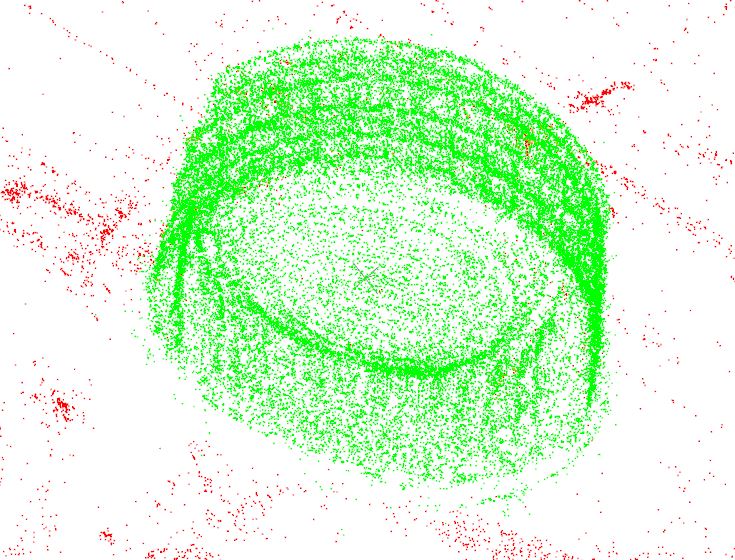}\\
(c) & (d)\\
\includegraphics[width=0.3\linewidth]{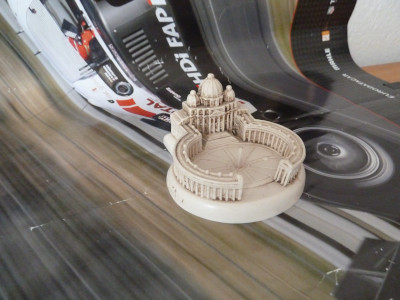}&  
\includegraphics[width=0.25\linewidth]{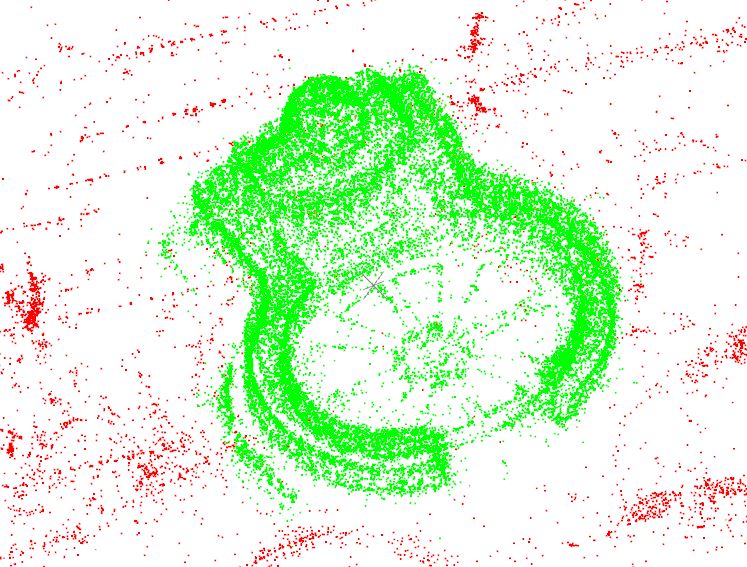}\\
(e) & (f)\\
\includegraphics[width=0.3\linewidth]{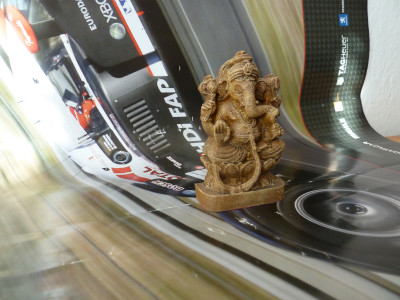}&  
\includegraphics[width=0.25\linewidth]{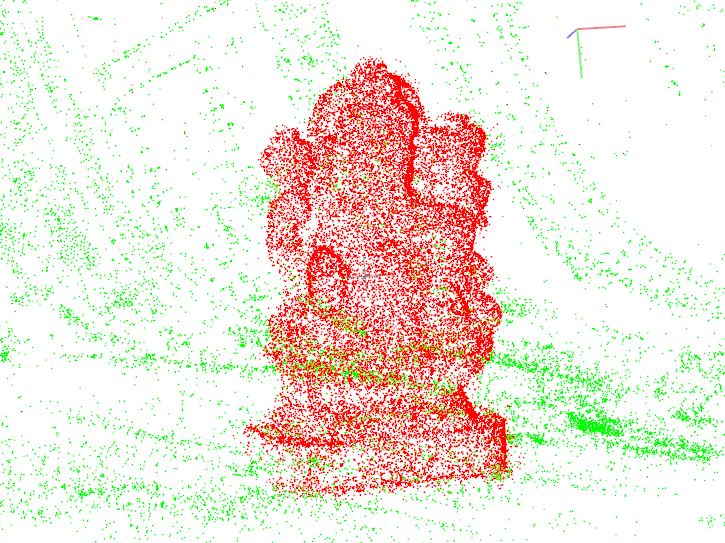}\\
(g) & (h)\\

\end{tabular}
\end{center}
\caption{Reconstructed datasets, Part 2.}
\label{fig:recon2}
\end{figure}

\begin{figure}[H]
\begin{center}
\begin{tabular}{c|c}
\includegraphics[width=0.3\linewidth]{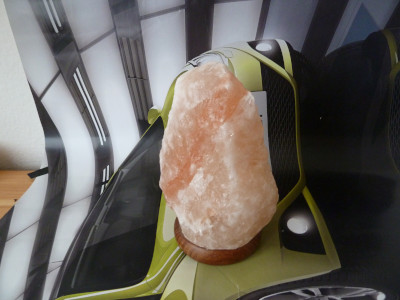}&
\includegraphics[width=0.25\linewidth]{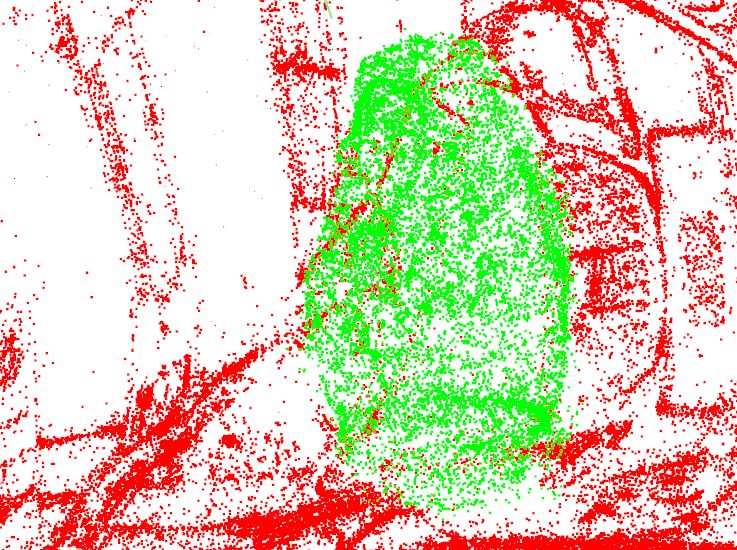}\\
(a) & (b)\\
\includegraphics[width=0.3\linewidth]{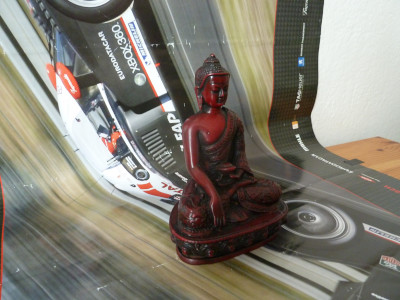}&  
\\
(c) & (d)\\
\includegraphics[width=0.3\linewidth]{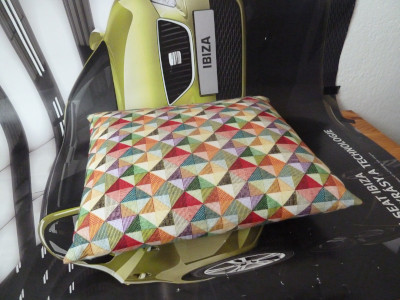}&  
\includegraphics[width=0.3\linewidth]{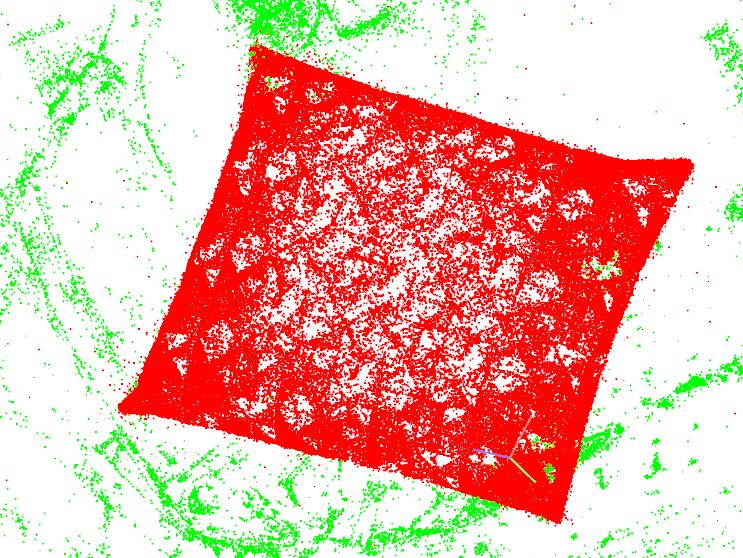}\\
(e) & (f)\\
\includegraphics[width=0.3\linewidth]{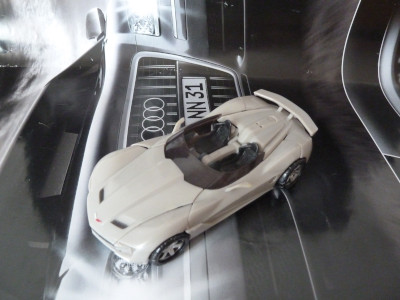}&  
\includegraphics[width=0.3\linewidth]{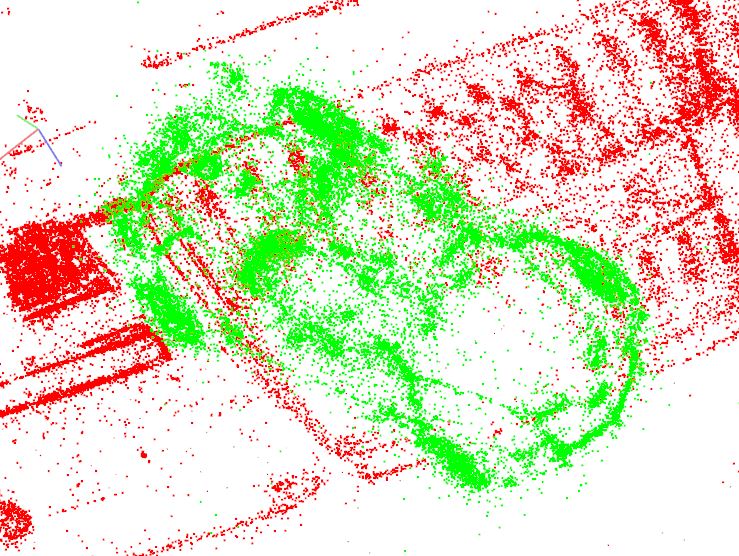}\\
(g) & (h)\\

\end{tabular}
\end{center}
\caption{Reconstructed datasets, Part 3.}
\label{fig:recon3}
\end{figure}

\begin{figure}[h!]
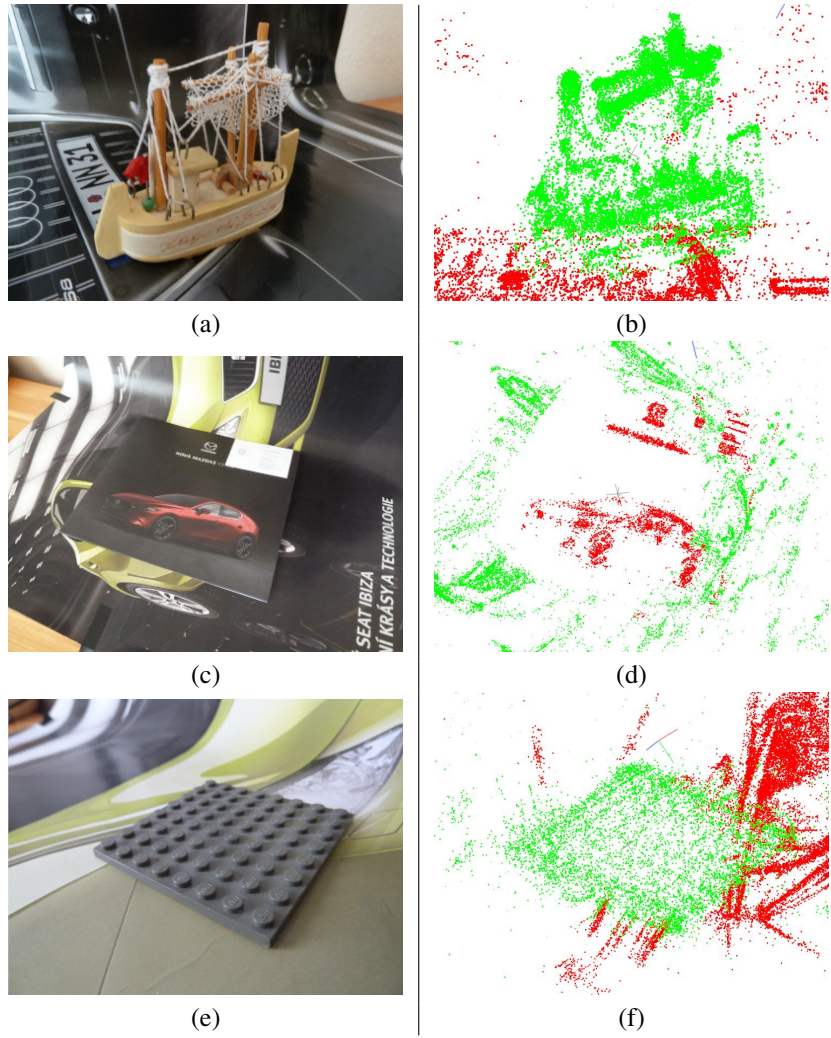

\begin{center}
\begin{tabular}{c|c}
\includegraphics[width=0.3\linewidth]{Figs/ship.jpg}&  
\includegraphics[width=0.3\linewidth]{Figs/ship_n.JPG}\\
(a) & (b)\\
\includegraphics[width=0.3\linewidth]{Figs/catalog.jpg}&  
\includegraphics[width=0.3\linewidth]{Figs/mazda_n.JPG}\\
(c) & (d)\\
\includegraphics[width=0.3\linewidth]{Figs/lego.jpg}&  
\includegraphics[width=0.3\linewidth]{Figs/lego_n.JPG}\\
(e) & (f)\\
\end{tabular}
\end{center}
\caption{Reconstructed datasets, Part 4.}
\label{fig:recon4}
\end{figure}

The images of the scenes together with the results of TBSfM on them are shown in Figures \ref{fig:recon1}, \ref{fig:recon2}, \ref{fig:recon3}, \ref{fig:recon4}. In each row, the image on the left is an image taken from the dataset, while the image on the right depicts the 3D model reconstructed from the dataset with TBSfM. TBSfM was not able to reconstruct the dataset "Buddha" (Fig.~\ref{fig:recon3}(c)), due to the small density of features, reflections on the surface of the object and small overlaps between neighboring takes. Other models were reconstructed successfully. The points assigned to the foreground are displayed green, while the points assigned to the background are red. Figures \ref{fig:recon1}(h), \ref{fig:recon2}(f), \ref{fig:recon3}(d) show that if the foreground is dominant, it may be swapped with the background. Figures \ref{fig:recon1}, \ref{fig:recon2}, \ref{fig:recon3}, \ref{fig:recon4} show that the quality of the segmentation, as well as the quality of the reconstruction is high.

\begin{table}[H]
\begin{center}
\begin{tabular}{ |c|c|c|c|c|c| } 
 \hline
 Foreground & Background & Takes & Images & Special & Figure \\ 
 \hline \hline
 Daliborka & Nissan GTR & 8 & 338 & NONE & \ref{fig:recon1}(a)\\ 
 Lycan & Nissan GTR & 8 & 140 & NONE & \ref{fig:recon2}(a)\\ 
 Colosseum & Peugeot 908 & 8 & 294 & NONE & \ref{fig:recon2}(c)\\
 Vatican & Peugeot 908 & 8 & 270 & NONE & \ref{fig:recon2}(e)\\
 Ganesha & Peugeot 908 & 4 & 160 & Only translation & \ref{fig:recon2}(g)\\
 Salt lamp & Seat Ibiza & 8 & 330 & Translucent & \ref{fig:recon3}(a)\\
 Buddha & Peugeot 908 & 8 & 300 & Shiny & \ref{fig:recon3}(c)\\
 Pillow & Seat Ibiza & 8 & 278 & Repetitive & \ref{fig:recon3}(e)\\
 Car & Audi S8 & 8 & 329 & NONE & \ref{fig:recon3}(g)\\
 Ship & Audi S8 & 8 & 277 & NONE & \ref{fig:recon4}(a)\\
 Catalog & Seat Ibiza & 5 & 150 & Planar & \ref{fig:recon4}(c)\\
 Lego & Seat Ibiza & 4 & 264 & Repetitive, Planar & \ref{fig:recon4}(e)\\
 \hline
\end{tabular}
\end{center}
\caption{Properties of the datasets.}
\label{tab:datasets1}
\end{table}

\begin{table}[H]
\begin{center}
\begin{tabular}{ |c|c|c|c|c| } 
 \hline
 Dataset & Figure  & Points Fg. & Points Bck. & Swapped \\ 
 \hline \hline
 Daliborka & \ref{fig:recon1}(b) & 37755 & 26324 & NO \\ 
 Lycan & \ref{fig:recon2}(b) & 40347 & 12345 & NO \\ 
 Colosseum & \ref{fig:recon2}(d) & 52397 & 33335 & NO \\
 Vatican & \ref{fig:recon2}(f) & 39024 & 35160 & NO \\
 Ganesha & \ref{fig:recon2}(h) & 35179 & 40643 & YES \\
 Salt lamp & \ref{fig:recon3}(b) & 15305 & 63828 & NO \\
 Pillow & \ref{fig:recon3}(d) & 134086 & 30532 & YES \\
 Car & \ref{fig:recon3}(f) & 25229 & 46878 & NO \\
 Ship & \ref{fig:recon4}(h) & 18244 & 21983 & NO \\
 Catalog & \ref{fig:recon4}(b) & 8004 & 20757 & YES \\
 Lego & \ref{fig:recon4}(d) & 7692 & 29964 & NO \\
 
 \hline
\end{tabular}
\end{center}
\caption{Results of the reconstructions.}
\label{tab:results}
\end{table}

\newpage
\newpage

\newpage

\subsection{Comparison with single body SfM}\label{sec:comparison}
\begin{figure}[H]
\begin{center}
\begin{tabular}{c c c}
\includegraphics[width=0.3\linewidth]{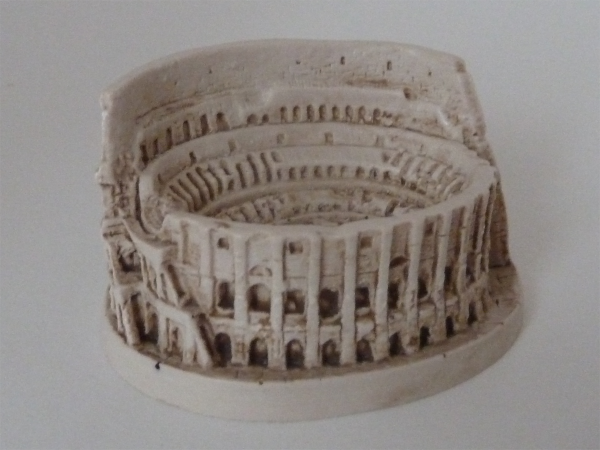}&  
\includegraphics[width=0.3\linewidth]{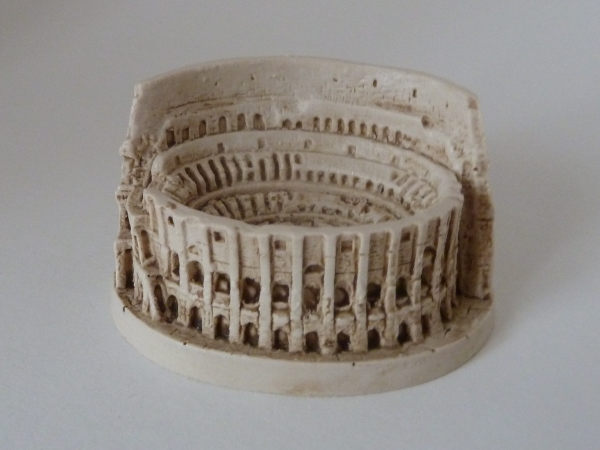}&
\includegraphics[width=0.3\linewidth]{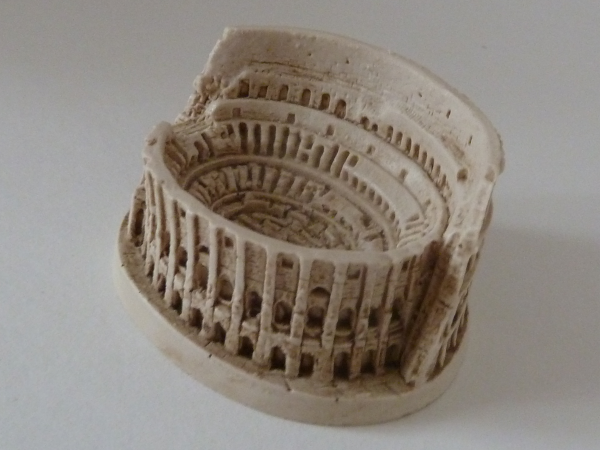}\\
\end{tabular}
\end{center}
\vspace{-0.4cm}
\caption{Example of the images of an object without the background.}
\label{fig:nobg}
\end{figure}

This section shows the result of the comparison proposed in Sec.~4.3 of the main paper. We show, for which kinds of objects and how much, TBSfM improves the reconstruction. We have created sets of photographs of the same objects but this time without any background, like in Figure \ref{fig:nobg}. We compare the models reconstructed with TBSfM on the data with background and the models of the same objects reconstructed with state-of-the-art SfM pipeline COLMAP \cite{schoenberger2016sfm} without the background. We measure two quantities: the number of reconstructed points and the median reprojection error.



Table \ref{tab:obj_points} compares the numbers of reconstructed points and the median reprojection error. In the results of TBSfM, only the points from the foreground are assumed. In 6 out of 9 cases the model reconstructed with TBSfM has more points than the model reconstructed without the background, including model "Lego", which the state-of-the-art pipeline COLMAP did not manage to reconstruct without the background. Table \ref{tab:obj_points} shows that TBSfM achieved lower median reprojection error for all considered models.

\begin{table}
\begin{center}
\begin{tabular}{|c|c|c||c|c|}
\hline
Dataset & \multicolumn{2}{c}{Points} \vline & \multicolumn{2}{c}{Error} \vline\\
\hline
& TBSfM & COLMAP & TBSfM & COLMAP \\
\hline\hline
Colosseum & \textbf{52397} & 49418 & \textbf{1.206} & 1.483\\
\hline
Vatican & \textbf{39024} & 38833 & \textbf{1.418} & 1.792\\
\hline
Ganesha & \textbf{35179} & 34539 & \textbf{1.142} & 1.852\\
\hline
Salt lamp & \textbf{15305} & 10049 & \textbf{1.298} & 1.889\\
\hline
Pillow & 134086 & \textbf{198728} & \textbf{1.147} & 2.342\\
\hline
Car & \textbf{25229} & 14372 & \textbf{1.463} & 2.644\\
\hline
Ship & 18244 & \textbf{20571} & \textbf{1.326} & 1.739\\
\hline
Catalog & 8004 & \textbf{10143} & \textbf{2.350} & 2.832\\
\hline
Lego & \textbf{7692} & 0 & \textbf{1.818} & -\\
\hline
\end{tabular}
\end{center}
\caption{Comparison of the number of points and the median reprojection error between TBSfM and COLMAP.}
\label{tab:obj_points}
\end{table}

\begin{figure}[H]
\begin{center}
\begin{tabular}{c|c}
\includegraphics[width=0.35\linewidth]{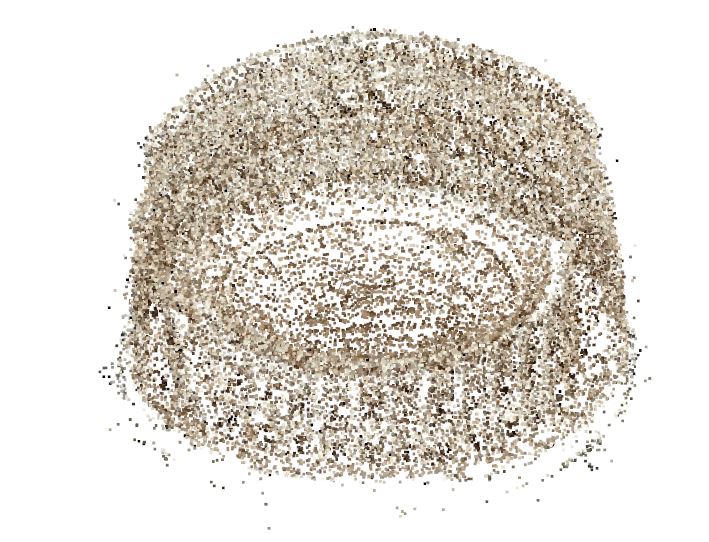}&  
\includegraphics[width=0.4\linewidth]{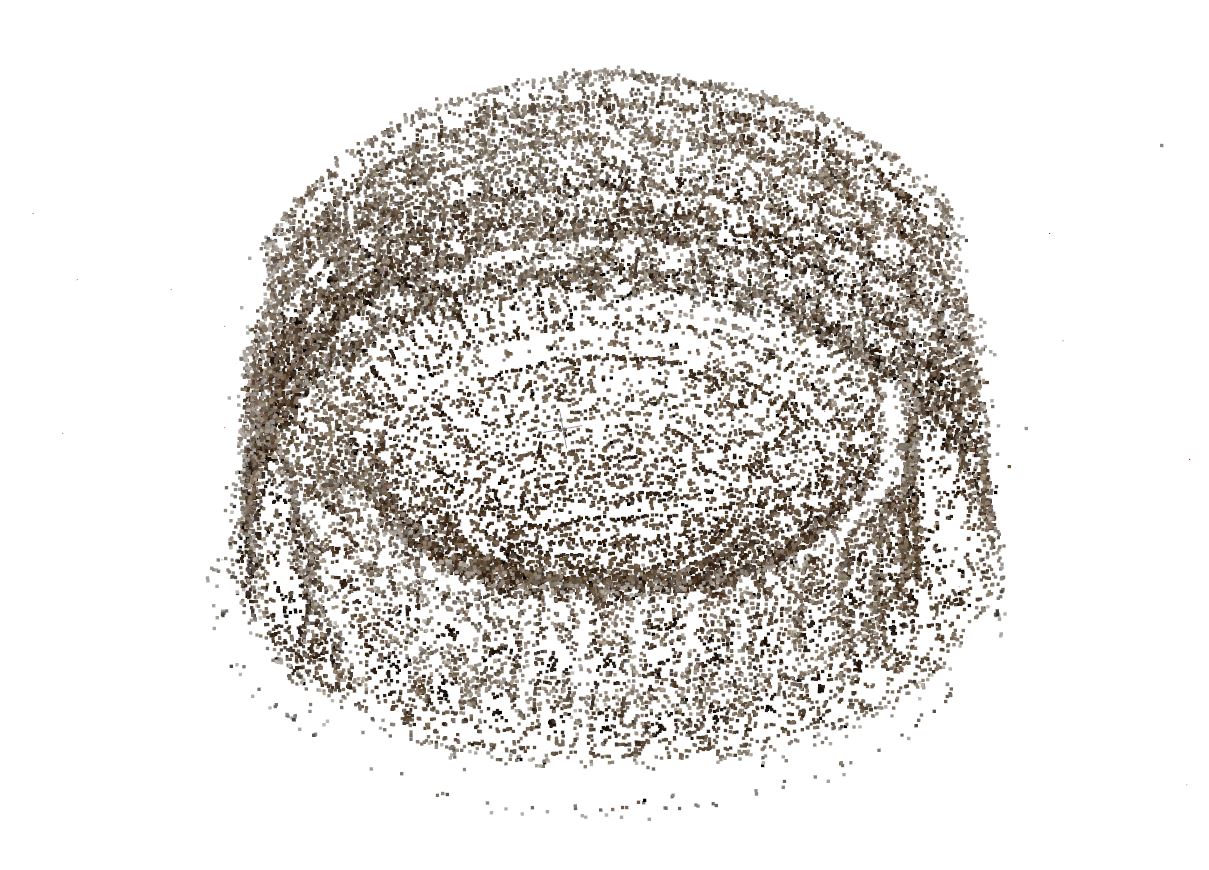}\\
(a) & (b)\\
\includegraphics[width=0.38\linewidth]{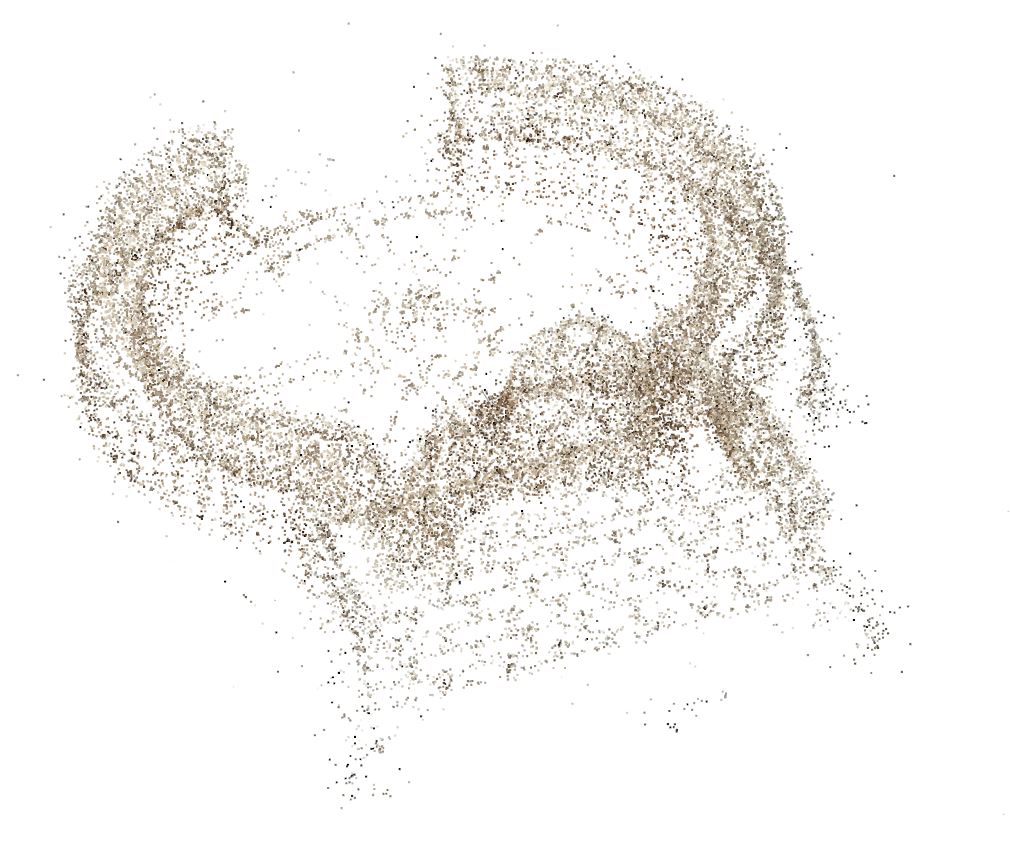}&
\includegraphics[width=0.40\linewidth]{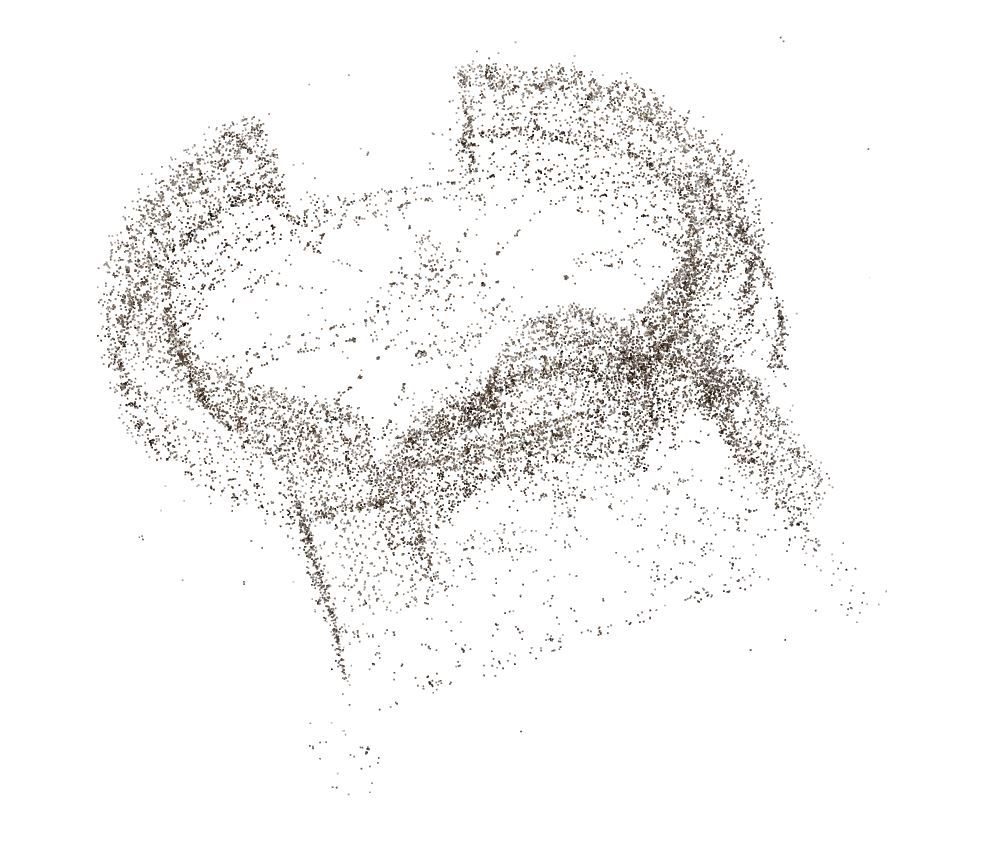}\\
(c) & (d)\\
\includegraphics[width=0.40\linewidth]{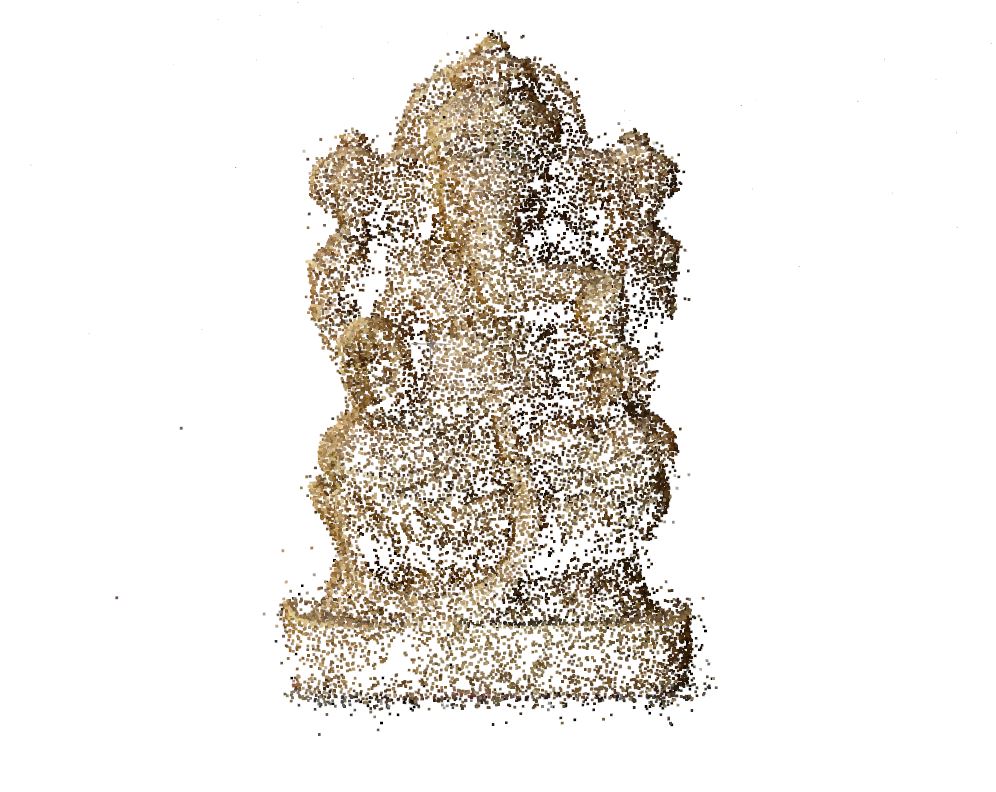}&
\includegraphics[width=0.35\linewidth]{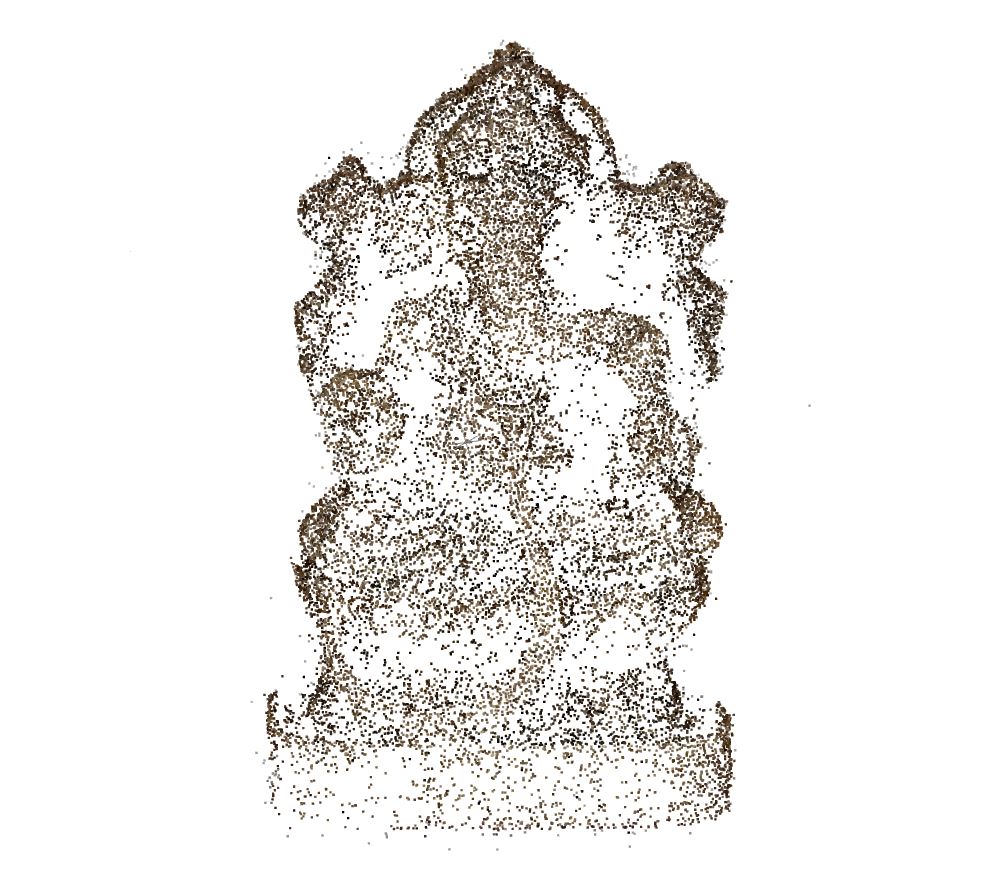}\\
(e) & (f)\\
\end{tabular}
\end{center}
\caption{Comparison of the models reconstructed with and without the background. The models on the left side have been reconstructed with the background.}
\label{fig:compare1}
\end{figure}

\begin{figure}[H]
\begin{center}
\begin{tabular}{c|c}
\includegraphics[width=0.35\linewidth]{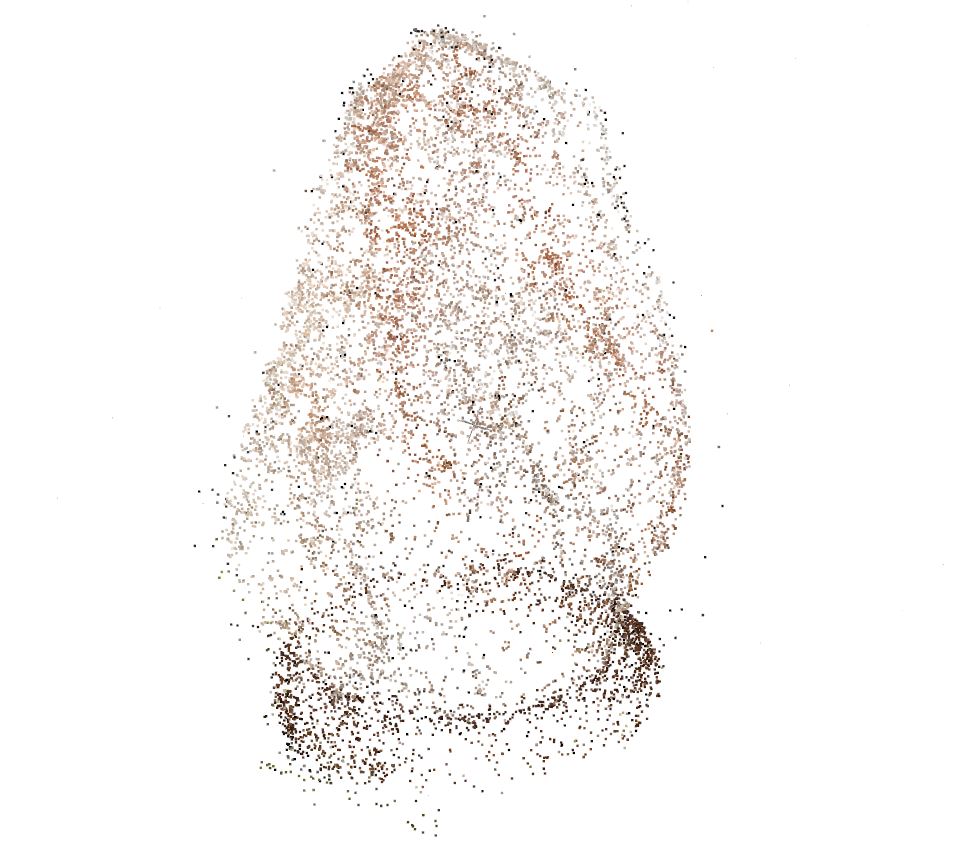}&
\includegraphics[width=0.35\linewidth]{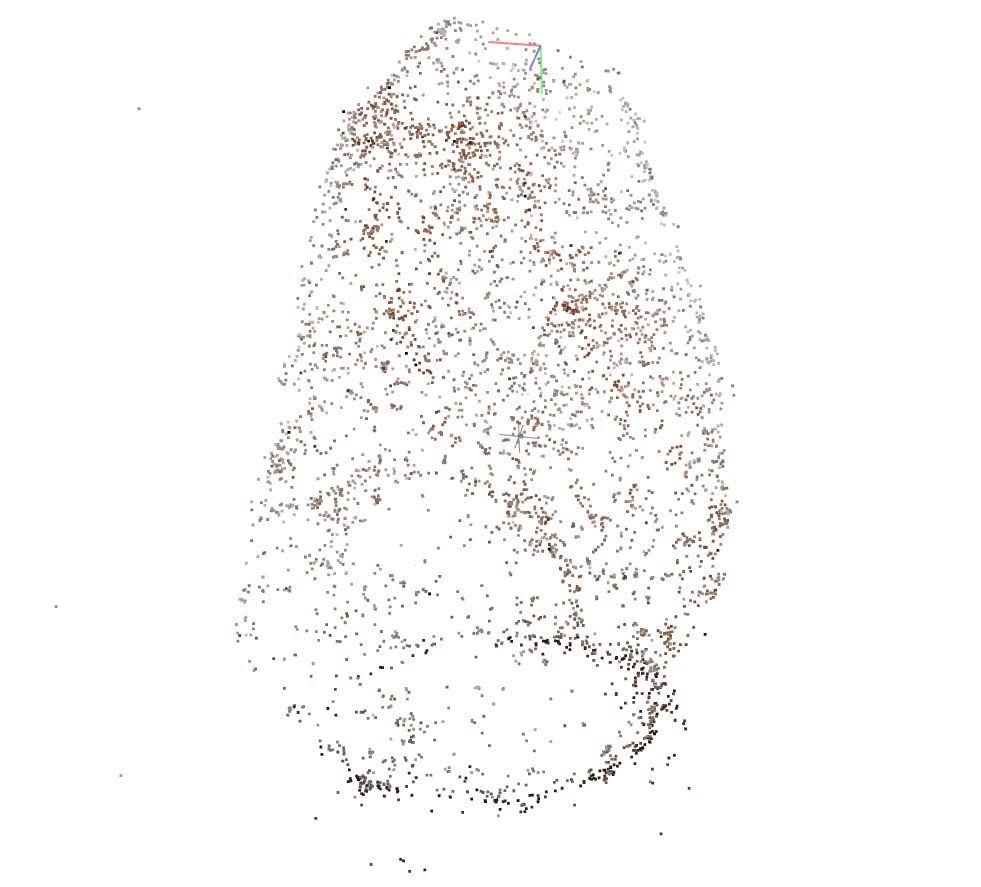}\\
(a) & (b)\\
\includegraphics[width=0.35\linewidth]{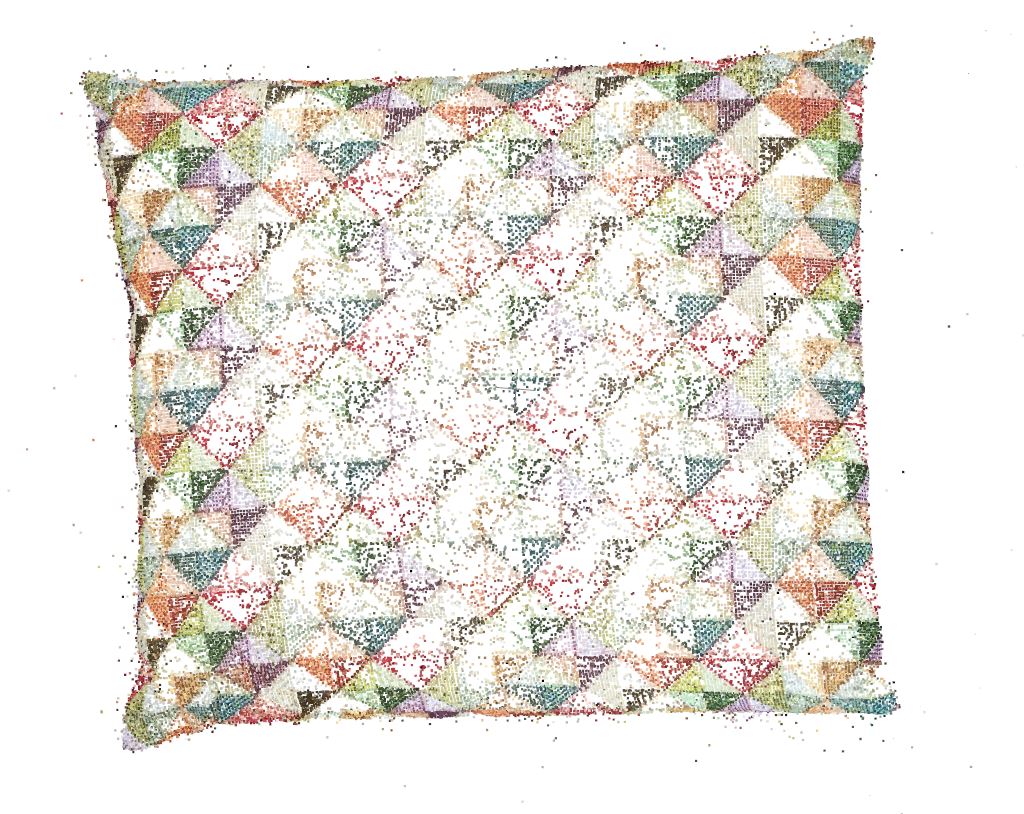}&
\includegraphics[width=0.40\linewidth]{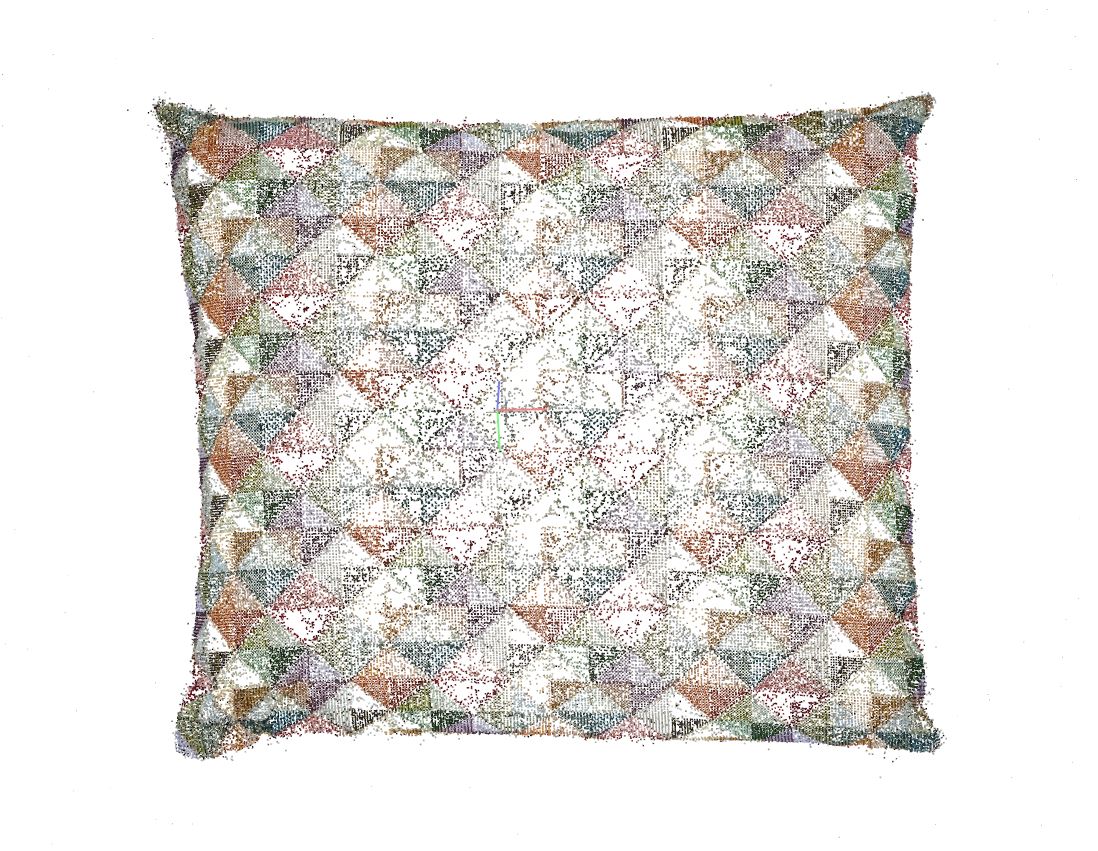}\\
(c) & (d)\\
\includegraphics[width=0.35\linewidth]{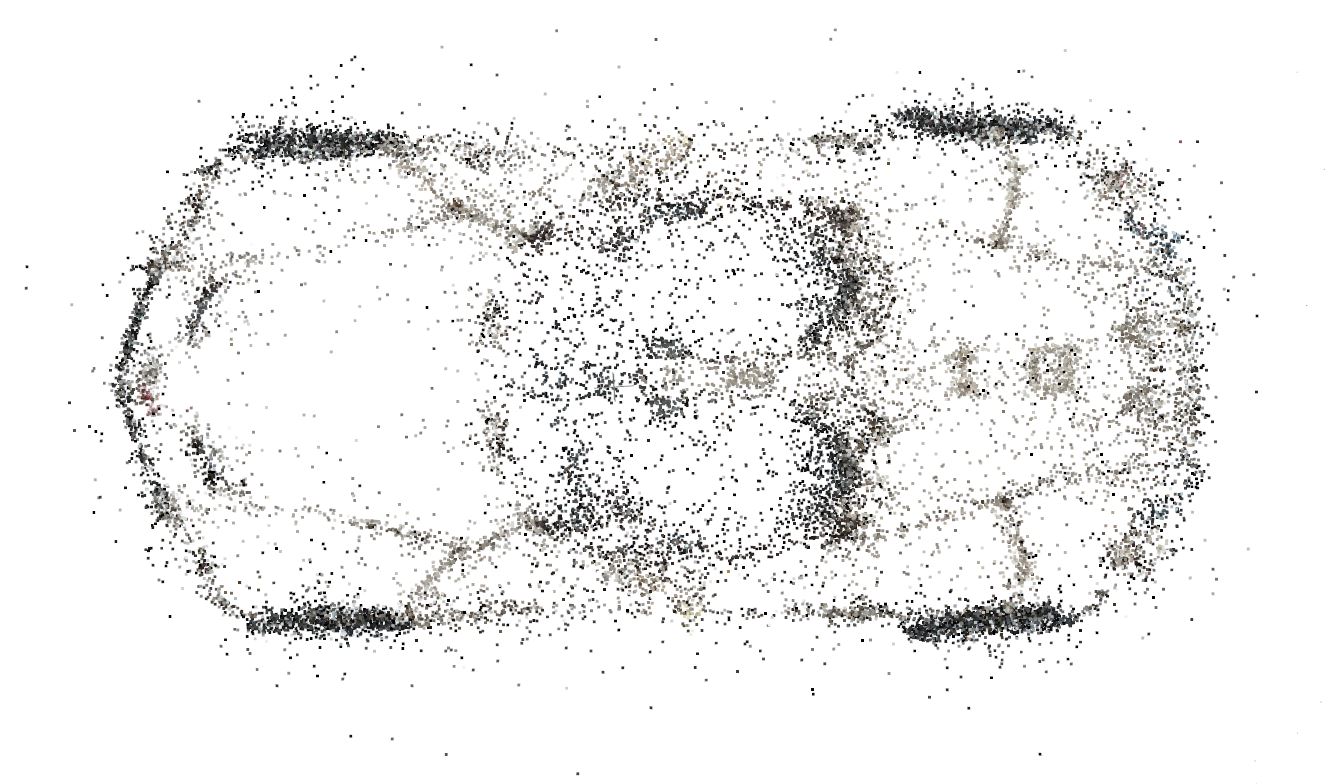}&
\includegraphics[width=0.35\linewidth]{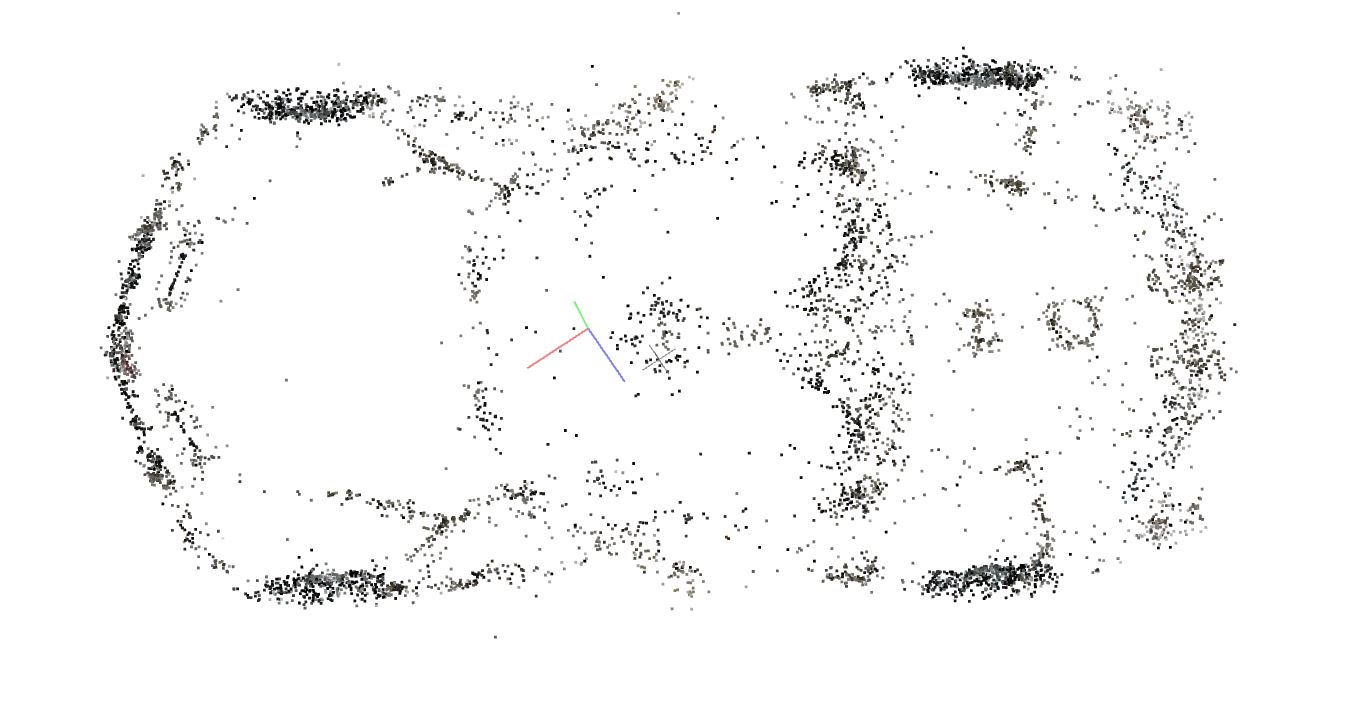}\\
\includegraphics[width=0.35\linewidth]{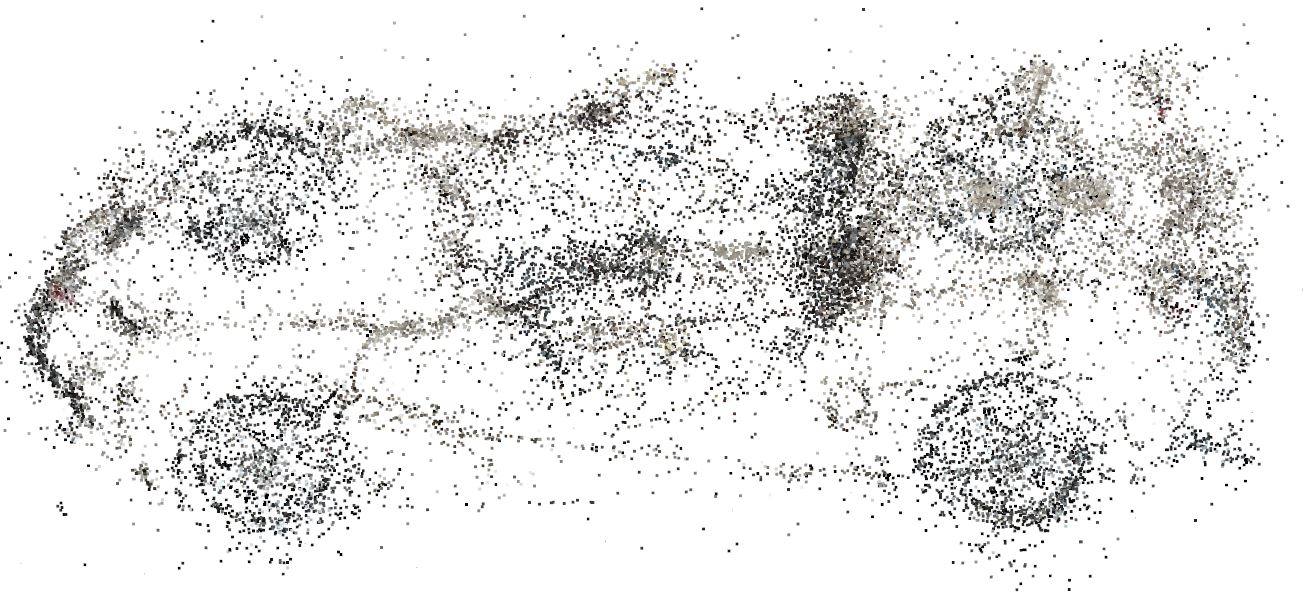}&
\includegraphics[width=0.35\linewidth]{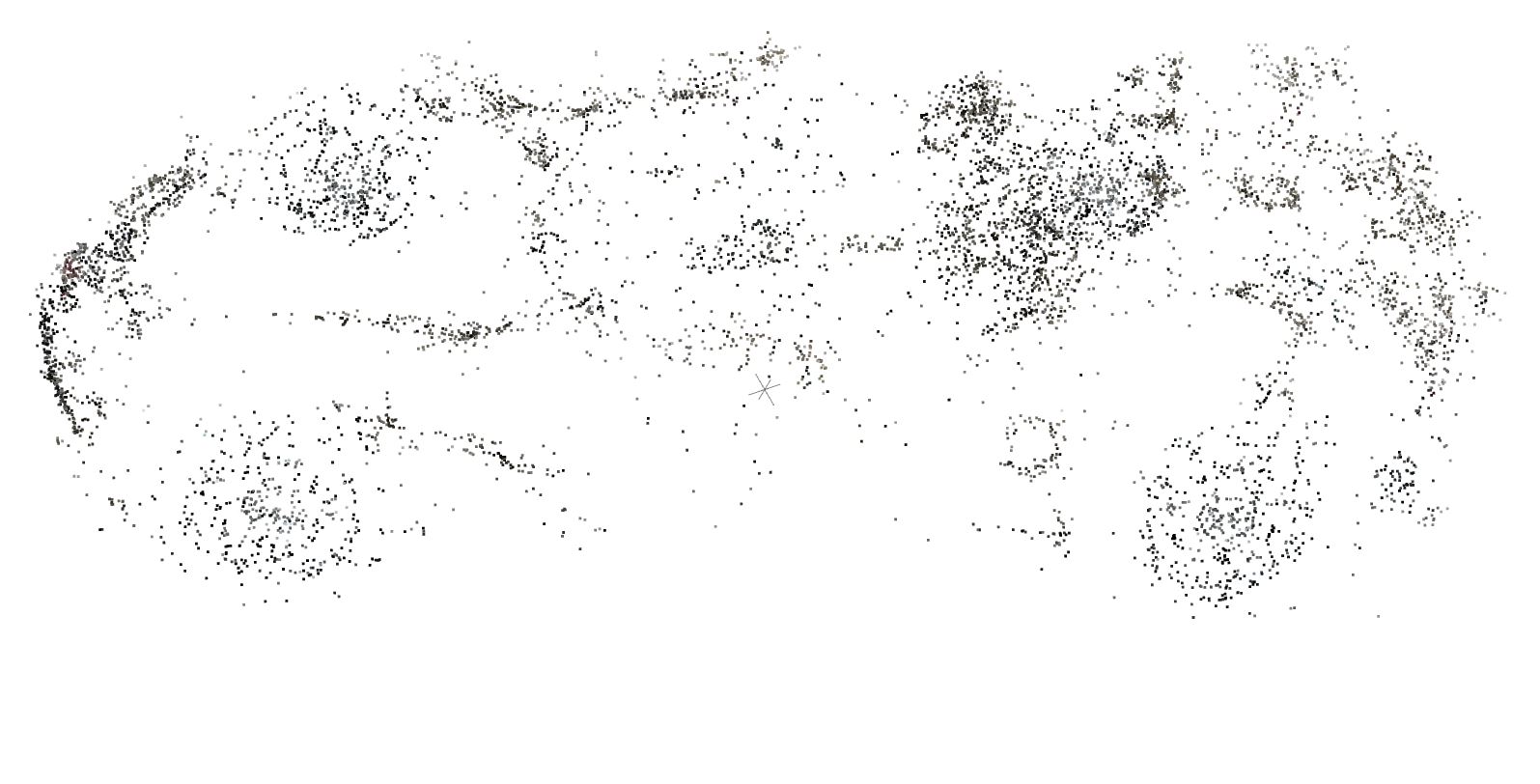}\\
(e) & (f)\\
\end{tabular}
\end{center}
\caption{Comparison of the models reconstructed with and without the background. The models on the left side have been reconstructed with the background.}
\label{fig:compare2}
\end{figure}

\begin{figure}[H]
\begin{center}
\begin{tabular}{c|c}
\includegraphics[width=0.39\linewidth]{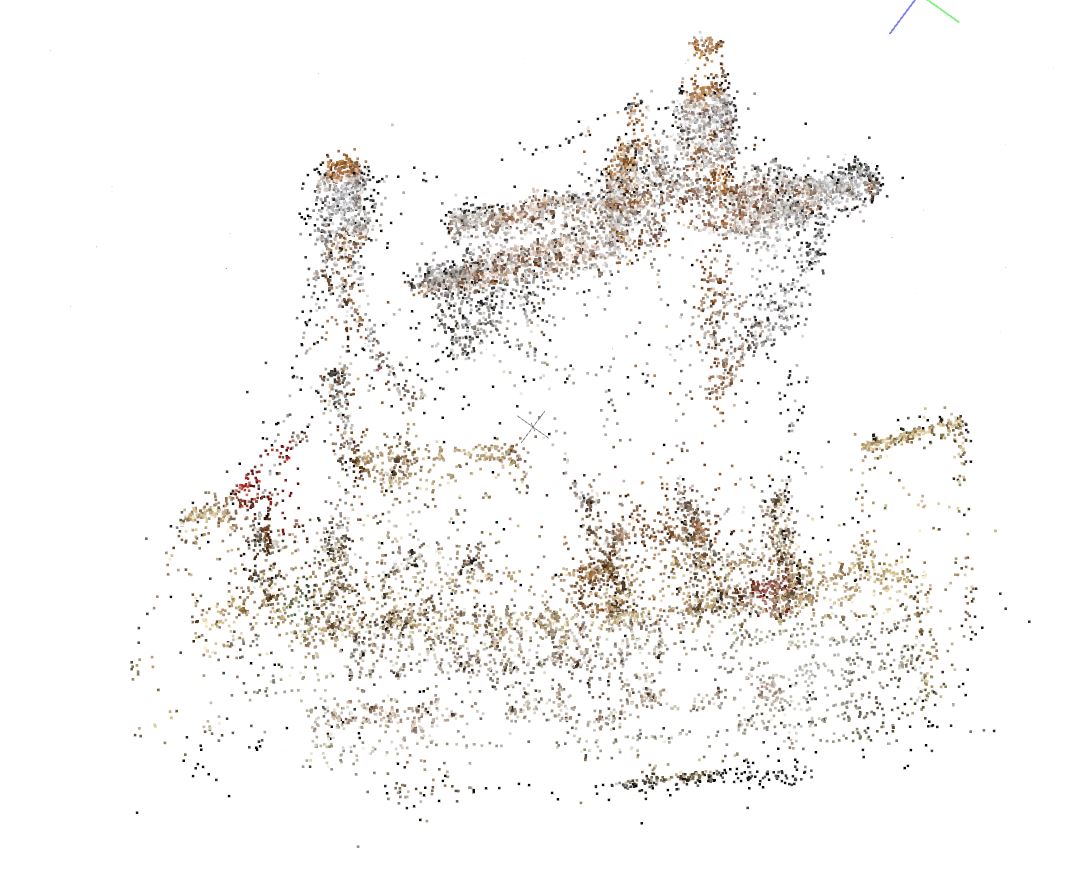}&
\includegraphics[width=0.35\linewidth]{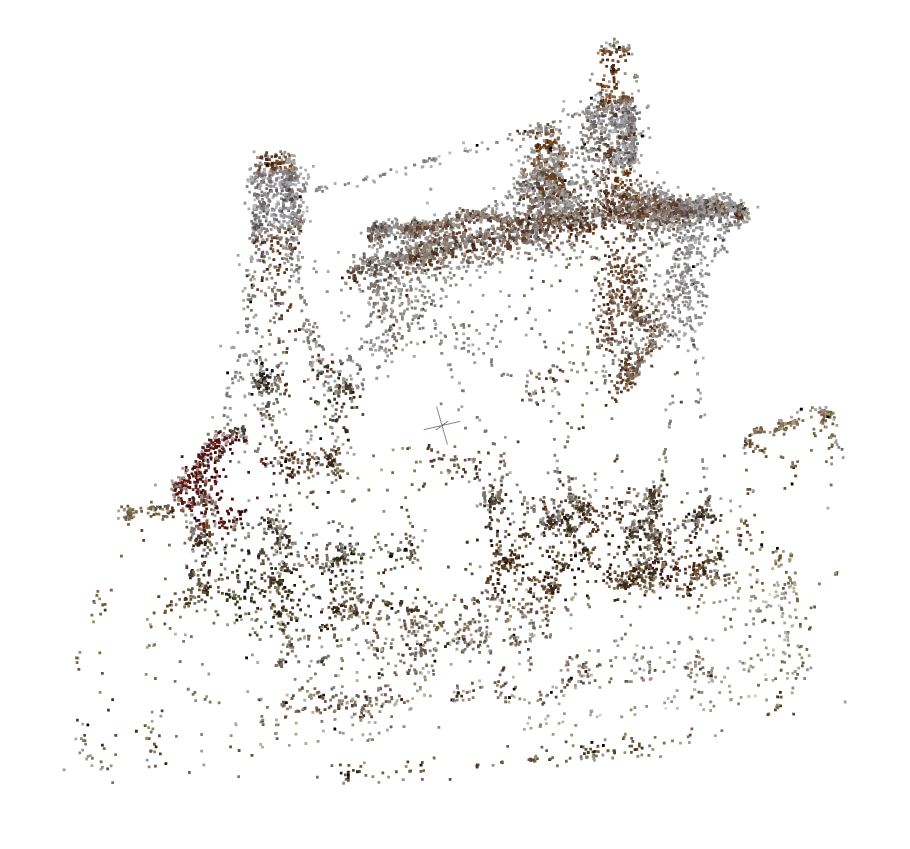}\\
(a) & (b)\\
\includegraphics[width=0.35\linewidth]{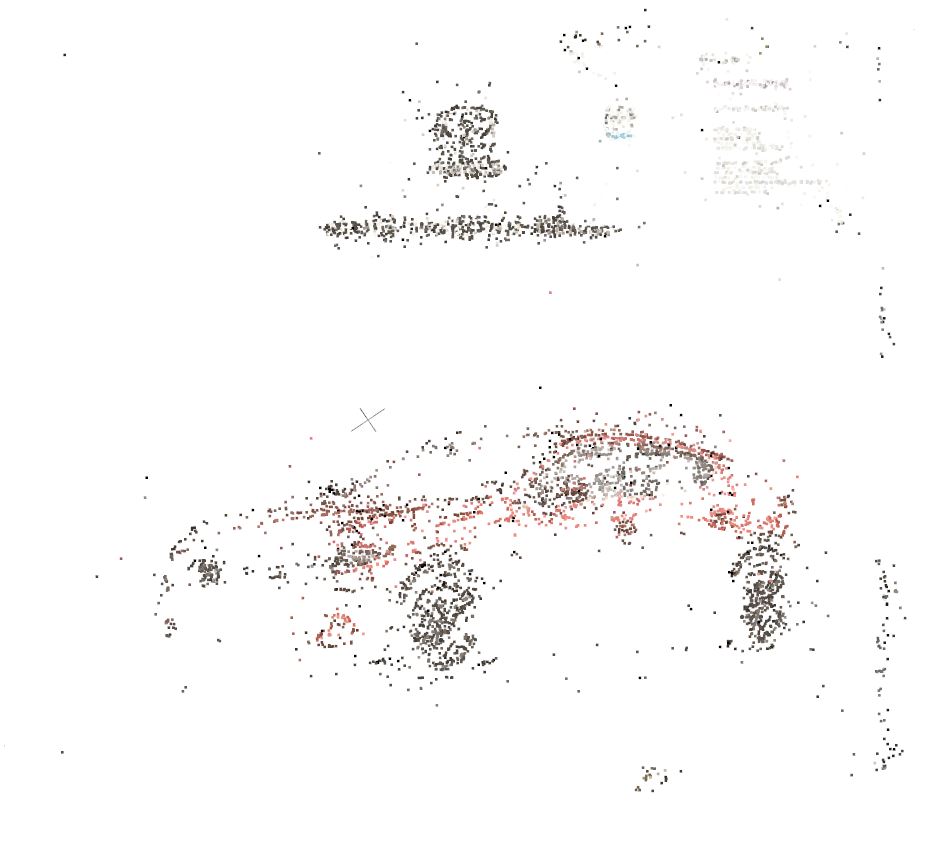}&
\includegraphics[width=0.35\linewidth]{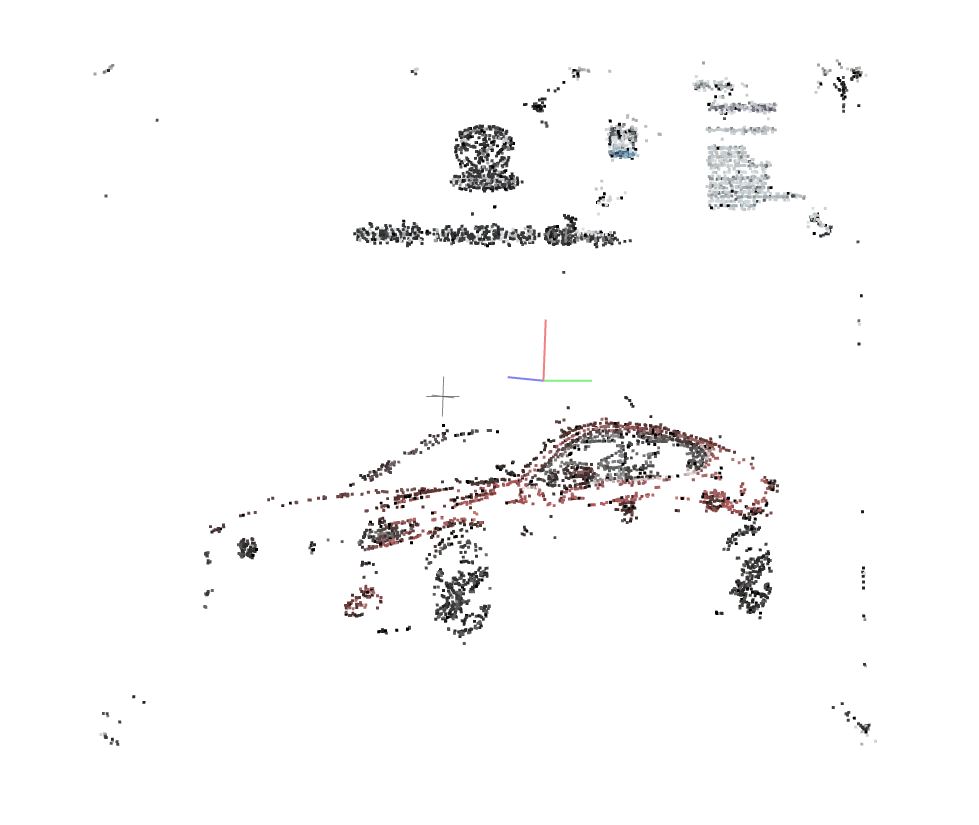}\\
(c) & (d)\\
\\
\includegraphics[width=0.30\linewidth]{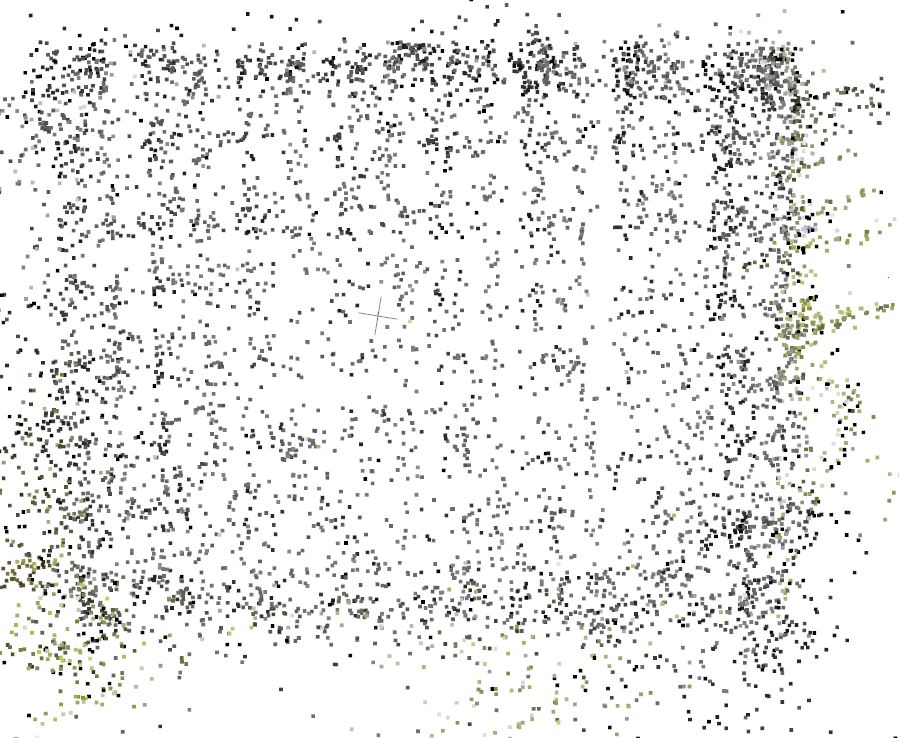}&
\includegraphics[width=0.35\linewidth]{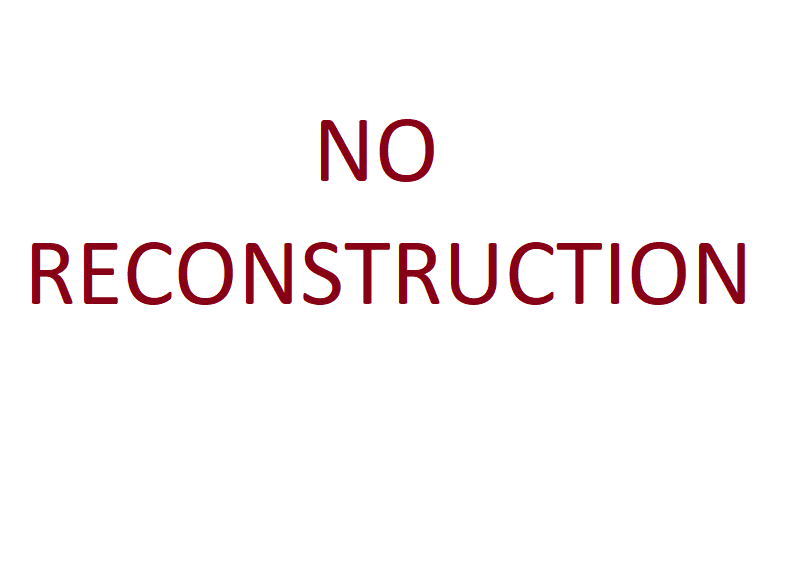}\\
(e) & (f)\\
\end{tabular}
\end{center}
\caption{Comparison of the models reconstructed with and without the background. The models on the left side have been reconstructed with the background.}
\label{fig:compare3}
\end{figure}

Figures \ref{fig:compare1}, \ref{fig:compare2}, \ref{fig:compare3} compare the models of the objects reconstructed with TBSfM and with COLMAP. In each row, the image on the left depicts a model of the foreground object reconstructed using TBSfM, while the image on the right depicts model of the same object reconstructed in COLMAP without background. TBSfM reconstructed significantly denser models of objects "Ganesha" (Fig.~\ref{fig:compare1}(e)), "Lamp" (Fig.~\ref{fig:compare2}(a)) and "Car" (Fig.~\ref{fig:compare2}(e)). Also notice, that TBSfM managed to reconstruct some parts of objects, which COLMAP could not reconstruct, like the wall on the back side of the "Vatican" object (Fig.~\ref{fig:compare1}(c)) or the base of the "Lamp" object (Fig.~\ref{fig:compare2}(a)). In addition to that, TBSfM managed to reconstruct object "Lego" (Fig.~\ref{fig:compare3}(e)), which could not be reconstructed by COLMAP without the background due to its planar and repetitive shape. On the other hand, COLMAP reconstructed better models of objects "Ship" (Fig.~\ref{fig:compare3}(a)) and "Catalog" (Fig.~\ref{fig:compare3}(c)).

\subsection{Results of YASFM}\label{sec:yasfm}
\begin{figure}[H]
    \begin{center}
        \begin{tabular}{c c}
            \includegraphics[width=0.33\linewidth]{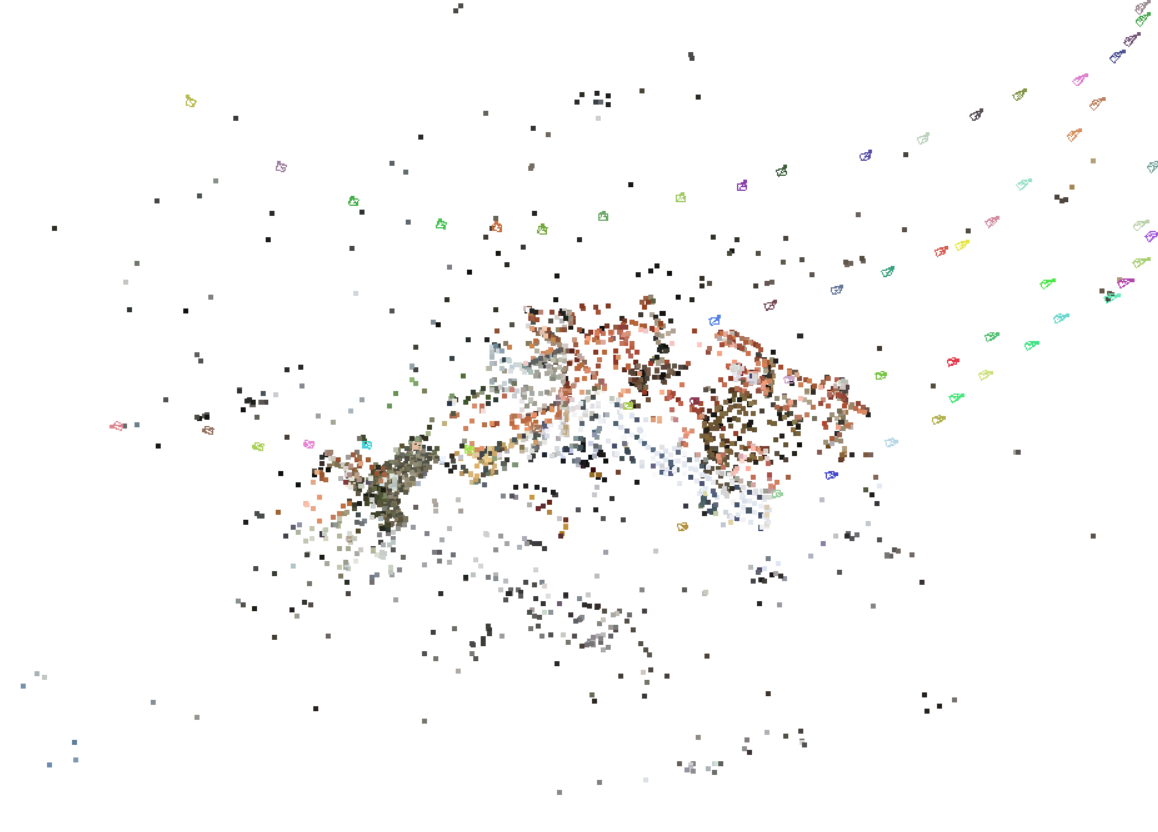}&
            \includegraphics[width=0.33\linewidth]{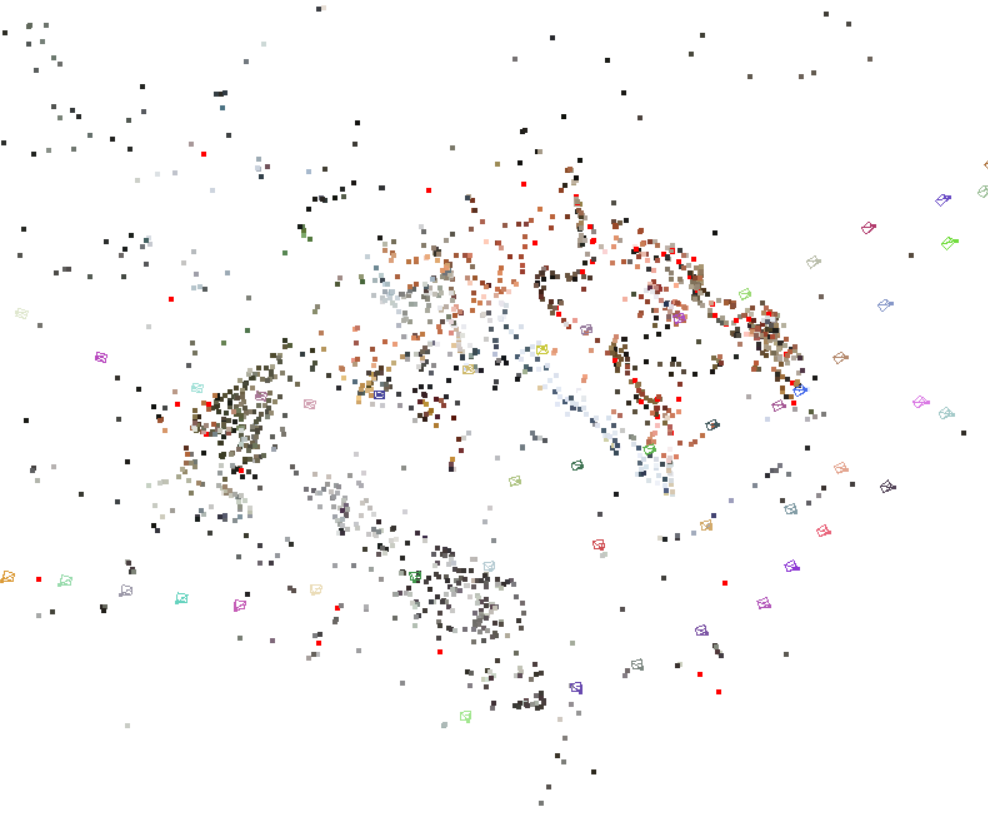}\\
            (a) & (b)\\
            \includegraphics[width=0.33\linewidth]{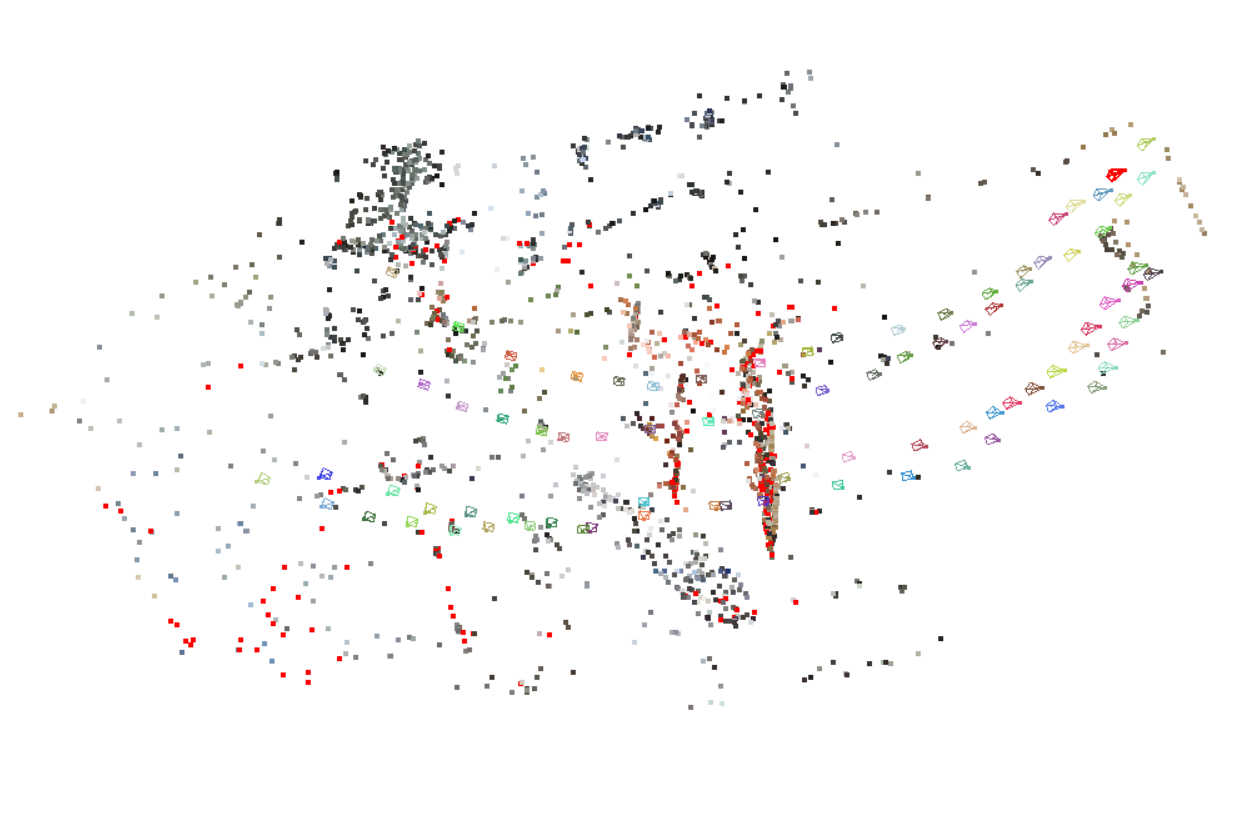}&
            \includegraphics[width=0.33\linewidth]{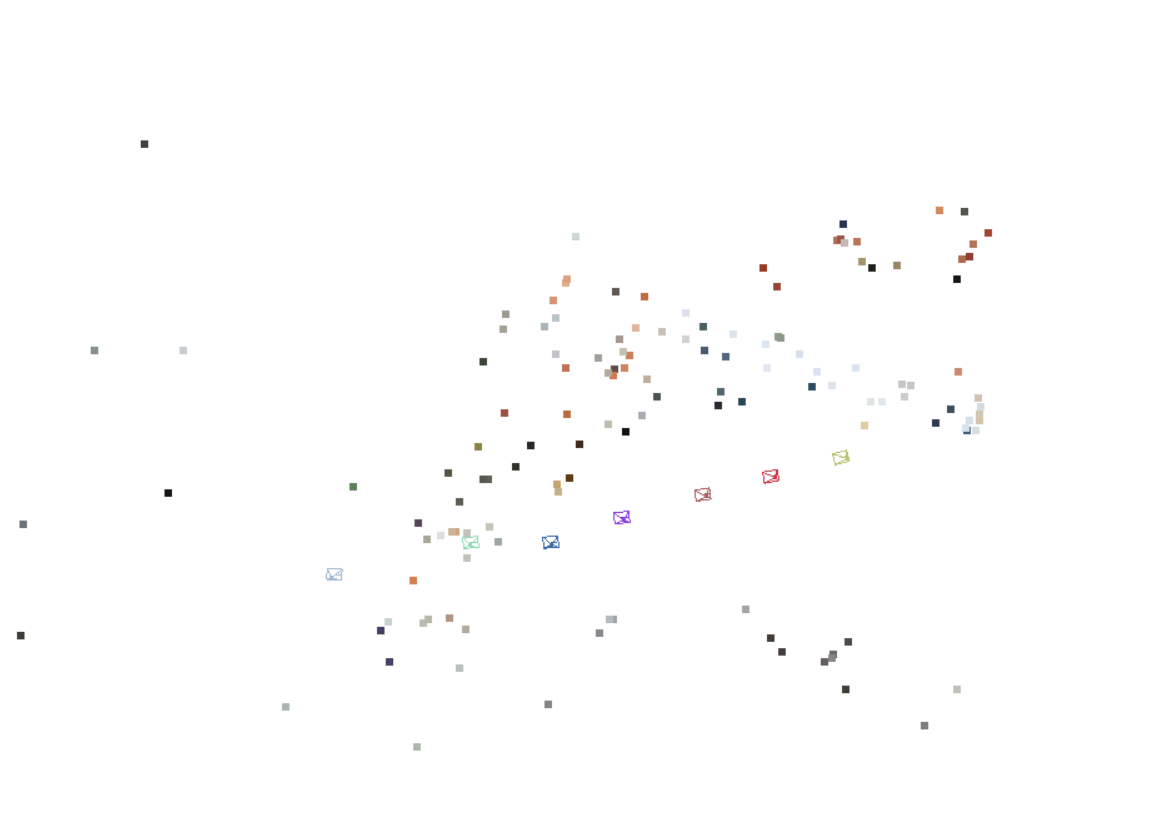}\\
            (c) & (d)\\
            \includegraphics[width=0.32\linewidth]{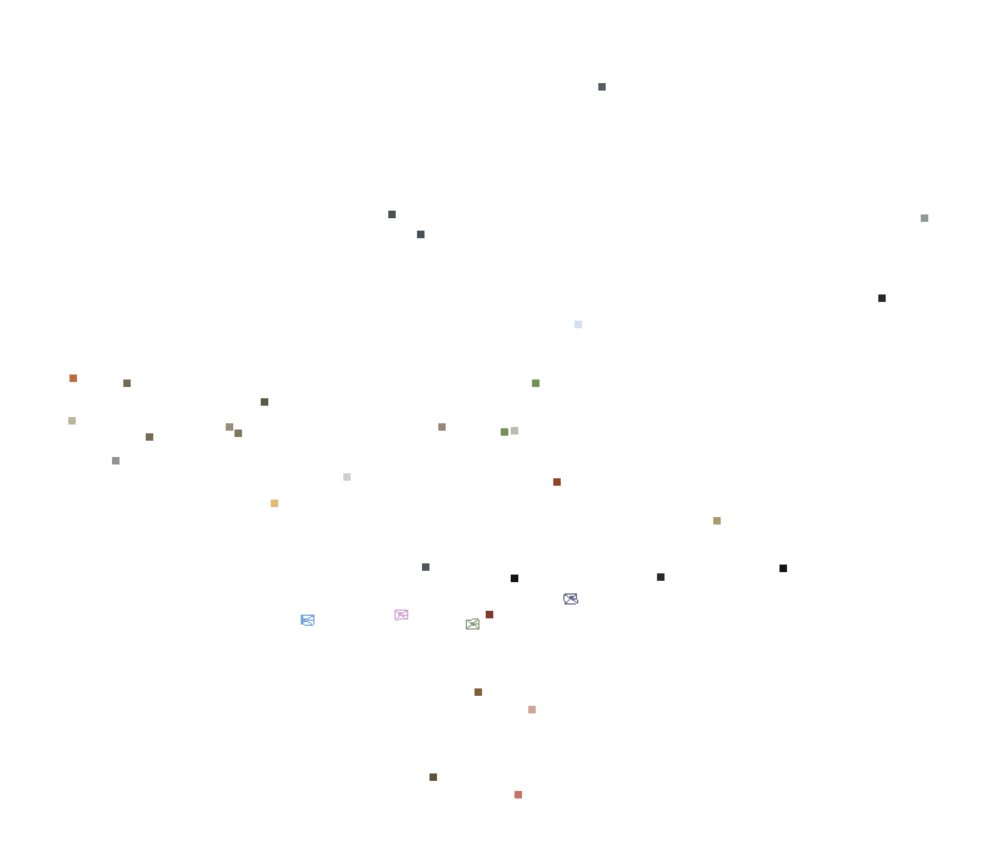}\\
            (e)
        \end{tabular}
    \end{center}
    \caption{Models reconstructed from Daliborka dataset in YASFM. The reconstruction has split into five models.}
\label{fig:daliborka_srajer}
\end{figure}

In this section we give an example of the result, which outputs a MBSfM pipeline YASFM \cite{Srajer16}, if it is given the images from our dataset. Figure \ref{fig:daliborka_srajer} shows that YASFM typically splits the model into more than two objects, most of which contain points from both the foreground and the background. For example the reconstruction of the "Daliborka" object was split into five models, which are shown in Figure \ref{fig:daliborka_srajer}. Figures \ref{fig:daliborka_srajer}(a), \ref{fig:daliborka_srajer}(b), \ref{fig:daliborka_srajer}(c) show models which contain points from both the foreground and the background. Figures \ref{fig:daliborka_srajer}(d), \ref{fig:daliborka_srajer}(e) show very sparse models. The result of our method on this dataset is shown in Figure \ref{fig:recon1}(b).

\subsection{Results of MODE}\label{sec:mode}
\begin{figure}[H]
    \begin{center}
        \begin{tabular}{c c}
            \includegraphics[width=0.45\linewidth]{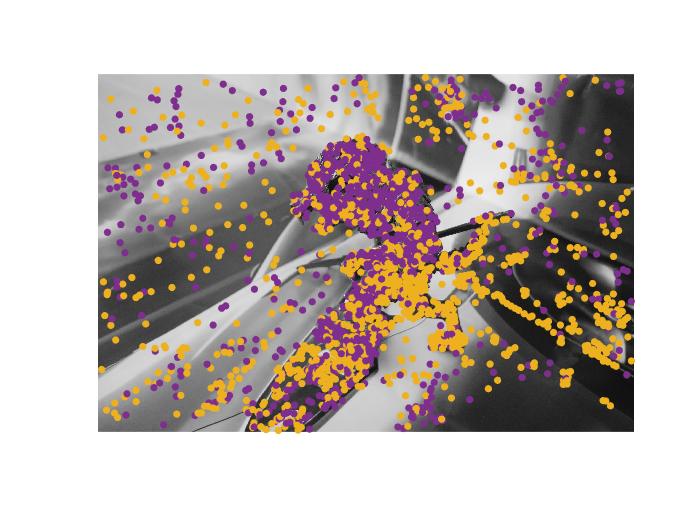}&
            \includegraphics[width=0.45\linewidth]{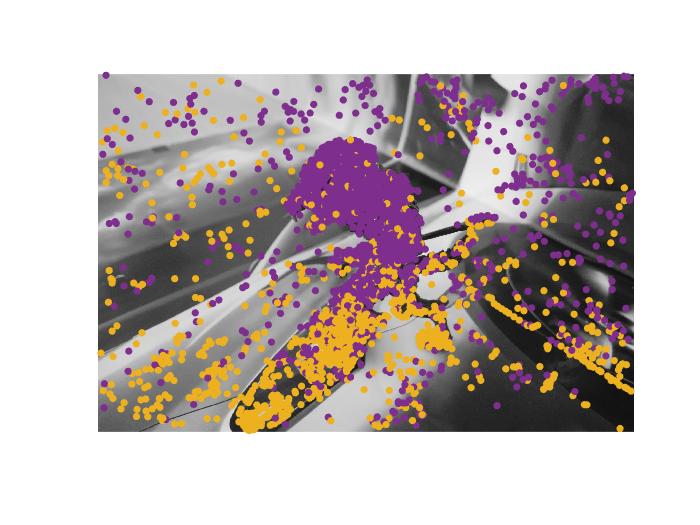}\\
            (a) & (b)\\
            \includegraphics[width=0.45\linewidth]{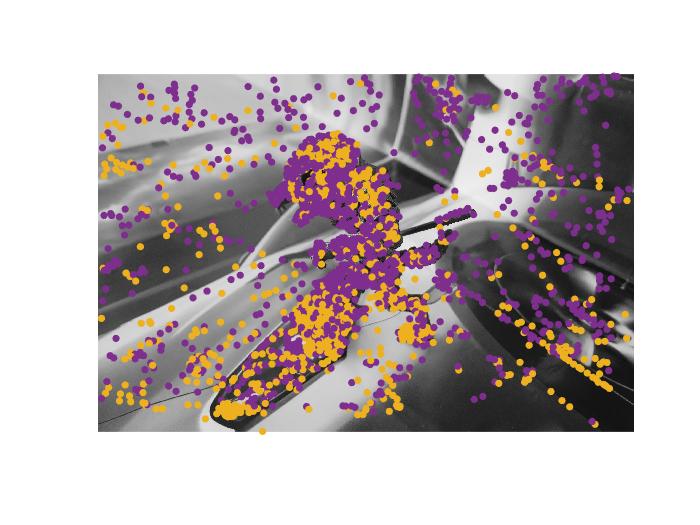}&
            \includegraphics[width=0.45\linewidth]{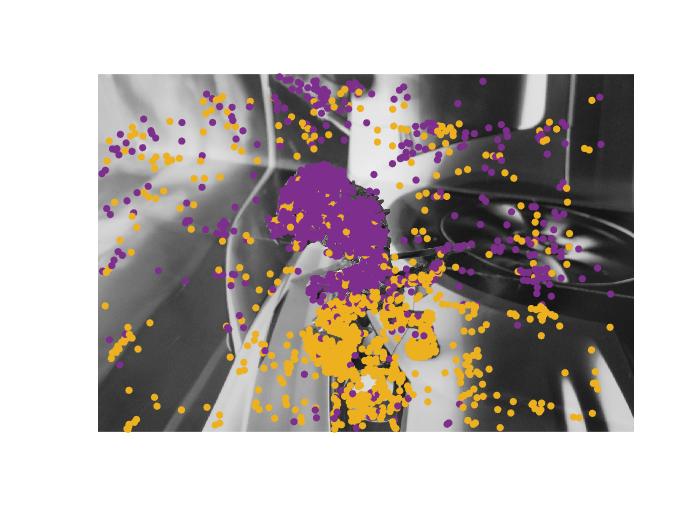}\\
            (c) & (d)\\
        \end{tabular}
    \end{center}
    \caption{Result of motion segmentation by MODE. Points belonging to the first object are purple and points belonging to the second object are yellow.}
\label{fig:mode}
\end{figure}

\begin{figure}[H]
    \begin{center}
            \includegraphics[width=0.5\linewidth]{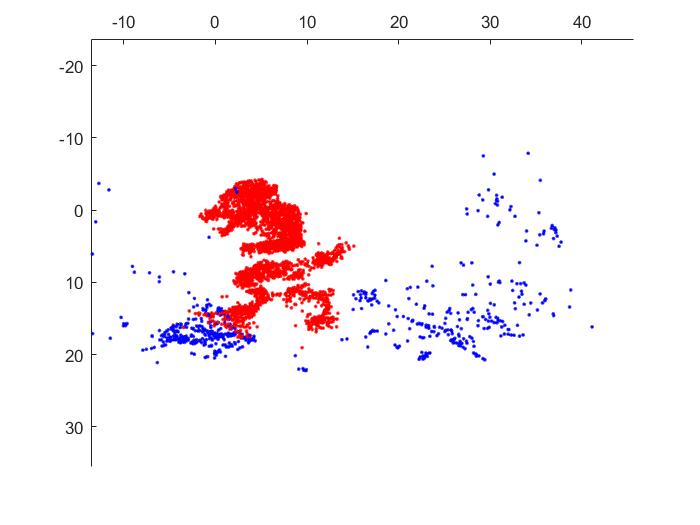}
    \end{center}
    \caption{Result of our method on the shortened version of Lycan dataset. Although the total number of images is equal to 7, our method gives a meaningful result.}
\label{fig:shortened}
\end{figure}

In this section we give an example of the result of the segmentation, which outputs MODE \cite{DBLP:journals/corr/abs-1905-09043}, if it is given the images from our dataset. If the whole datasets are used as the input, MODE runs out of memory. We have, therefore, shortened our dataset to 7 images, which were used as the input for the method. Figure \ref{fig:mode} shows that the result of the segmentation is not correct and that both clusters contain points from the foreground object, as well as from the background. Our method is able to correctly segment and reconstruct the shortened version of the dataset, the result of our method is shown in Figure \ref{fig:shortened}.

\subsection{Results of Subset}\label{sec:subset}

\begin{figure}[H]
    \begin{center}
        \begin{tabular}{c c}
            \includegraphics[width=0.4\linewidth]{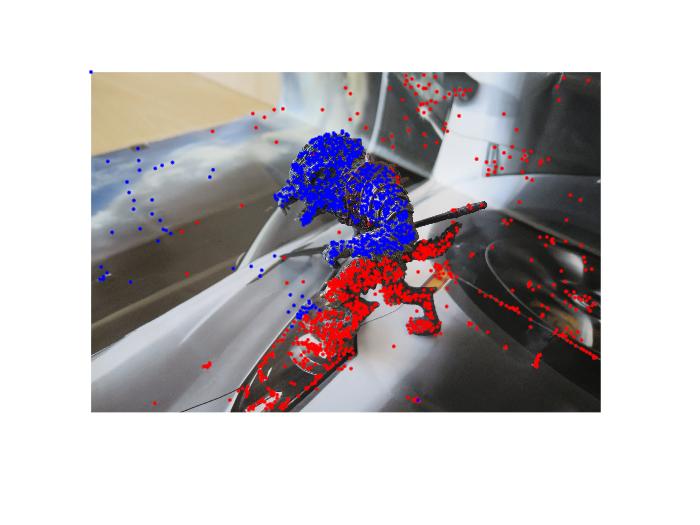}&
            \includegraphics[width=0.4\linewidth]{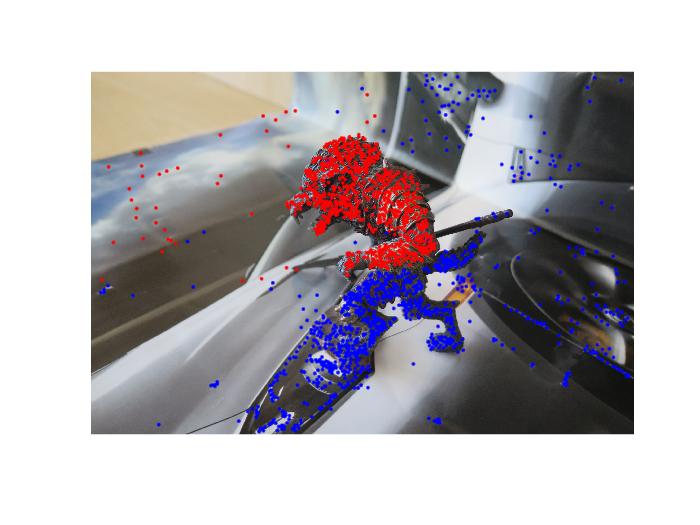}\\
            (a) & (b)\\
            \includegraphics[width=0.4\linewidth]{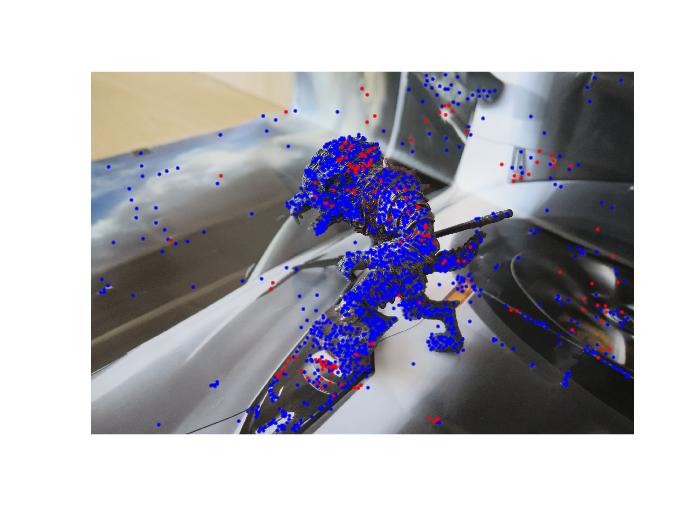}&
            \includegraphics[width=0.4\linewidth]{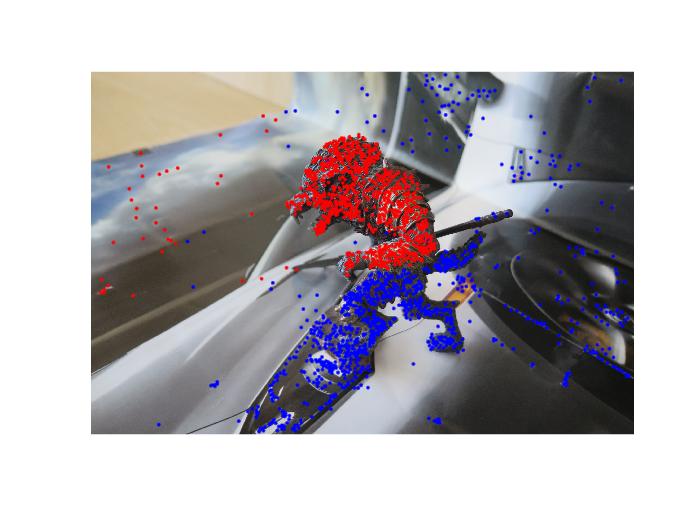}\\
            (c) & (d)\\
            \includegraphics[width=0.4\linewidth]{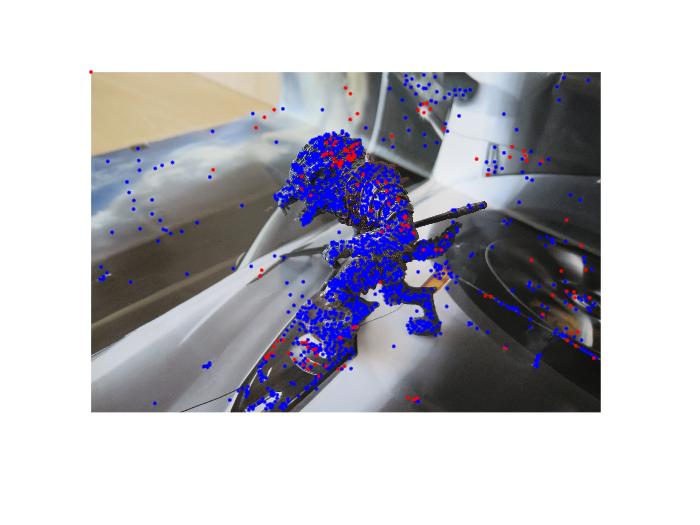}&
            \includegraphics[width=0.4\linewidth]{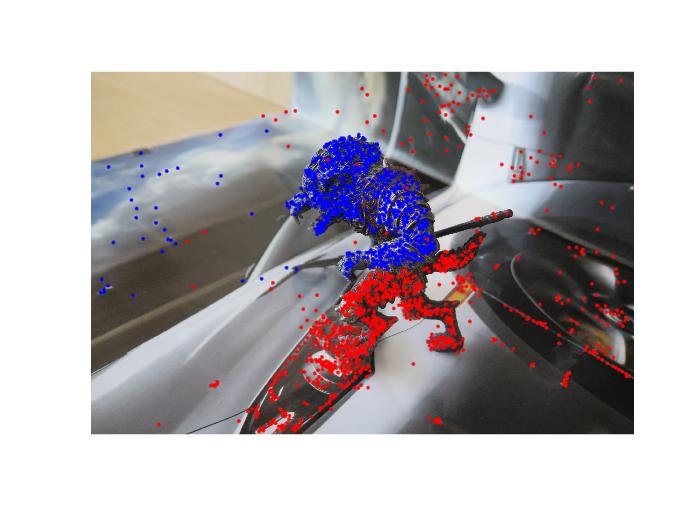}\\
            (e) & (f)\\
        \end{tabular}
    \end{center}
    \caption{Result of motion segmentation by Subset. Points belonging to the first object are blue and points belonging to the second object are red. Particular images are results of (a) Affine, (b) Homography, (c) Fundamental, (d) Kernel Addition, (e) Co-Regularization, (f) Subset Constrained Multi-View Spectral Clustering.}
\label{fig:subset}
\end{figure}

In this section we give an example of the result of the segmentation, which outputs Subset \cite{Xu18} if it is given the images from our dataset. If the whole datasets are used as the input, Subset runs out of memory. Therefore, we have shortened our dataset to 7 images, which were used as the input for the method. Paper \cite{Xu18} describes six different spectral clustering schemes to fuse the clustering information: (a) Affine, (b) Homography, (c) Fundamental, (d) Kernel Addition, (e) Co-Regularization, (f) Subset Constrained Multi-View Spectral Clustering. All the schemes assign all the points to one object. If we enforce 2 objects, all spectral clustering schemes give incorrect segmentation results, which are shown in Figure \ref{fig:subset}. Our method is able to correctly segment and reconstruct the shortened version of the dataset, the result of our method is shown in Fig.~\ref{fig:shortened}.

\subsection{Results of COLMAP}\label{sec:colmap}
\begin{figure}[H]
\begin{center}

\includegraphics[width=0.9\linewidth]{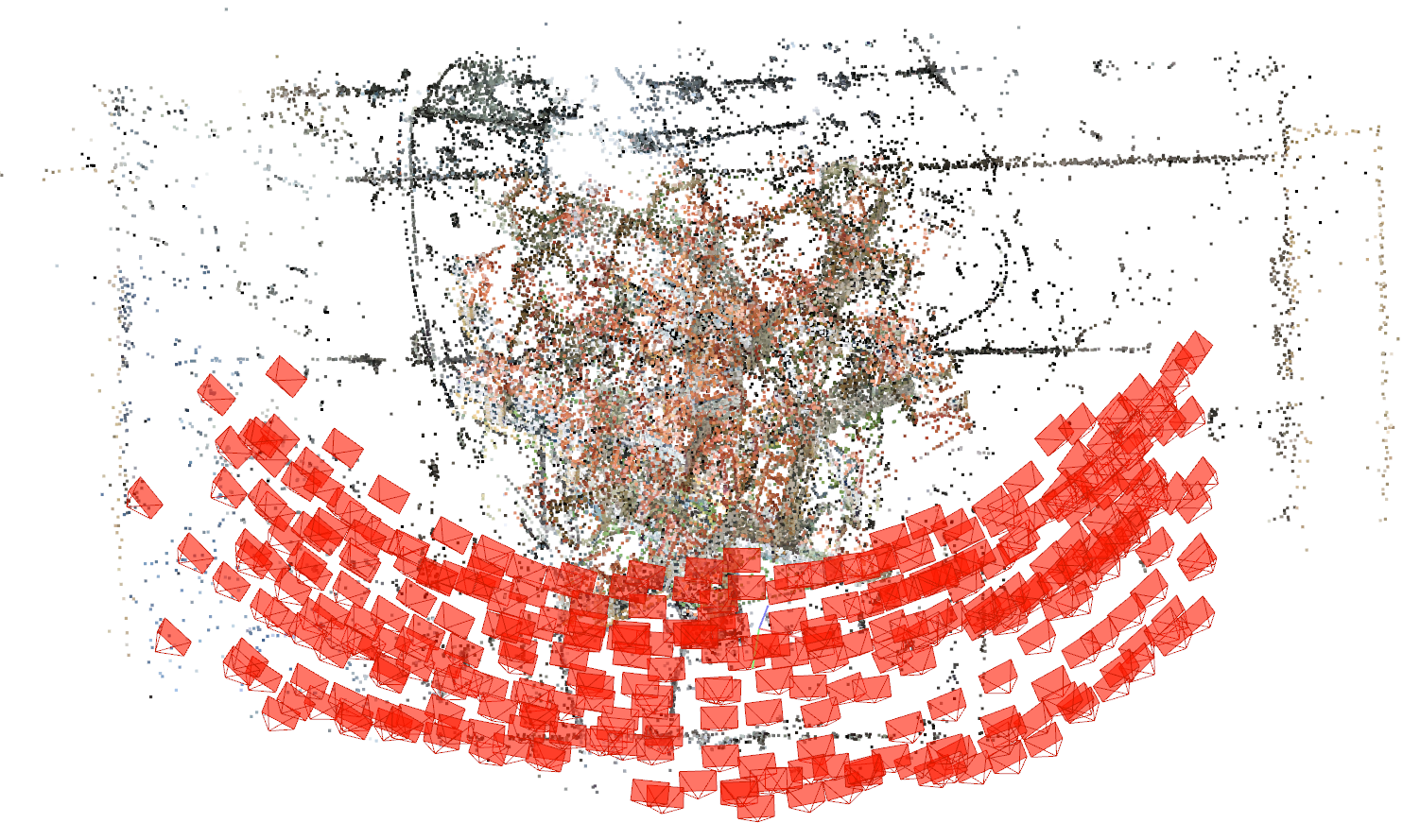}\\
\caption{Model reconstructed from the whole Daliborka dataset in COLMAP.}
\label{fig:colmap3}
\end{center}
\end{figure}

In this section we show the result, which outputs the single body SfM pipeline COLMAP \cite{schoenberger2016sfm}, if it is given images from different takes. COLMAP is not able to segment the points from different objects. The section is related to Sec.~4.2 of the main paper. We see in Figure \ref{fig:colmap3} that the resulting model contains one background, on which the foreground object is reconstructed multiple times on different positions, once for each take.

\subsection{Experiments with ETH 3D dataset}\label{sec:eth}

In this section we show the result of the experiments with the sequences, which have been extracted from the ETH 3D SLAM datasets \cite{DBLP:conf/cvpr/SchopsSP19}. The experiment has been proposed in Sec.~4.4 of the main paper. The sequences have been extracted from datasets motion\_1, motion\_2, motion\_3, motion\_4, which depict a dynamic scene with a synchronized rig of two cameras. Although the objects in the scene perform an independent motion, the objects do not move between any two images taken at the same time by the two cameras of the rig. Therefore, we obtain the semi-dynamic setting, where each pair of images taken at the same time builds one take. We have extracted every sequence in such a way that it contains one object moving with respect to the background. The properties of the sequences are in Table \ref{tab:ETH_sequences}.\\

\begin{table}
\begin{center}
\begin{tabular}{|c|c|c|c|c|c|}
\hline
ID & Dataset & Object & First Image & Last Image & Length\\
\hline\hline
(a) & motion\_1 & Einstein Bust & 7603.638892 & 7607.214650 & 98\\
(b) & motion\_2 & Backpack & 7978.918332 & 7981.867380 & 130\\
(c) & motion\_3 & Brown Bear & 8020.794811 & 8024.628573 & 105\\
(d) & motion\_3 & White Bear & 8005.828393 & 8007.855864 & 56\\
(e) & motion\_3 & Ball & 8012.353162 & 8017.403406 & 138\\
(f) & motion\_3 & Frieze 1 & 8028.019978 & 8032.885907 & 133\\
(g) & motion\_4 & Backpack & 8086.964071 & 8088.991541 & 56\\
(h) & motion\_4 & White Bear & 8075.094154 & 8076.863582 & 49\\
(i) & motion\_4 & Frieze 2 & 8077.674570 & 8081.508333 & 105\\
(j) & motion\_4 & Dinosaur & 8107.607406 & 8110.150959 & 70\\
(k) & motion\_4 & Ball & 8111.478031 & 8114.021584 & 70\\
(l) & motion\_4 & Brown Bear & 8132.637449 & 8138.277503 & 154\\
\hline

\end{tabular}
\end{center}
\caption{Properties of the sequences extracted from ETH 3D dataset, which have been used to evaluate our method.}
\label{tab:ETH_sequences}
\end{table}

\begin{figure}[H]
\begin{center}
\begin{tabular}{c c c}
\includegraphics[width=0.2\linewidth]{Figs/einstein_img.png}&  
\includegraphics[width=0.24\linewidth]{Figs/einstein_front_.jpg}&
\includegraphics[width=0.24\linewidth]{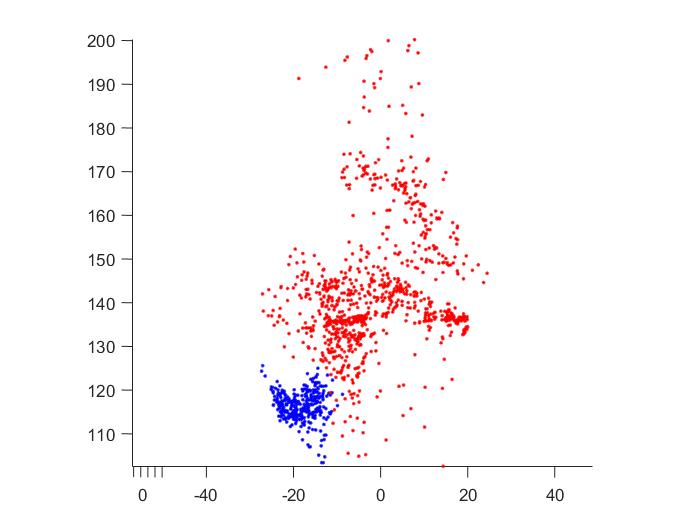}\\
 & (a)\\
\includegraphics[width=0.2\linewidth]{Figs/backpack_img.png}&  
\includegraphics[width=0.24\linewidth]{Figs/backpack_front.jpg}&
\includegraphics[width=0.24\linewidth]{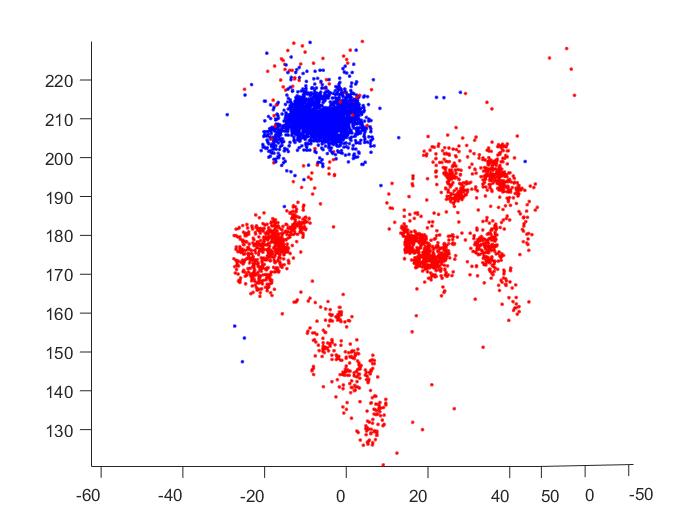}\\
 & (b)\\
\includegraphics[width=0.2\linewidth]{Figs/bear1_img.png}&  
\includegraphics[width=0.24\linewidth]{Figs/bear1_front.jpg}&
\includegraphics[width=0.24\linewidth]{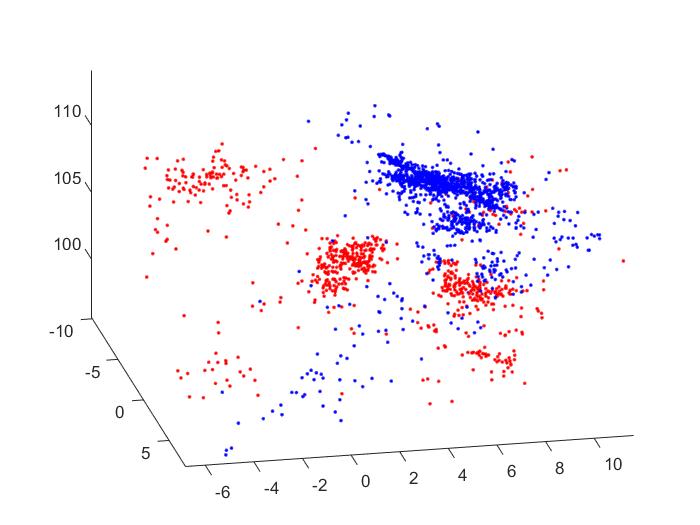}\\
 & (c)\\
\includegraphics[width=0.2\linewidth]{Figs/bear2_img.png}&  
\includegraphics[width=0.24\linewidth]{Figs/bear2_front.jpg}&
\includegraphics[width=0.24\linewidth]{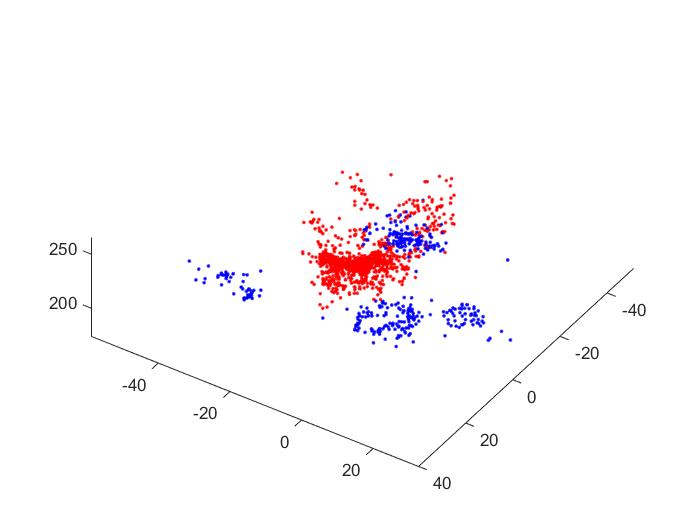}\\
 & (d)\\
\includegraphics[width=0.2\linewidth]{Figs/ball_img.png}&  
\includegraphics[width=0.24\linewidth]{Figs/ball_front.jpg}&
\includegraphics[width=0.24\linewidth]{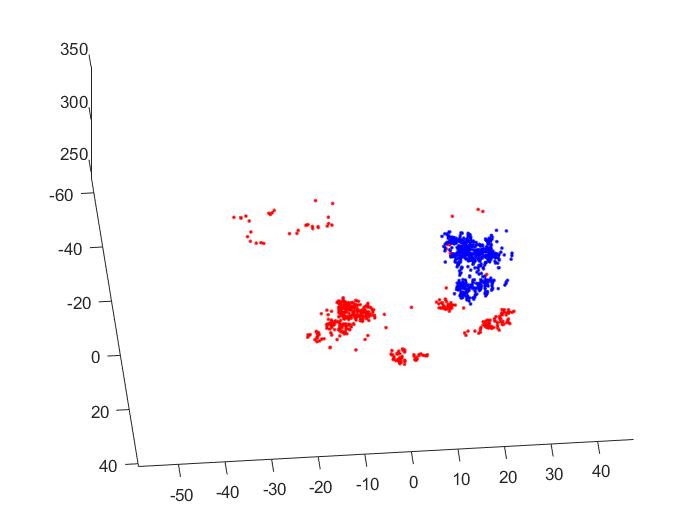}\\
 & (e)\\
\end{tabular}
\end{center}
\caption{Results of our method on the sequences extracted from the ETH Dataset. Left: image from the sequence, Middle: front view, Right: top view.}
\label{fig:ETH_results1}
\end{figure}

The quality of the images in these sequences is lower than in the case of our dataset, the images are darker and every take contains only two images. Moreover, the scene contains the person, who moves the objects. Still, our method is able to reconstruct these scenes and segment the objects. Examples of the images from the sequences together with the results are shown in Figure~\ref{fig:ETH_results1}. The properties of the resulting models are shown in Table \ref{tab:ETH_results}. In the most of the cases, the person is segmented together with the foreground, as in the sequences the person moves together with the foreground object. In sequences (a), (i) the person is segmented together with the background. In sequences (b), (c), (d), (g) the person is not reconstructed. In sequence (k) the hand of the person is assigned to the foreground, while the rest of the body is assigned to the background. In sequence (l), the background and the foreground are swapped. Model (c) has been reconstructed together with its shadow.

The results on the sequences extracted from the ETH 3D dataset show, that our method is able to reconstruct and segment scenes which are less controlled than the scenes from our dataset. It also shows, that although our semi-dynamic setting may seem to be too restrictive and impractical, it indeed has a practical use when reconstructing fully dynamic scenes observed by multiple synchronized cameras.

\begin{table}[H]
\begin{center}
\begin{tabular}{|c|c|c|c|c|c|}
\hline
ID & Figure & $\#$Points Fg. & $\#$Points Bck. & Person & Swapped\\
\hline\hline
(a) & \ref{fig:ETH_results1} (a) & 353 & 1015 & background & NO\\
(b) & \ref{fig:ETH_results1} (b) & 2734 & 2345 & not reconstructed & NO\\
(c) & \ref{fig:ETH_results1} (c) & 1402 & 852 & not reconstructed & NO\\
(d) & \ref{fig:ETH_results1} (d) & 1130 & 382 & not reconstructed & YES\\
(e) & \ref{fig:ETH_results1} (e) & 616 & 539 & foreground & NO\\
(f) & \ref{fig:ETH_results2} (a) & 624 & 1206 & foreground & NO\\
(g) & \ref{fig:ETH_results2} (b) & 2091 & 2073 & not reconstructed & NO\\
(h) & \ref{fig:ETH_results2} (c) & 2042 & 1362 & foreground & NO\\
(i) & \ref{fig:ETH_results2} (d) & 233 & 1337 & background & NO\\
(j) & \ref{fig:ETH_results2} (e) & 891 & 442 & foreground & NO\\
(k) & \ref{fig:ETH_results3} (a) & 314 & 1202 & split, hand with Fg. & NO\\
(l) & \ref{fig:ETH_results3} (b) & 1564 & 793 & foreground & YES\\
\hline

\end{tabular}
\end{center}
\caption{Results of the models reconstructed from the sequences extracted from ETH 3D dataset.}
\label{tab:ETH_results}
\end{table}

\newpage

\newpage
\begin{figure}[H]
\begin{center}
\begin{tabular}{c c c}
\includegraphics[width=0.2\linewidth]{Figs/frieze_img.png}&  
\includegraphics[width=0.24\linewidth]{Figs/frieze_front.jpg}&
\includegraphics[width=0.24\linewidth]{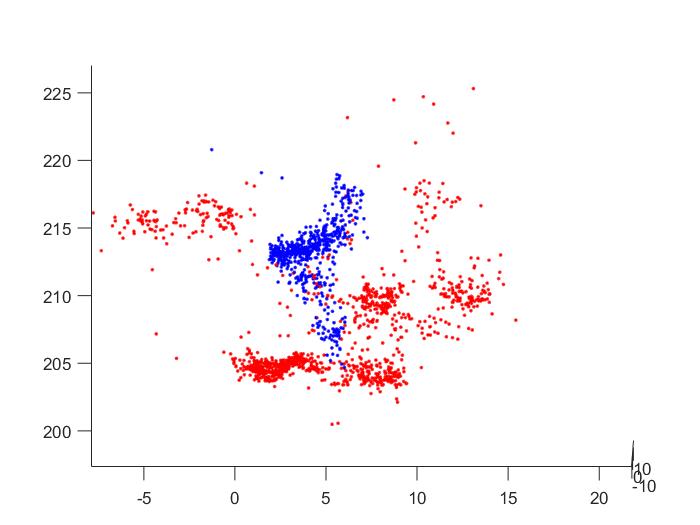}\\
 & (a)\\
\includegraphics[width=0.2\linewidth]{Figs/backpack4_img.png}&  
\includegraphics[width=0.24\linewidth]{Figs/backpack4_front.jpg}&
\includegraphics[width=0.24\linewidth]{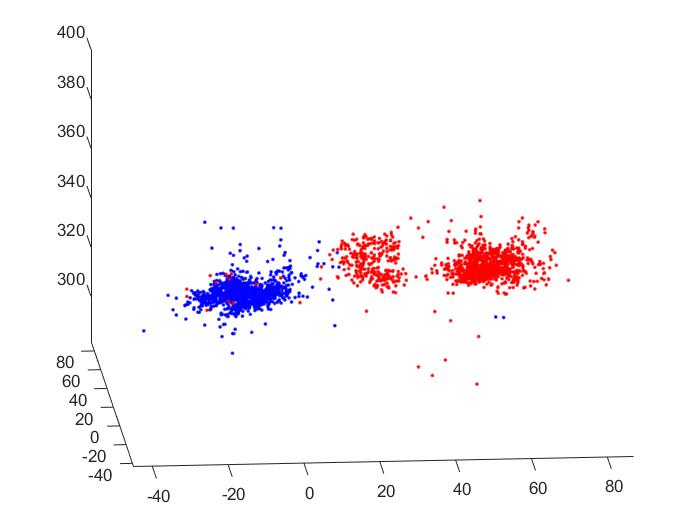}\\
 & (b)\\
\includegraphics[width=0.2\linewidth]{Figs/bear4_img.png}&  
\includegraphics[width=0.24\linewidth]{Figs/bear4_front.jpg}&
\includegraphics[width=0.24\linewidth]{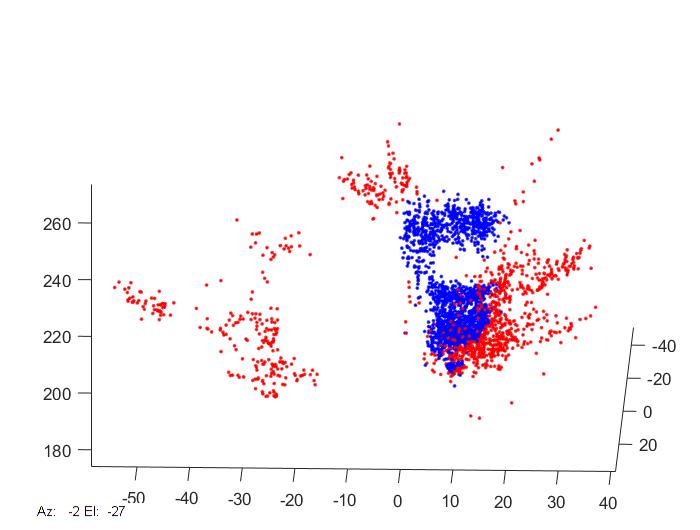}\\
 & (c)\\
\includegraphics[width=0.2\linewidth]{Figs/frieze2_img.png}&  
\includegraphics[width=0.24\linewidth]{Figs/frieze2_front.jpg}&
\includegraphics[width=0.24\linewidth]{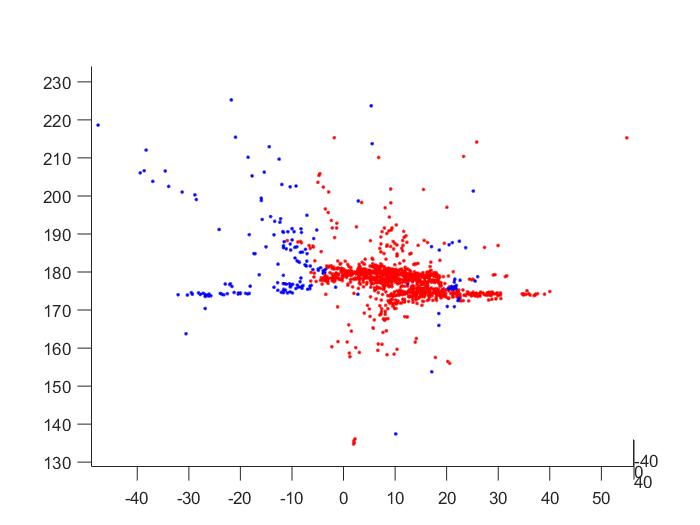}\\
 & (d)\\
\includegraphics[width=0.2\linewidth]{Figs/dinosaur2_img.png}&  
\includegraphics[width=0.24\linewidth]{Figs/dinosaur2_front.jpg}&
\includegraphics[width=0.24\linewidth]{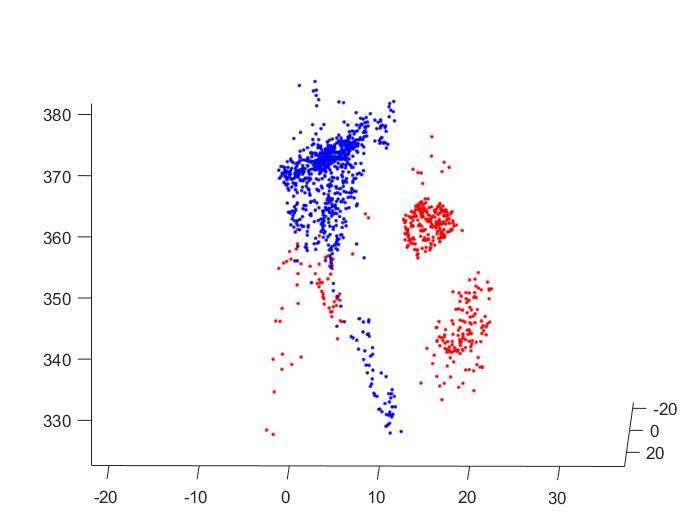}\\
 & (e)\\
\end{tabular}
\end{center}
\caption{Results of our method on the sequences extracted from the ETH Dataset. Left: image from the sequence, Middle: front view, Right: top view.}
\label{fig:ETH_results2}
\end{figure}

\newpage
\begin{figure}[H]
\begin{center}
\begin{tabular}{c c c}
\includegraphics[width=0.2\linewidth]{Figs/ball2_img.png}&  
\includegraphics[width=0.24\linewidth]{Figs/ball2_front.jpg}&
\includegraphics[width=0.24\linewidth]{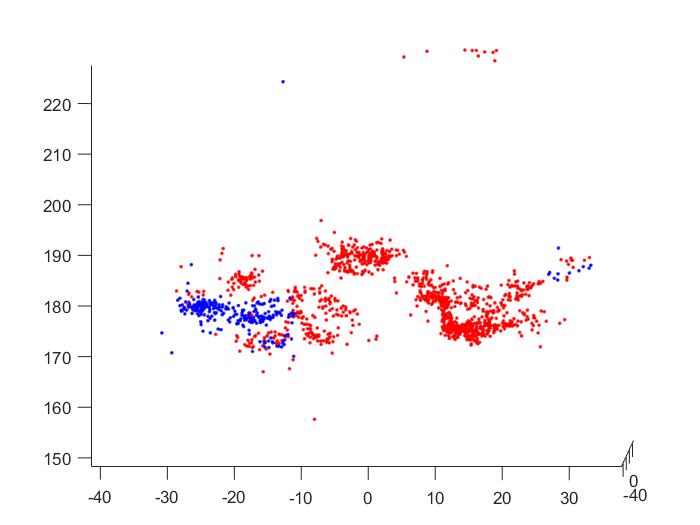}\\
 & (a)\\
\includegraphics[width=0.2\linewidth]{Figs/bear5_img.png}&  
\includegraphics[width=0.24\linewidth]{Figs/bear5_front.jpg}&
\includegraphics[width=0.24\linewidth]{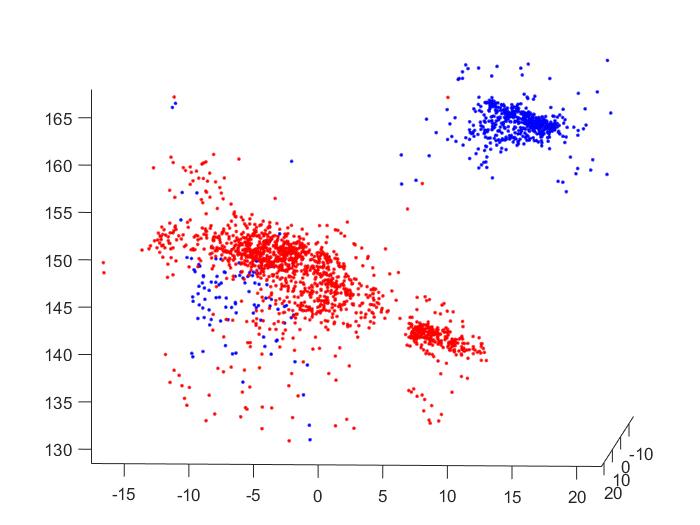}\\
 & (b)\\
\end{tabular}
\end{center}
\caption{Results of our method on the sequences extracted from the ETH Dataset. Left: image from the sequence, Middle: front view, Right: top view.}
\label{fig:ETH_results3}
\end{figure}

\newpage
\section{Foreground motion calculation}\label{sec:motion1}

In this section we show that we can compute the motion $(A_{t,s,\delta_t}^F, a_{t,s,\delta_t}^F)$ of the foreground from a camera pair ($P_{s,\delta_t}^{j_s, B}, P_{s,\delta_t}^{j_s, F}$) obtained with the sequential registration of an image $P_{s,\delta_s}^{j_s}$ capturing configuration $s$ towards model of configuration $t$. Camera $P_{s,\delta_t}^{j_s, B}$ is registered towards the background while $P_{s,\delta_t}^{j_s, F}$ is registered towards the foreground. The computed motion is in coordinate system $\delta_t$. The result is used in Sec.~3.3 of the main paper.

Let $X_{s, i, \delta_s}^{F}$ be a point from the foreground in model of take $s$. Let $x_{s,i}^{F, j_s}$ be the projection of $X_{s, i, \delta_s}^{F}$ onto camera $P_{s,\delta_s}^{j_s}$.

\begin{center}
\begin{figure}[H]
\includegraphics[width=0.9\linewidth]{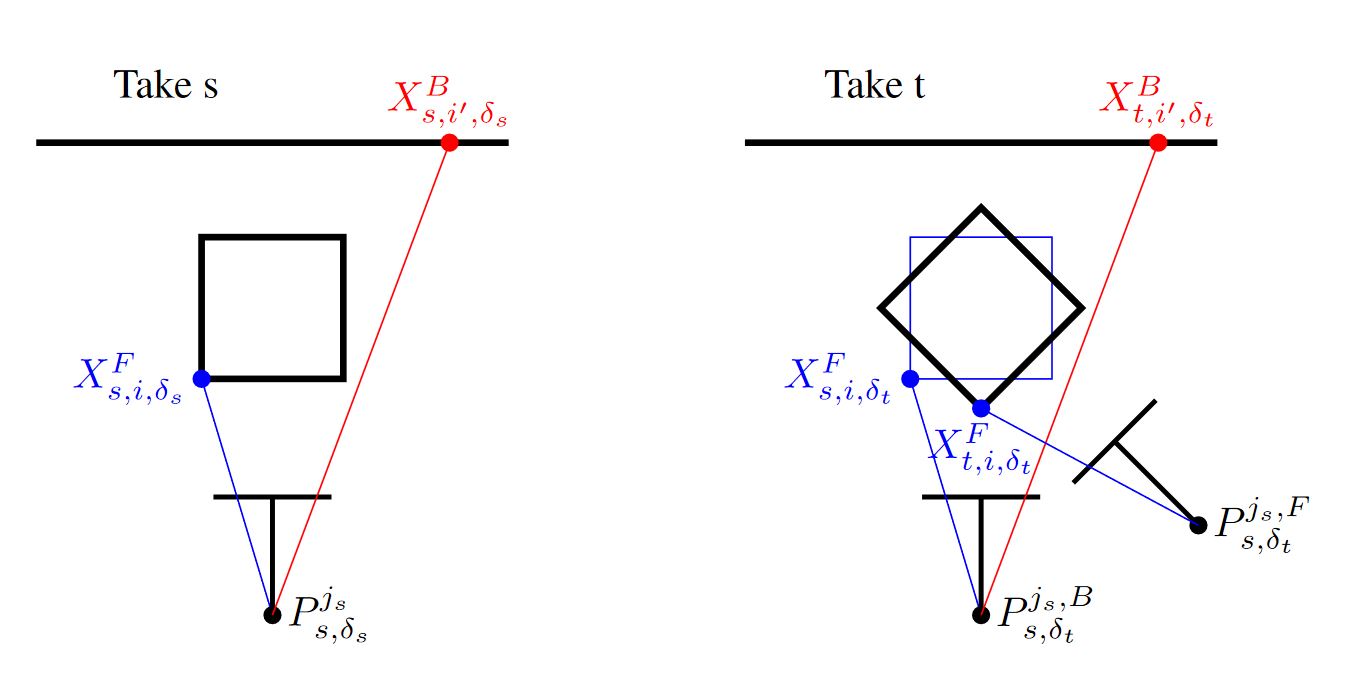}

\caption{Example of a camera from take $s$ after the sequential PnP registration towards take $t$.}
\label{fig:twotakes}
\end{figure}
\end{center}
\noindent
Figure \ref{fig:twotakes} shows that if camera $P_{s,\delta_s}^{j_s}$ from take $s$ is registered towards the foreground points from model of take $t$, it projects the foreground points from model of take $t$ onto the same 2D points onto which they are projected in the model of take $s$. Let $X_{t, i, \delta_t}^F$ be reconstruction of point $X_{s, i, \delta_s}^{F}$ in model of take $t$. Then $x_{s,i}^{F, j_s}$ is a projection of point $X_{t, i, \delta_t}^F$ into camera $P_{s, \delta_t}^{j_s, F}$\\
\\
Figure \ref{fig:twotakes} also shows that if the same camera $P_{s,\delta_s}^{j_s}$ from the second take is registered towards the background points from the model of take $t$, the position of the camera towards the background is the same as in take $s$. This means that if a foreground point is transformed to the configuration of take $s$, it will be projected onto the same 2D feature onto which it is projected in take $s$. Let $X_{s, i, \delta_t}^F = A_{t,s,\delta_t}^F X_{t, i, \delta_t}^{F} + a_{t,s,\delta_t}^F$ be point $X_{t, i, \delta_t}^{F}$ in coordinate system $\delta_t$ and position of take $s$. Then $x_{s,i}^{F, j_s}$ is a projection of point $X_{s, i, \delta_t}^F$ onto camera $P_{s, \delta_t}^{j_s, B}$.\\
\\
The following equation must therefore hold true:\\
\begin{equation}\alpha P_{s, \delta_t}^{j_s, F} \begin{bmatrix} X_{t, i, \delta_t}^F\\ 1 \end{bmatrix} = \alpha P_{s, \delta_t}^{j_s, B} \begin{bmatrix}X_{s, i, \delta_t}^F\\1\end{bmatrix}\\
\end{equation}
We can replace $X_{s, i, \delta_t}^{F}$ with $A_{t,s,\delta_t}^F X_{t, i, \delta_t}^{F} + a_{t,s,\delta_t}^F$ and rewrite the equation using symbols for rotation, center and camera calibration matrix for cameras $P_{s, \delta_t}^{j_s, F}$ and $P_{s, \delta_t}^{j_s, B}$ as:\\
\begin{equation}
    \alpha K R_{s, \delta_t}^{j_s, F} (X_{t, i, \delta_t}^{F} - c_{s, \delta_t}^{j_s, F}) = \alpha K R_{s, \delta_t}^{j_s, B} (A_{t,s,\delta_t}^F X_{t, i, \delta_t}^{F} + a_{t,s,\delta_t}^F - c_{s, \delta_t}^{j_s, B})
\end{equation}
We can eliminate $K$ and $\alpha$ and replace :
\begin{equation}
    R_{s, \delta_t}^{j_s, F} (X_{t, i, \delta_t}^{F} - c_{s, \delta_t}^{j_s, F}) = R_{s, \delta_t}^{j_s, B} (A_{t,s,\delta_t}^F X_{t, i, \delta_t}^{F} + a_{t,s,\delta_t}^F - c_{s, \delta_t}^{j_s, B})
\end{equation}

\noindent Because the equation has to hold true for all $X_{t, i, \delta_t}^F$, it holds true also for $X_{t, i, \delta_t}^F = 0$, which implies that the following two equations also hold true:\\
\begin{equation}
-R_{s, \delta_t}^{j_s, F} c_{s, \delta_t}^{j_s, F} = R_{s, \delta_t}^{j_s, B} a_{t,s,\delta_t}^F - R_{s, \delta_t}^{j_s, B} c_{s, \delta_t}^{j_s, B}\label{eq:m4}\\
\end{equation}
\begin{equation}
R_{s, \delta_t}^{j_s, F} = R_{s, \delta_t}^{j_s, B} A_{t,s,\delta_t}^F\\
\end{equation}
We can easily find the relative rotation $A_{t,s,\delta_t}^F$ from the latter equation as\\
\begin{equation}
A_{t,s,\delta_t}^F = (R_{s, \delta_t}^{j_s, B})^{-1} R_{s, \delta_t}^{j_s, F}\label{eq:rot}\\
\end{equation}
We can substitute \eqref{eq:rot} into \eqref{eq:m4} to obtain:\\
\begin{equation}
    -R_{s, \delta_t}^{j_s, F} c_{s, \delta_t}^{j_s, F} = R_{s, \delta_t}^{j_s, B} a_{t,s,\delta_t}^F - R_{s, \delta_t}^{j_s, B} c_{s, \delta_t}^{j_s, B}\label{eq:m3}\\
\end{equation}
We can find translation $a_{t,s,\delta_t}^F$ as \\
\begin{equation}
a_{t,s,\delta_t}^F = c_{s, \delta_t}^{j_s, B} - A_{t,s,\delta_t}^F c_{s, \delta_t}^{j_s, F}\label{eq:m2}\\
\end{equation}

\noindent We can compute the motions of the foreground from take $s$ to take $t$ according to equations \eqref{eq:rot}, \eqref{eq:m2}, if we have a pair of cameras obtained with sequential registration of a camera $P_{s, \delta_s}^{j_s}$ towards take $t$. One camera pair is enough for finding the relative motion.\\

\section{Additional criterion for local observation grouping}\label{sec:loc-group-s}

Local observation grouping, which is based on common observed points, is described in Main paper in Section~3.3. In this section, we describe an additional criterion, which is based on the motion of the object. This criterion compares the motion of the object, which is computed from the pair of sequentially registered cameras according to Section \ref{sec:motion1}. As this criterion requires multiple pairs from the same take to be registered towards the current take, it is suitable for datasets, where every take contains many images. Now, we will show that the cameras can be clustered by the motion of the foreground, which can be computed from the sequentially registered poses.



If poses $P_{s, \delta_t}^{j_s, B}$ and $P_{s, \delta_t}^{j_s, F}$ are known, the motion $(A_{t, s, \delta_t}^F, a_{t, s, \delta_t}^F)$ can be computed by \eqref{eq:rot}, \eqref{eq:m2}. Note that the motion \eqref{eq:rot}, \eqref{eq:m2} is in coordinate system $\delta_t$ of take $t$.

Let us have two different poses $P_{s, \delta_t}^{j_s, 1}, P_{s, \delta_t}^{j_s, 2} \in \mathcal{P}_{s, \delta_t}^{j_s}$  of the same camera $P_{s, \delta_s}^{j_s}$ Suppose that we follow formulas \eqref{eq:rot}, \eqref{eq:m2} to compute the motion. Four different situations can occur

\noindent \textbf{1. Forward motion} Pose $P_{s, \delta_t}^{j_s, 1}$ observes the background and pose $P_{s, \delta_t}^{j_s, 2}$ observes the foreground; then formulas \eqref{eq:rot}, \eqref{eq:m2} give motion $(A_{t, s, \delta_t}^F, a_{t, s, \delta_t}^F)$.

\noindent \textbf{2. Backward motion} Pose $P_{s, \delta_t}^{j_s, 1}$ observes the foreground and pose $P_{s, \delta_t}^{j_s, 2}$ observes the background. It can be easily shown that formulas \eqref{eq:rot}, \eqref{eq:m2} give motion $(A_{s, t, \delta_t}^F, a_{s, t, \delta_t}^F)$, which is the inverse of the original motion $(A_{t, s, \delta_t}^F, a_{t, s, \delta_t}^F)$.

\noindent \textbf{3. Zero motion} Both poses observe the same object; formulas \eqref{eq:rot}, \eqref{eq:m2} give zero motion, i.e., \ it is close to $(I, 0)$.

\noindent \textbf{4. Wrong result} At least one of the poses is badly estimated, \eqref{eq:rot}, \eqref{eq:m2} give a result different from the previous ones.

For given takes $s$, $t$ we select all the sets of poses $\mathcal{P}_{s,\delta_t}^{j_s}, j_s \in [1, m_s]$ obtained with sequential registration of cameras from take $s$ towards take $t$. From each of these sets, we extract camera pairs $P_{s, \delta_t}^{j_s, 1}, P_{s, \delta_t}^{j_s, 2}$ and then we compute the motion from these camera pairs using formulas~\eqref{eq:rot}, \eqref{eq:m2}.

\begin{figure}[t]
\begin{center}
\begin{tabular}{c c}
\includegraphics[height=0.27\linewidth]{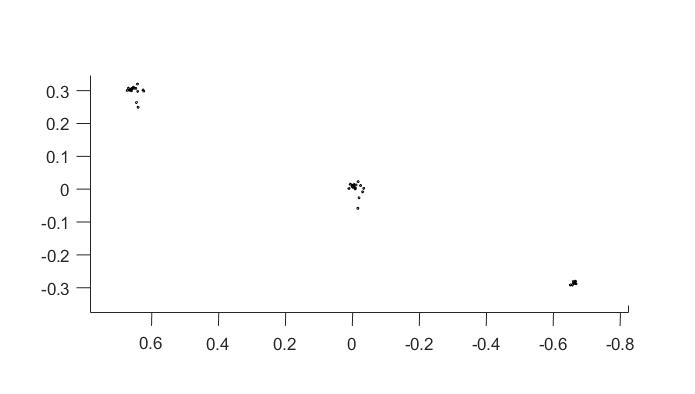}&  
\includegraphics[height=0.24\linewidth]{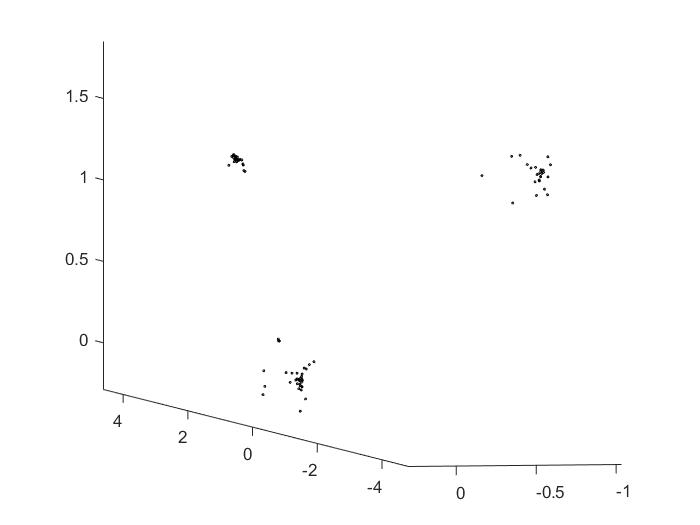}\\
(a) & (b)\\
\end{tabular}
\end{center}
\caption{Euler vectors (a) and translations (b) of the motions calculated with \eqref{eq:rot}, \eqref{eq:m2}. Three clusters representing the forward, backward and zero motions are observable.}
\label{fig:motions}
\end{figure}

The computed motions cluster around the correct motion values $(A_{t, s, \delta_t}^F, a_{t, s, \delta_t}^F)$, $(A_{s, t, \delta_t}^F, a_{t, s, \delta_t}^F)$, $(0, I)$, Fig.~\ref{fig:motions}. First, we remove the zero motions with spectral clustering~\cite{Luxburg07}. Then, we use Sequential RANSAC to find the clusters of the forward motions and the backward motions.

For every two takes $s$, $t$, $s \neq t$, we have obtained clusters $\mathcal{C}_{s, t}^{l}, l \in [1, q_{s,t}]$ of ordered camera pairs. In the forward motion cluster, the first camera in a pair observes the background and the second one observes the foreground. In the backward motion cluster it is vice versa. We can merge the clusters which observe the same reconstruction using the linkage scheme described in Sec~3.3 of the main paper. Then, the clusters obtained in Sec~3.3 of the main paper, are further merged according to Sec~3.5 of the main paper.

\section{Calculation of transformation of points between takes}\label{sec:trans}

In this section we compute transformation $(B_{s,t}^o, b_{s,t}^o, \beta_{s,t}^o)$ of the points of object $o$ from model of take $s$ to model of take $t$ from a camera $P_{s,\delta_s}^{j_s}$ from take $s$ and its pose $P_{s,\delta_t}^{j_s, o}$ registered towards object $o$ in model of take $t$. The result of this section is used in Section~\ref{sec:cameras}.

Let $X_{s, i, \delta_s}^{o}$ be an arbitrary point from object $o$ in model of take $s$. Let $X_{t, i, \delta_t}^{o}$ be the point $X_{s, i, \delta_s}^{o}$ in model of take $t$. Then, there holds true:

\begin{equation}
    X_{t,i,\delta_t}^{o} = \beta_{s,t}^o B_{s,t}^o X_{s,i,\delta_s}^{o} + b_{s,t}^o\label{eq:b0}
\end{equation}

\begin{center}
\includegraphics[width=0.8\linewidth]{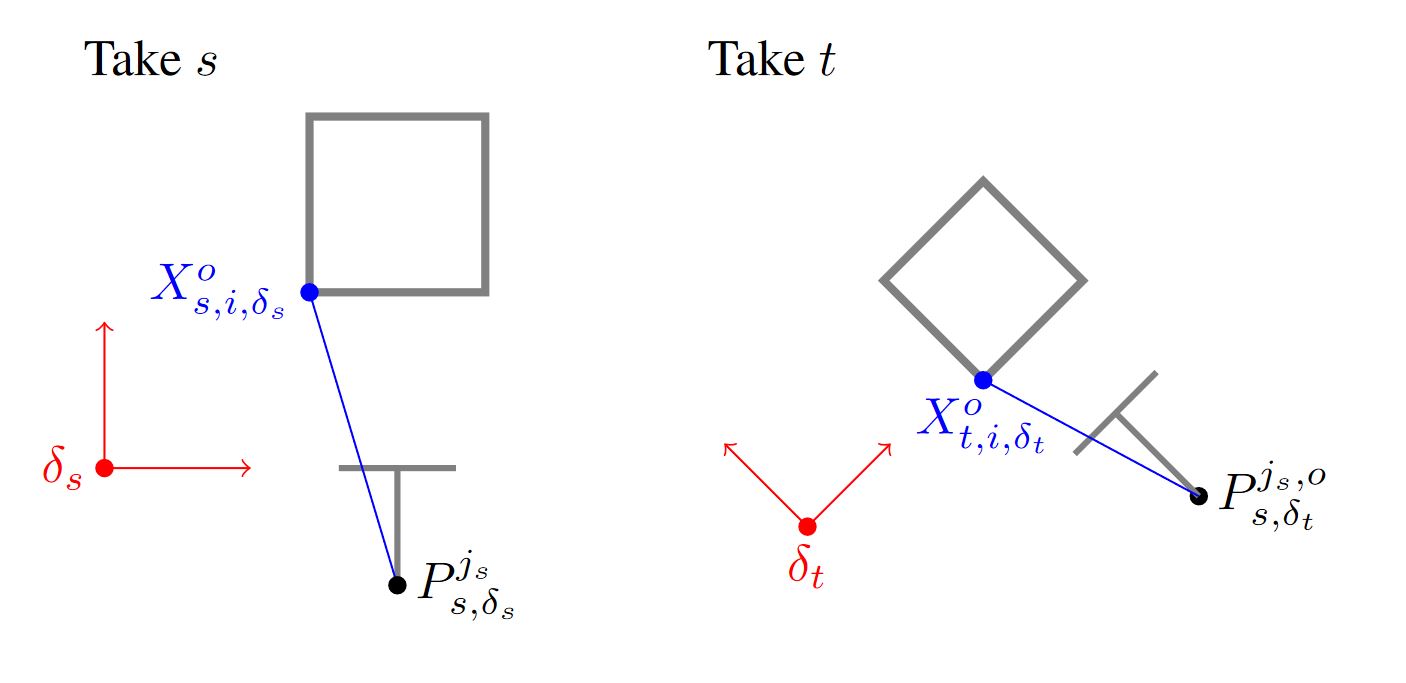}
\end{center}

\noindent The 3D point is projected onto the same point on the camera in both reconstructions, so there holds true:\\
\begin{equation}
    \overrightarrow{x} = \alpha P_{s,\delta_t}^{j_s, o} \begin{bmatrix} X_{t, i, \delta_t}^{o}\\ 1 \end{bmatrix} = \alpha P_{s,\delta_s}^{j_s} \begin{bmatrix} X_{s, i, \delta_s}^{o}\\ 1 \end{bmatrix} 
\end{equation}
We assume that both poses belong to the same camera, so the camera calibration matrix $K$ is the same for both pictures. We rewrite the equation using symbols for rotation, center and camera calibration matrix as:
\begin{equation}
    \alpha_1 K R_{s, \delta_t}^{j_s, o} (X_{t, i, \delta_t}^{o} - c_{s, \delta_t}^{j_s, o}) = \alpha_2 K R_{s, \delta_s}^{j_s} (X_{s, i, \delta_s}^{o} - c_{s, \delta_s}^{j_s})
\end{equation}
We eliminate $K$, introduce $\sigma = \frac{\alpha_1}{\alpha_2}$ and replace $X_{t, i, \delta_t}^{o}$ with $\beta_{s,t}^o B_{s,t}^o X_{s,i,\delta_s}^{o}~+~b_{s,t}^o$:
\begin{equation}
    \sigma R_{s, \delta_t}^{j_s, o} (\beta_{s,t}^o B_{s,t}^o X_{s,i,\delta_s}^{o} + b_{s,t}^o - c_{s, \delta_t}^{j_s, o}) = R_{s, \delta_s}^{j_s} (X_{s, i, \delta_s}^{o} - c_{s, \delta_s}^{j_s})
\end{equation}
This equation has to hold true for every $X_{s, i, \delta_s}^{o}$, so following two equations hold true:\\
\begin{equation}
    \sigma \beta_{s,t}^o R_{s,\delta_t}^{j_s, o} B_{s,t}^o = R_{s,\delta_s}^{j_s}\label{eq:b1}
\end{equation}
\begin{equation}
    \sigma R_{s,\delta_t}^{j_s,o} b_{s,t}^o - \sigma R_{s,\delta_t}^{j_s, o} c_{s,\delta_t}^{j_s, o} = -R_{s,\delta_s}^{j_s} c_{s,\delta_s}^{j_s}\label{eq:b2}
\end{equation}
Because $R_{s,\delta_t}^{j_s,o}$, $B_{s,t}^o$ and $R_{s,\delta_s}^{j_s}$ are all rotation matrices, in order for the first equation to hold true, $\sigma \beta_{s, t}^o$ must be equal to 1, so we can compute the rotation $B_{s,t}^o$ from \eqref{eq:b1} as:
\begin{equation}
    B_{s,t}^o = (R_{s,\delta_t}^{j_s, o})^{-1} R_{s,\delta_s}^{j_s}\label{eq:b3}
\end{equation}
Rotation between coordinate systems can be found with one camera pair.\\
\\
We substitute $\beta_{s, t}^o$ into \eqref{eq:b2}:\\
\begin{equation}
    R_{s,\delta_t}^{j_s,o} b_{s,t}^o - R_{s,\delta_t}^{j_s,o} c_{s,\delta_t}^{j_s,o} = -\beta_{s,t}^o R_{s,\delta_s}^{j_s} c_{s,\delta_s}^{j_s}\label{eq:b5}
\end{equation}
We can rewrite \eqref{eq:b5} in matrix form as:
\begin{equation}
    \begin{bmatrix} R_{s, \delta_t}^{j_s, o} & R_{s, \delta_s}^{j_s} c_{s, \delta_s}^{j_s} \end{bmatrix} \begin{bmatrix} b_{s, t}^o \\ \beta_{s, t}^o \end{bmatrix} = R_{s, \delta_s}^{j_s} c_{s, \delta_s}^{j_s} \label{eq:b4}
\end{equation}
If only one camera pair is considered, equation \eqref{eq:b4} is underdetermined. It can be solved with least squares if at least two camera pairs are available.

\section{Transformation of the cameras}\label{sec:cameras}

In this section we describe the transformation of the cameras to the coordinate system of the reference take $r$. The result is used in Sec.~3.6 in the main paper. Let $P_{t, \delta_s}^{j_t, o}$ be a camera from take $t$ which is registered towards points from object $o$ in model $s$. If $s = t$, the camera observes both objects. In that case we define $o = A$.

The task is to find poses $P_{t, \delta_r}^{j_t, B}$, $P_{t, \delta_r}^{j_t, F}$ of camera $P_{t, \delta_s}^{j_t, o}$ transformed to take $r$, where $P_{t, \delta_r}^{j_t, B}$ is towards the background and $P_{t, \delta_r}^{j_t, F}$ is towards the foreground. Apart from $P_{t, \delta_s}^{j_t, o}$, we know the transformations $(B_{u, r}^B, b_{u, r}^B, \beta_{u, r}^B)$, $(B_{u, r}^F, b_{u, r}^F, \beta_{u, r}^F)$, of the points from take $u \in [1,k]$ to take $r$, and the motions $(A_{u,r,\delta_r}^F, a_{u,r,\delta_r}^F)$ of the foreground from take $u \in [1,k]$ to take $r$.

First, we find $P_{t, \delta_r}^{j_t, B}$, then we follow to find the transformation which transforms $P_{t, \delta_r}^{j_t, B}$ to $P_{t, \delta_r}^{j_t, F}$. The formula to compute $P_{t, \delta_r}^{j_t, B}$ depends on $s, t, o$.

\paragraph{\boldmath $s=r, o \in \{A, B\}$}
The camera is already in the desired position, no transformation is necessary.

\paragraph{\boldmath $s \neq r; o \in \{A, B\}$}
The camera is in another coordinate system.

\begin{center}
\includegraphics[width=0.8\linewidth]{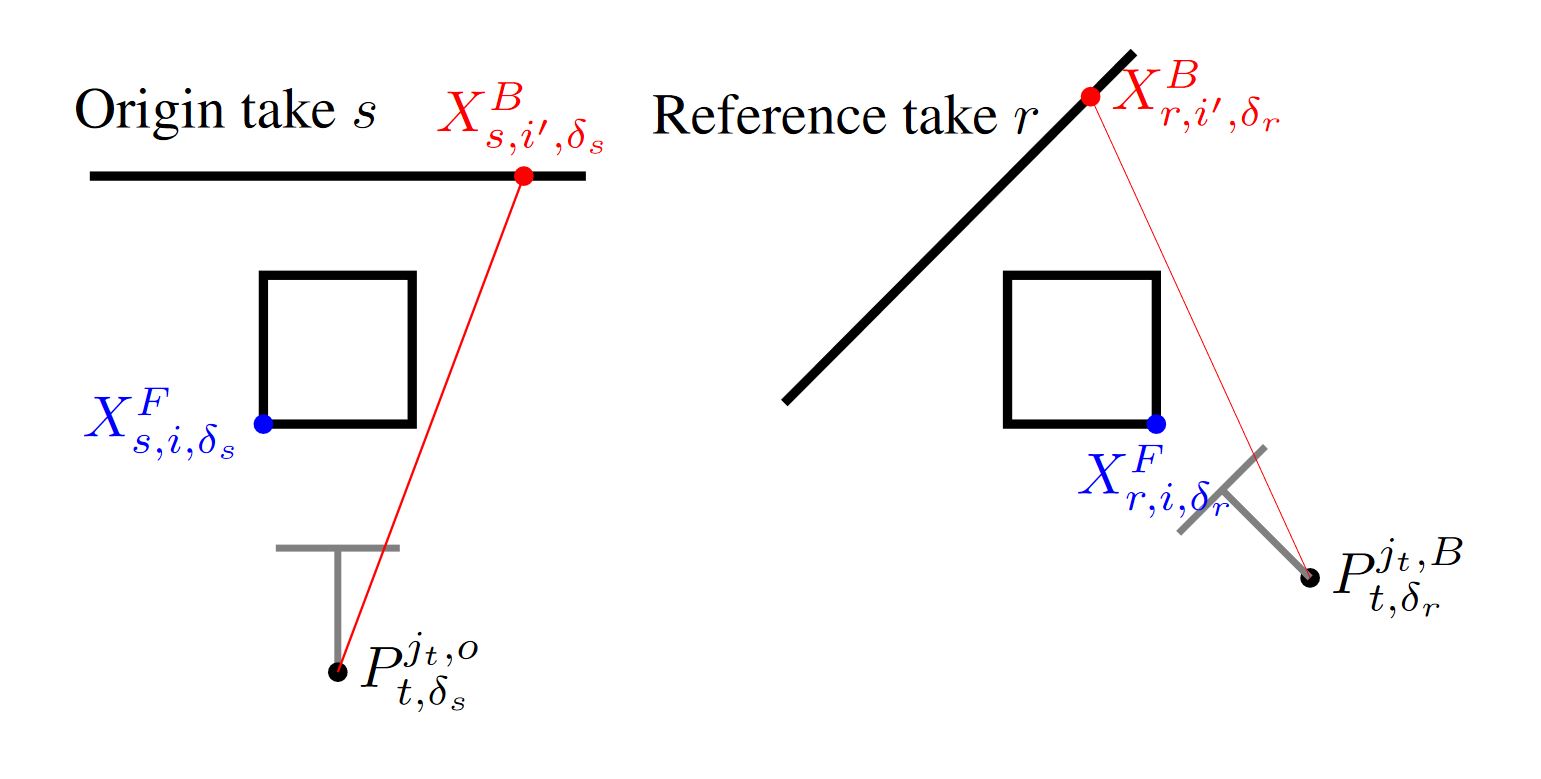}
\end{center}
\noindent
The transformation to the reference take is performed according to the equations \eqref{eq:b3}, \eqref{eq:b5} as:
\begin{equation}
    R_{t, \delta_r}^{j_t, B} = R_{t, \delta_s}^{j_t, o} (B_{s,r}^B)^{-1}
\end{equation}
\begin{equation}
    c_{t, \delta_r}^{j_t, o} = b_{s,r}^B + \beta_{s,r}^B (R_{t, \delta_r}^{j_t, B})^{-1} R_{t, \delta_s}^{j_t, o} c_{t, \delta_s}^{j_t, o}
\end{equation}
\noindent
Transformation $(B_{s, r}^B, b_{s, r}^B, \beta_{s, r}^B)$ transforms the points from the background in the coordinate system $s$ to the reference take $r$. The transformed camera with pose $(R_{t, \delta_r}^{j_t, B}, c_{t, \delta_r}^{j_t, B})$, therefore, observes the points from the background in the reference take, which is the desired result.

\paragraph{\boldmath $s \neq r; o = F; t = r$}
The camera comes from the reference take $r$ but it has been registered onto another reconstruction towards the foreground points.

\begin{center}
\includegraphics[width=0.8\linewidth]{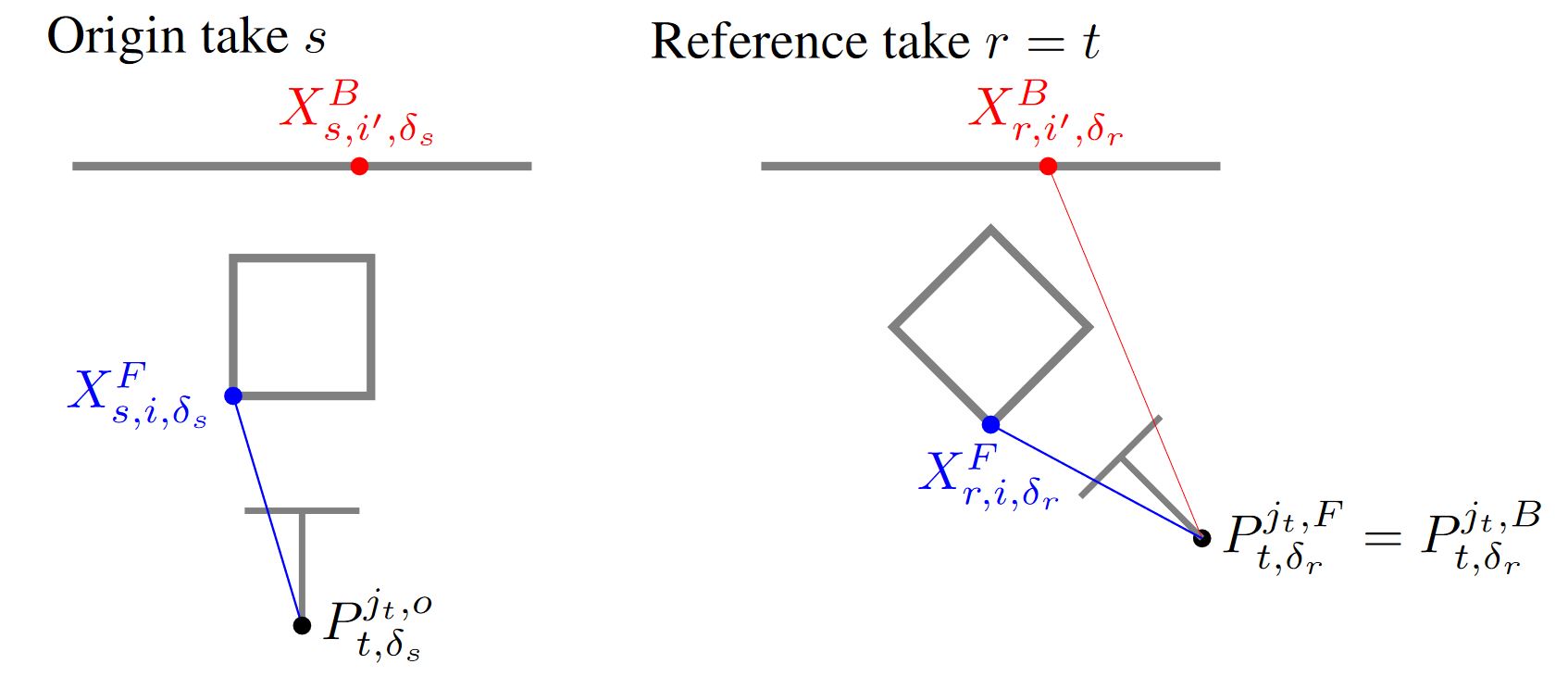}
\end{center}
\noindent
The pose of the camera $P_{t, \delta_r}^{j_t, F}$ observing the foreground points in the reference take transformed from take $s$ by $(B_{s, r}^F, b_{s,r}^F, \beta_{s, r}^F)$ can be computed according to the equations \eqref{eq:b3}, \eqref{eq:b5}:
\begin{equation}
    R_{t, \delta_r}^{j_t, F} = R_{t, \delta_s}^{j_t, o} (R_{s,r}^O)^{-1} \label{eq:mc1}
\end{equation}
\begin{equation}
    c_{t, \delta_r}^{j_t, F} = \overrightarrow{o}_{s,r}^O + \sigma_{s,r} (R_{t, \delta_r}^{j_t, F})^{-1} R_{t, \delta_s}^{j_t, o} c_{t, \delta_s}^{j_t, o} \label{eq:mc2}
\end{equation}
The pose of the transformed camera is towards the foreground points. But because the camera arises from the reference take, the pose towards the foreground and the background in the reference take is the same, so $R_{t, \delta_r}^{j_t, B} = R_{t, \delta_r}^{j_t, F}, c_{t, \delta_r}^{j_t, B} = c_{t, \delta_r}^{j_t, F}$.

\paragraph{\boldmath $s = r; o = F; t \neq r$}
The camera is already in the target coordinate system but it is registered towards the foreground. It has to be moved to the position where it would observe the background points.\\

\begin{center}
\includegraphics[width=0.8\linewidth]{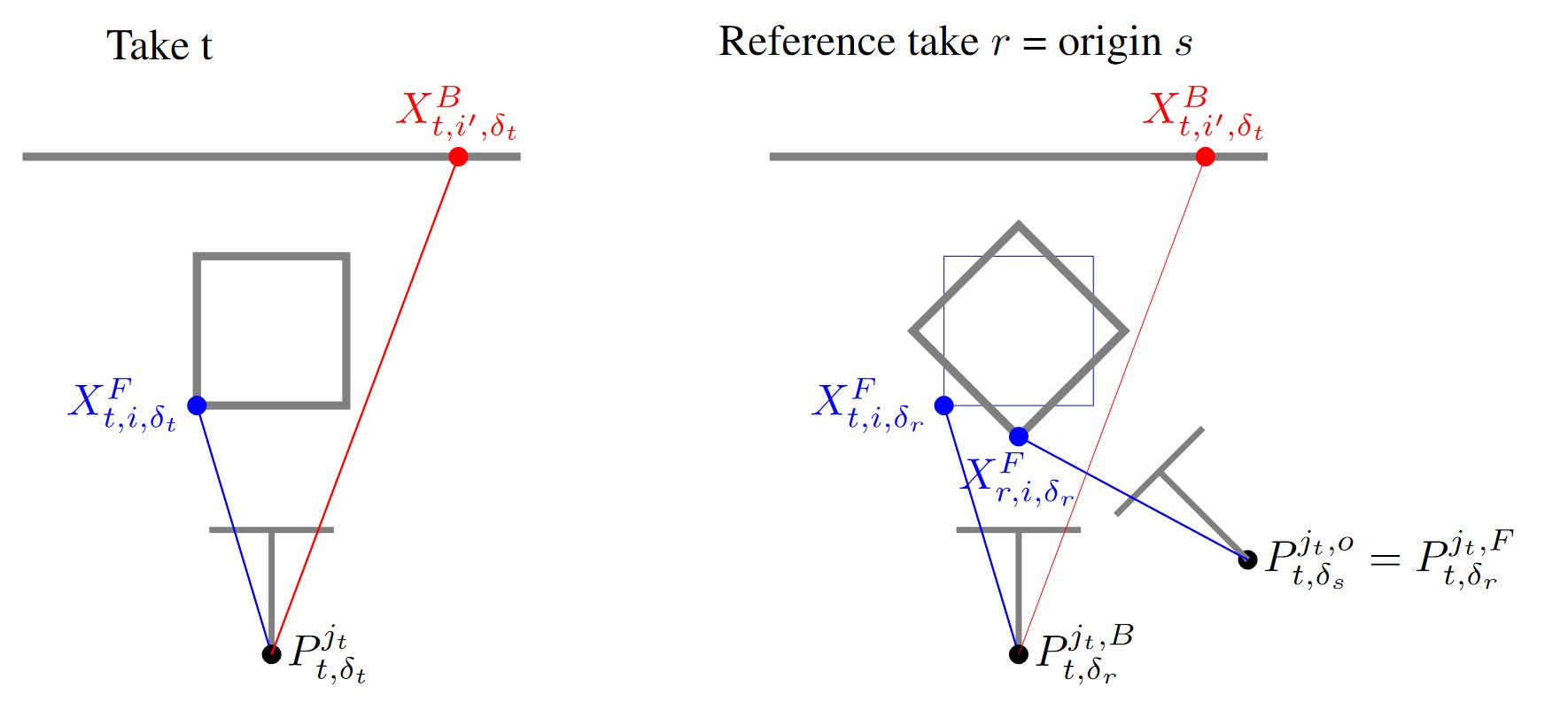}
\end{center}
\noindent
Camera $P_{t, \delta_s}^{j_t, o}$ observes point $X_{t, i, \delta_t}^F$ on position $X_{r, i, \delta_r}^F$ where it is in the reference take. If the camera was registered towards the background, it would observe the point on its original position in take $t$: $X_{t, i, \delta_r}^F$.
\begin{equation}
    X_{t, i, \delta_r}^F = A_{r,t,\delta_r}^F X_{r, i, \delta_r}^F + a_{r,t,\delta_r}^F
\end{equation}
Rotation and centre of the camera registered towards the background points can be found according to equations \eqref{eq:rot}, \eqref{eq:m3} as:
\begin{equation}
    R_{t, \delta_r}^{j_t, B} = R_{t, \delta_s}^{j_t, o} (A_{r,t,\delta_r}^F)^{-1} = R_{t, \delta_s}^{j_t, o} A_{t,r,\delta_r}^F\label{eq:mc3}
\end{equation}
\begin{equation}
    c_{t, \delta_r}^{j_t, B} = a_{r,t,\delta_r}^F + (R_{t, \delta_r}^{j_t, B})^{-1} R_{t, \delta_s}^{j_t, o} c_{t, \delta_s}^{j_t, o}\label{eq:mc4}
\end{equation}

\paragraph{\boldmath $s \neq r; o = F; t \neq r$}
The camera does not come from the reference take, so transformation $(B_{s, r}^F, b_{s, r}^F, \beta_{s, r}^F)$ does not bring the desired result. This case can however be solved as a combination of the two previous ones.

\begin{center}
\includegraphics[width=1\linewidth]{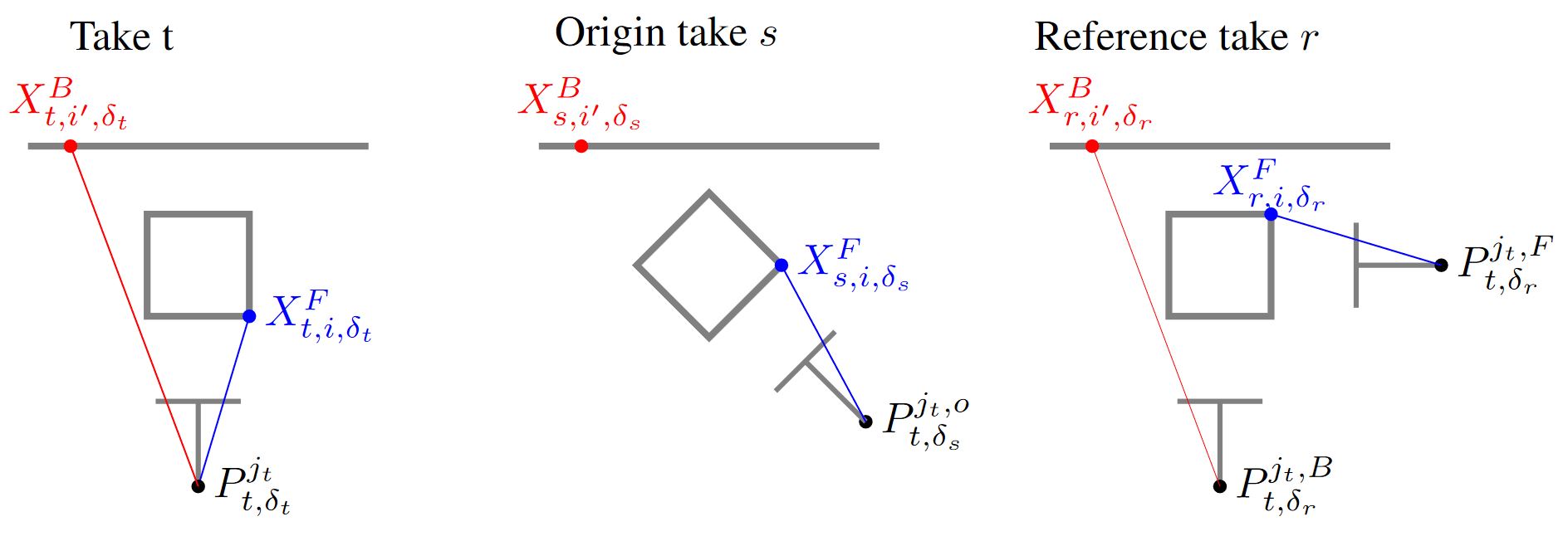}
\end{center}
\noindent
If camera $P_{t, \delta_s}^{j_t, o}$ is transformed with $(B_{s, r}^F, b_{s,r}^F, \beta_{s, r}^F)$ according to equations \eqref{eq:mc1}, \eqref{eq:mc2} to the reference take $r$, the transformed camera $P_{t, \delta_r}^{j_t, F}$ observes the foreground points of the reference take on the same features where the original camera $P_{t, \delta_s}^{j_t, o}$ observed the foreground points in take $s$.
\begin{equation}
    R_{t, \delta_r}^{j_t, F} = R_{t, \delta_s}^{j_t, o} (B_{s,r}^F)^{-1}
\end{equation}
\begin{equation}
    c_{t, \delta_r}^{j_t, F} = b_{s,r}^F + \beta_{s,r}^F (R_{t, \delta_r}^{j_t, F})^{-1} R_{t, \delta_s}^{j_t, o} c_{t, \delta_s}^{j_t, o}
\end{equation}
\noindent
This converts the problem to the previous one where the task is to transform the camera $P_{t,\delta_r}^{j_t,F}$ observing the foreground to the camera $P_{t,\delta_r}^{j_t,B}$ which observes the background. This can be done according to the equations \eqref{eq:mc3}, \eqref{eq:mc4}
\begin{equation}
    R_{t, \delta_r}^{j_t, B} = R_{t, \delta_r}^{j_t, F} (A_{r,t,\delta_r}^F)^{-1} = R_{t, \delta_r}^{j_t, F} A_{t,r,\delta_r}^F
\end{equation}
\begin{equation}
    c_{t, \delta_r}^{j_t, B} = a_{r,t,\delta_r}^F + (R_{t, \delta_r}^{j_t, B})^{-1} R_{t, \delta_s}^{j_t, o} c_{t, \delta_s}^{j_t, o}
\end{equation}\\

\paragraph{\textbf{Finding the camera towards the foreground}}
We have found the camera $P_{t, \delta_r}^{j_t, B}$ towards the background. The pose $P_{t, \delta_r}^{j_t, F}$ towards the foreground can be calculated from the pose $P_{t, \delta_r}^{j_t, B}$ using the inverse of the equations \eqref{eq:mc3}, \eqref{eq:mc4}:
\begin{equation}
    R_{t, \delta_r}^{j_t, F} = R_{t, \delta_r}^{j_t, B} A_{r,t,\delta_r}^F\label{eq:mc5}
\end{equation}
\begin{equation}
    c_{t, \delta_r}^{j_t, F} = (A_{r,t,\delta_r}^F)^{-1} (c_{t, \delta_r}^{j_t, B} - a_{r,t,\delta_r}^F)\label{eq:mc6}
\end{equation}
\noindent
The observations of the camera $P_{t, \delta_r}^{j_t}$ are split in such way, that the camera $P_{t, \delta_r}^{j_t, B}$ observes the points which belong to the background and the camera $P_{t, \delta_r}^{j_t, F}$ observes the points which belong to the foreground.

\end{document}